\documentclass[10pt,twocolumn,letterpaper]{article}

\usepackage{cvpr}

\usepackage{multirow}
\usepackage{subcaption}
\usepackage{makecell}
\usepackage{tabularx}
\usepackage[accsupp]{axessibility}

\definecolor{cvprblue}{rgb}{0.21,0.49,0.74}
\usepackage[pagebackref,breaklinks,colorlinks,allcolors=cvprblue]{hyperref}

\def\paperID{23} %
\def\confName{CVPR}
\def\confYear{2025}

\newcommand{\method}[1]{FreBIS}
\newcommand{\level}[1]{N}

\title{\method~: Frequency-Based Stratification for \\Neural Implicit Surface Representations}

\author{Naoko Sawada\textsuperscript{1,2}\quad
Pedro Miraldo\textsuperscript{1}\quad
Suhas Lohit\textsuperscript{1}\quad
Tim K. Marks\textsuperscript{1}\quad 
Moitreya Chatterjee\textsuperscript{1}\\[5pt]
\textsuperscript{1}Mitsubishi Electric Research Laboratories~(MERL), Cambridge, MA, USA\\
\textsuperscript{2}Information Technology R\&D Center, Mitsubishi Electric Corporation, Kanagawa, Japan\\
{\tt\small Sawada.Naoko@df.MitsubishiElectric.co.jp,\{miraldo,slohit,tmarks,chatterjee\}@merl.com}
}

\begin{document}
\maketitle
\begin{abstract}

Neural implicit surface representation techniques are in high demand for advancing technologies in augmented reality/virtual reality, digital twins, autonomous navigation, and many other fields.
With their ability to model object surfaces in a scene as a continuous function, 
such techniques have made remarkable strides recently, especially over classical 3D surface reconstruction methods, such as those that use voxels or point clouds. 
However, these methods struggle with scenes that have varied and complex surfaces principally because they model any given scene with a single encoder network that is tasked to capture all of low through high-surface frequency information in the scene simultaneously.
In this work, we propose a novel, neural implicit surface representation approach called \method~ to overcome this challenge.
\method~ works by stratifying the scene based on the frequency of surfaces into multiple frequency levels, 
with each level (or a group of levels) encoded by a dedicated encoder. 
Moreover, \method~ encourages these encoders to capture complementary information 
by promoting mutual dissimilarity of the encoded features via a novel, redundancy-aware weighting module. 
Empirical evaluations on the challenging BlendedMVS dataset indicate that replacing the standard encoder in an off-the-shelf neural surface reconstruction method with our frequency-stratified encoders yields significant improvements. These enhancements are evident both in the quality of the reconstructed 3D surfaces and in the fidelity of their renderings from any viewpoint.
\end{abstract}
    
\section{Introduction}
\label{sec:intro}

\begin{figure}[tb]
    \centering
    \includegraphics[width=0.8\linewidth]{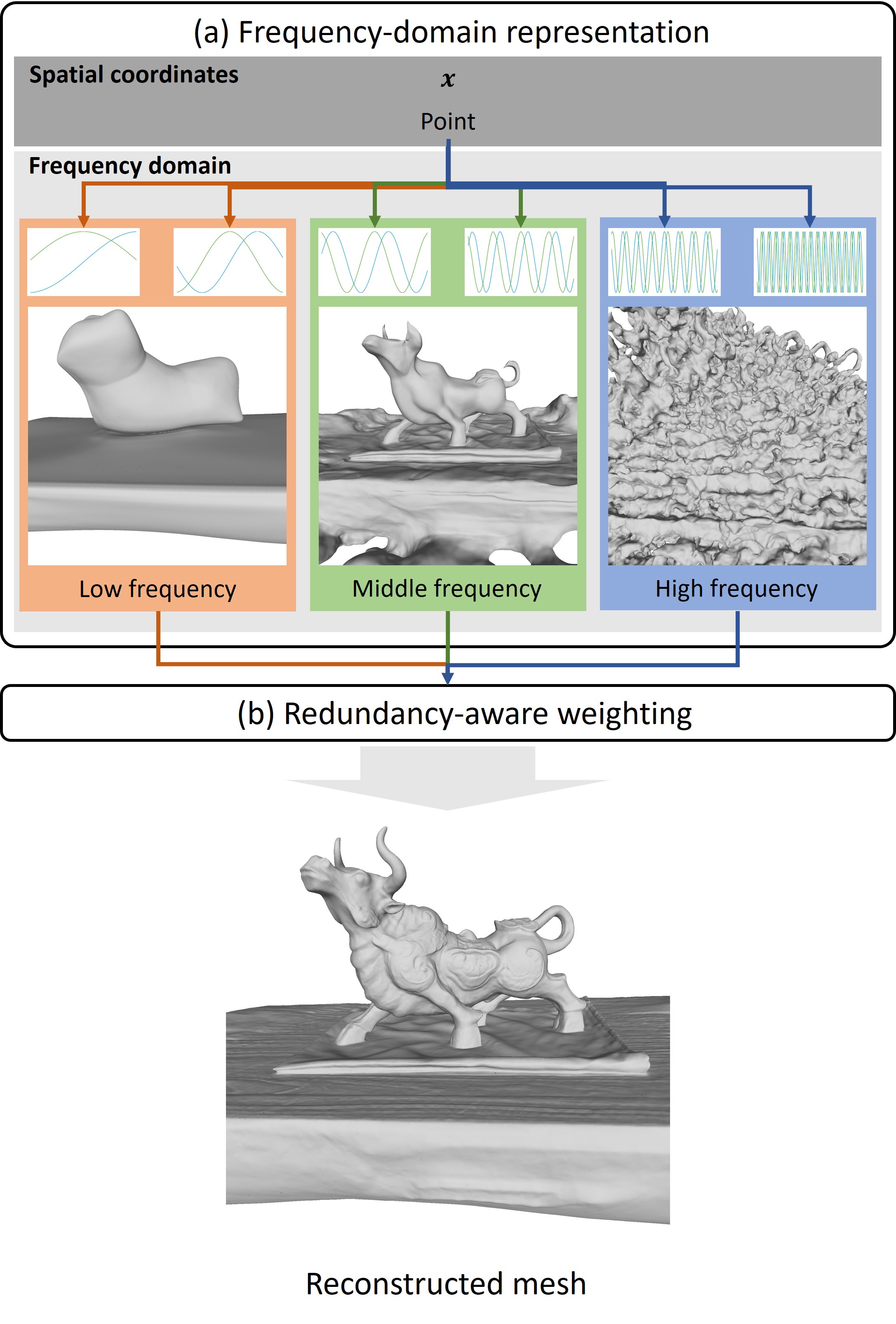}
    \caption{Overview of \method~:
    (a) \emph{Frequency-domain Representation:} \method~ works by mapping the input point coordinate to the frequency domain and encoding it via three frequency-band encoders -- one each for low, middle, and high.
    (b) \emph{Redundancy-aware Weighting}: This module computes weights that indicate the importance of the three encoded features according to the dissimilarity of each to the other two. These weights are then used to combine the encoded features.
    The 3D surface is reconstructed by decoding the combined feature into a SDF value. 
    }
    \label{fig:teaser}
\end{figure}

While a picture is worth a thousand words, yet 2D image understanding methods miss out on critical details, including depth cues and occluded structures, driving research into techniques for reconstructing complete 3D surfaces from images. 
Approaches for reconstructing 3D surfaces find wide use in a broad swathe of applications, 
including Augmented Reality (AR), Virtual Reality (VR), robotics, archaeology
and allow users to easily create 3D content.

Conventional methods for the task of 3D scene reconstruction leverage explicit representations, 
such as voxels~\cite{Broadhurst2001, Bonet1999PoxelsPV,Seitz1997} and point clouds~\cite{PatchMatch2009,Furukawa2010,Galliani2015,Schönberger2016},
where the granularity of the voxels or the 3D points determines the resolution of the reconstructed mesh,
thereby limiting the quality of the reconstruction. 
Neural implicit surface representation methods overcome this challenge by learning continuous functions to model the 3D surfaces, including signed distance functions (SDF)~\cite{yariv_volume_2021, wang_neus_2021} and occupancy fields~\cite{oechsle_unisurf_2021}. These implicit representations can encode 3D geometries at infinite resolution and reduce memory requirements, thereby realizing high-fidelity 3D surface reconstruction from 2D images.

Prior works on neural implicit surface representation~\cite{yariv_volume_2021, wang_neus_2021} and their variants can reconstruct 3D surfaces with high details.
However, their ability to simultaneously represent the correct shape of complex surfaces, while capturing their fine details is limited. This is primarily because they employ a single encoder network that attempts to capture all the various surface frequencies present in the scene (possibly from a very low to a very high one) simultaneously. %

In this paper, we propose \emph{\underline{Fre}quency-\underline{B}ased Stratification for Neural \underline{I}mplicit \underline{S}urface Representation} (\method~) -- a novel approach to neural implicit surface representation,
where multiple encoder networks are specialized to encode different frequency bands so that each encoder can capture complementary information about the scene,
allowing \method~ to effectively learn low-- through high--frequency information simultaneously.
In practice, \method~ employs three encoders dedicated to capturing information in the low--, middle--, and high--frequency bands, respectively, from the scene which is then assimilated and decoded by a single decoder network to
estimate the SDF value and a RGB feature vector which encodes the color information, 
as shown in Fig.~\ref{fig:teaser}~(a).
Thus, instead of a unified latent feature encoding, features corresponding to different frequency bands can be derived from three different encoders.
To effectively combine the disparate information learned by the different encoders, \method~ introduces a novel \emph{redundancy-aware weighting} module, as shown in Fig.~\ref{fig:teaser}~(b). Given the different feature encodings, this module estimates normalized importance scores for each of them and uses them as weights to combine the encodings to derive a unified representation.
Subsequently, a decoder module decodes this unified representation to predict the SDF value and RGB feature, corresponding to a 3D point in the scene.

\method~ makes it possible to recover high-quality surfaces of 3D scenes that contain various levels of detail. Additionally, it provides a flexible mechanism to combine the stratified encoders with any off-the-shelf decoder backbones.
Empirical evaluations on the challenging BlendedMVS~\cite{yao_blendedmvs_2020} dataset show that our strategy of frequency-based stratification results in improved reconstruction of 3D surfaces while better preserving the fidelity of their renderings from any given viewpoint.

In summary, the key contributions of our work are as follows:
\begin{itemize}
    \item A novel, frequency-based 3D surface representation method (called \method~) that works by stratifying the scene into non-overlapping frequency bands. %
    \item \method~ employs a \emph{redundancy-aware weighting} module that encourages the stratified encoders to capture complementary information by promoting mutual dissimilarity of the encoded features. 
    \item Empirical evaluations demonstrate the effectiveness of \method~ on the challenging BlendedMVS~\cite{yao_blendedmvs_2020} dataset.
\end{itemize}

\section{Related Work}
\label{sec:relatedwork}
\noindent\textbf{Early multi-view surface reconstruction methods:}
Multi-view stereo (MVS) technologies have traditionally been used to recover 3D shapes from multiple RGB images of a scene.
Classical MVS approaches can be classified into voxel-based~\cite{Broadhurst2001, Bonet1999PoxelsPV,Seitz1997, SpaceCarving2000,LevelSet1998,HierarchicalVolumetric2006}, point-cloud-based~\cite{pang2021quasidensesimilaritylearningmultiple,Furukawa2010,Goesele2007,PatchMatch2009,Schönberger2016}, and mesh-based~\cite{Faugeras1998,Vu2012,Labatut2007} methods.
While promising, these methods suffer from quantization artifacts, and noisy or disconnected reconstructed points. Moreover, the quality of the recovered surfaces is voxel/point-resolution-dependent.
We, on the other hand, 
learn an implicit, continuous function, resulting in smoother, more detailed, and robust reconstructions.

\noindent\textbf{Neural implicit surface representation approaches:}
Neural implicit surface representation techniques represent a 3D surface as a continuous function defined by a neural network, such as SDF or occupancy function.
Early methods~\cite{niemeyer_differentiable_2020,yariv_multiview_2020} achieved 3D surface reconstruction from multi-view images by leveraging object mask priors.
The advent of NeRF~\cite{mildenhall2020nerfrepresentingscenesneural} heralded a paradigm shift in this field,
integrating implicit surface representation methods with radiance-field-based approaches. For instance, VolSDF~\cite{yariv_volume_2021} and NeuS~\cite{wang_neus_2021} transform SDF into a differentiable volume density, 
enabling 3D surface reconstruction solely from 2D images while also permitting a rendering of the mesh from any viewpoint.
UNISURF~\cite{oechsle_unisurf_2021} formulates occupancy-based implicit surface representation and radiance field in a unified framework.
Different from previous approaches~\cite{niemeyer_differentiable_2020,yariv_multiview_2020}, they eliminate the need for object masks.
These methods have paved the way for newer neural implicit surface representation methods.
Several variants built upon VolSDF~\cite{yariv_volume_2021} and NeuS~\cite{wang_neus_2021} 
enhance the input feature encoder, venturing beyond a simple Multilayer Perceptron (MLP), to be capable of capturing the fine details of the scene~\cite{wang_hf-neus_2022,wang_pet-neus_2023,gu_hive_2024}.
NeuralWarp~\cite{darmon_improving_2022} and Geo-NeuS~\cite{fu_geo-neus_2022} add explicit multi-view geometry constraints to enforce photo consistency and depth consistency across views.
Other approaches~\cite{yu_monosdf_2022,wang_neuris_2022,wang_go-surf_2022,park_h2o-sdf_2024,liang_helixsurf_2023, guo_neural_2023, azinovic_neural_2022, patel_normal-guided_2024} try to enhance the robustness and details of the representation by integrating priors, such as monocular depth and normal estimates, in addition to RGB images.
Recent works~\cite{li_neuralangelo_2023, wu_voxurf_2023, wang_neus2_2023} have leveraged multi-resolution grid structures to accelerate training and boost the accuracy of the reconstructed surfaces.
Some extensions of these approaches~\cite{wu_object-compositional_2022,wu_objectsdf_2023,li_rico_2023} adapt neural implicit surface representations to object-compositional scenes.
Despite the noteworthy strides made by prior methods, to the best of our knowledge, none have looked at the efficacy of stratifying the scene based on surface frequencies as a cue towards achieving improved 3D surface reconstruction and rendering. Additionally, our approach is complementary to many of these approaches and can be integrated with them for possibly additive performance gains.

\noindent\textbf{Neural radiance field (NeRF):}
Some prior works extract explicit 3D surfaces from radiance field representations of 3D scenes obtained via Neural Radiance Field (NeRF)~\cite{mildenhall2020nerfrepresentingscenesneural}.
MobileNeRF~\cite{chen2023mobilenerfexploitingpolygonrasterization}, NeRF2Mesh~\cite{tang2023delicatetexturedmeshrecovery}, NeRFMeshing~\cite{rakotosaona_nerfmeshing_2023}, and BakedSDF~\cite{yariv2023bakedsdfmeshingneuralsdfs} extract an explicit textured mesh from a trained NeRF model, by having a separate network (in addition to the NeRF model) which predicts the SDF value of a point, given a feature encoding of the point and the viewing direction obtained from the NeRF network.
However, these methods require a fully trained NeRF to begin with, which can be prohibitively slow to train. %

\noindent\textbf{Gaussian splatting (GS):}
3D Gaussian Splatting (3DGS)~\cite{kerbl3Dgaussians} has emerged as a fast and accurate novel view synthesis method,
where scenes are modeled as sets of 3D Gaussians, which are splatted in any novel viewing direction to obtain the color. %
To leverage 3DGS for 3D surface representation, SuGaR~\cite{guedon_sugar_2023} introduces a new regularization term to encourage Gaussians to scatter on surfaces,
while Gaussian Surfels~\cite{dai_high-quality_2024} and 2DGS~\cite{Huang_2024} flatten 3D Gaussians into 2D ellipses.
SplatSDF~\cite{splatsdf} and 3DGSR~\cite{lyu_3dgsr_2024} fuse SDF and 3DGS to achieve both high accuracy and efficiency.
While these methods offer fast training and rendering, and some of them achieve surface reconstructions that are comparable in quality to the best implicit methods, however, they result in high memory consumption.
Additionally, some of these methods are sensitive to noise and thereby lack robustness.

\section{Background}
\label{sec:background}

\begin{figure*}
    \centering
    \includegraphics[width=0.9\linewidth]{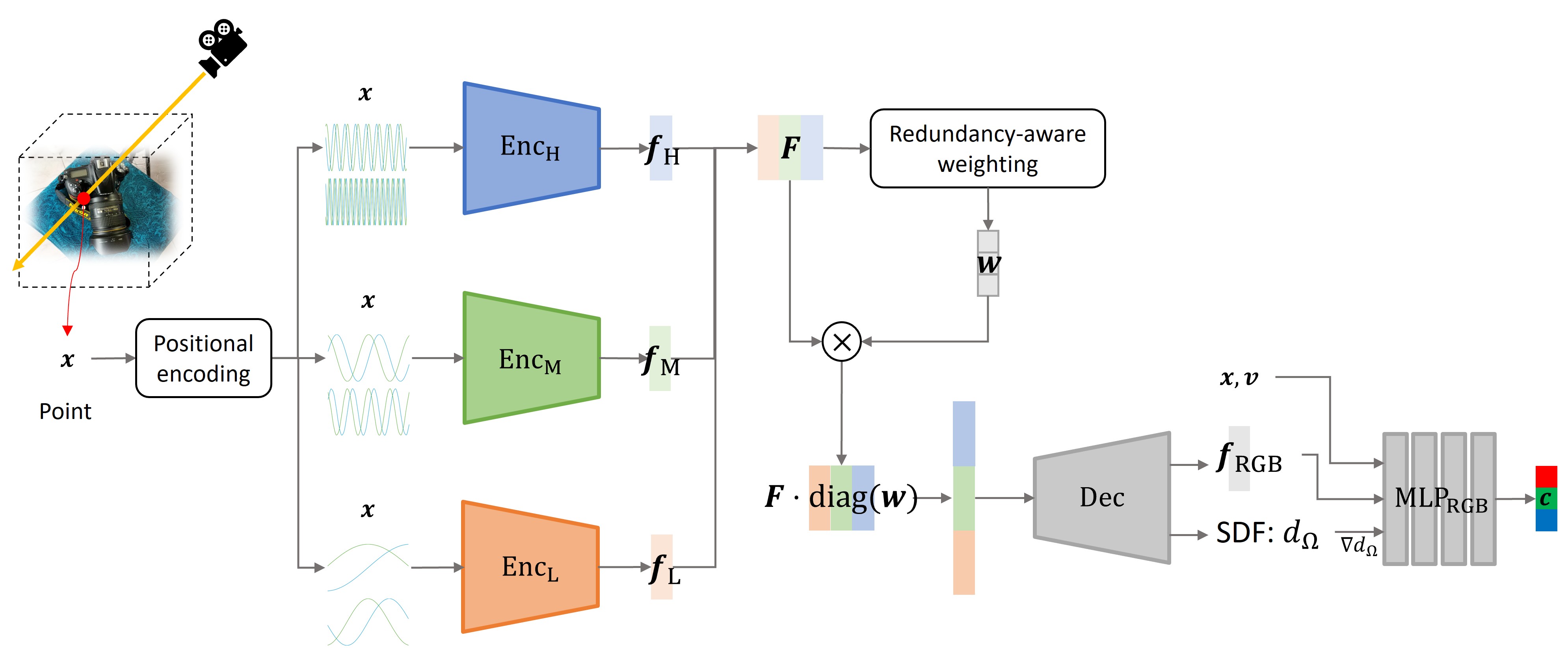}
    \caption{\method~ framework: Given an input 3D point $\boldsymbol{x}$, positional encoding maps it to the frequency domain. 
    The output of the positional encoding is then encoded into latent feature vectors corresponding to low--, middle--, and high--frequencies, respectively ($\boldsymbol{f}_{\rm{L}}, \boldsymbol{f}_{\rm{M}}, \boldsymbol{f}_{\rm{H}}$) by leveraging our frequency-stratified encoders $\rm{Enc_L, Enc_M}$, and $\rm{Enc_H}$.
    The \emph{redundancy-aware weighting} module takes the concatenated feature encodings ($\boldsymbol{F} = [\boldsymbol{f}_{\rm{L}}, \boldsymbol{f}_{\rm{M}}, \boldsymbol{f}_{\rm{H}}]$) and decides on the relative importance of these features according to the dissimilarity of each to the other two, estimating a normalized weight vector ($\boldsymbol{w}$).
    Finally, the weighted features ($\boldsymbol{F}\cdot\rm{diag}(\boldsymbol{w})$) are passed to a decoder $\rm{Dec}$ to extract a SDF value $d_{\Omega}$ and an appearance feature $\boldsymbol{f}_{\rm{RGB}}$ for the point $\boldsymbol{x}$. 
    $\rm{MLP}_{\rm{RGB}}$ predicts $\boldsymbol{x}$'s color given the appearance feature, point position $\boldsymbol{x}$, view direction $\boldsymbol{v}$, and point normal $\nabla d_{\Omega}$.
    }
    \label{fig:framework}
\end{figure*}

\subsection{Positional Encoding}
\label{sec:positionalencoding}
Positional encodings have assumed a critical role in neural implicit models, such as NeRF~\cite{mildenhall2020nerfrepresentingscenesneural} or VolSDF~\cite{yariv_volume_2021}. In these models, positional encoding is used to map the input coordinates into vectors in the frequency domain. 
Such a transformation injects ordering information into the input and enables 
the encoder network to capture the scene frequency information.
Eq.~\ref{equ:PE} shows a prototypical definition of the positional encoding, as used in neural implicit networks, such as VolSDF~\cite{yariv_volume_2021}.

\begin{multline}
\label{equ:PE}
    \gamma(\boldsymbol{x})=(\sin{(2^0\boldsymbol{x})}, \cos{(2^0\boldsymbol{x})},\cdots,\\\sin{(2^{\level~-1}\boldsymbol{x})},\cos{(2^{\level~-1}\boldsymbol{x})}),
\end{multline}
where $\boldsymbol{x} \in \mathbb{R}^3$ denotes the coordinate of the input point, 
while a total of $\level~$ frequencies are used for the encoding.

\subsection{Neural Volume Rendering}
\label{sec:neuralvolumerendering}
Neural volume rendering approaches, such as NeRF~\cite{mildenhall2020nerfrepresentingscenesneural}, have achieved tremendous success at the task of novel view rendering of 3D scenes. These models learn an implicit representation of the scene via a mapping from any 3D point $\boldsymbol{x}$ in the scene, 
encoded using positional encodings, to a volume density $\sigma(\boldsymbol{x}) \in [0, 1]$ and a RGB color $\boldsymbol{c}(\boldsymbol{x}) \in \mathbb{R}^3$, given a viewing direction $\boldsymbol{v}$. Such a mapping is typically implemented via a MLP network. The novel view rendering of the scene is generated pixel by pixel by casting a ray ($\boldsymbol{r}(t)=\boldsymbol{o}+t\boldsymbol{v}, t \geq 0, t \in \mathbb{R}$) emanating from the position of the camera center $\boldsymbol{o}$ in the viewing direction $\boldsymbol{v}$.

Using volume rendering, each pixel color $\hat{\boldsymbol{C}}_{\boldsymbol{p}}(\boldsymbol{r})$ at pixel $\boldsymbol{p}$
is calculated as the accumulation of all color contributions along the ray $\boldsymbol{r}$,
weighed by the accumulated transmittance $T(t)$ from the near bound $t_{\rm{near}}$ upto $t$, where
the transmittance is defined as: $T(t)=\exp(\int_{t_{\rm{near}}}^{t}\sigma(\boldsymbol{r}(s))ds)$
and opacity of the point  being captured by the density $\sigma(\boldsymbol{r}(t)) \in [0, 1]$). More formally, the pixel color $\hat{\boldsymbol{C}}_{\boldsymbol{p}}(\boldsymbol{r})$ is given by the following equation:
\begin{equation}
\label{equ:rendering}
    \hat{\boldsymbol{C}}_{\boldsymbol{p}}(\boldsymbol{r}) = \int_{t_{\rm{near}}}^{t_{\rm{far}}}T(t)\sigma(\boldsymbol{r}(t))\boldsymbol{c}(\boldsymbol{r}(t))dt,
\end{equation}
where $t_{\rm{near}}, t_{\rm{far}}$ denote the nearest and farthest points that could be sampled along the ray  $\boldsymbol{r}$.

\subsection{Signed Distance Function (SDF)}
Signed Distance Function (SDF) has recently emerged as a very effective tool for representing 3D surfaces~\cite{yariv_volume_2021,wang_neus_2021}. An SDF is a continuous function that denotes the distance of any point in 3D to the closest surface in the scene. The zero-level set of an SDF implicitly represents the scene's outer surface, points inside objects in the scene have a negative SDF value, while those that are outside have a positive SDF value. In practice, the SDF network is often instantiated by a MLP~\cite{yariv_volume_2021,wang_neus_2021}.
To train the SDF network without ground truth 3D mesh information,
prior works, such as VolSDF~\cite{yariv_volume_2021} and NeuS~\cite{wang_neus_2021}, convert SDF values into a density field and use it to synthesize RGB images from the viewing direction of the training views, via volume rendering. Such a design allows for the SDF model to derive a training signal by comparing ground truth RGB images with those estimated by the volume rendering step.

More concretely, given a scene $\Omega\subset\mathbb{R}^3$,
the volume density at a point $\boldsymbol{x}$ is derived from its SDF value $d_\Omega(\mathbf{x}) \in [-1, 1]$ (estimated from a neural network), using the following equation~\cite{yariv_volume_2021}:
\begin{multline}
\sigma(\boldsymbol{x})=
    \begin{cases}
        \frac{\alpha}{2}\rm{exp}(\frac{d_\Omega(\boldsymbol{x})}{\beta})\quad &\text{if $d_\Omega(\boldsymbol{x}) \leq 0$},\\
        \alpha\left(1-\frac{1}{2}\rm{exp}(\frac{-d_\Omega(\boldsymbol{x})}{\beta})\right)\quad &\text{if $d_\Omega(\boldsymbol{x})>0$},
    \end{cases}
\end{multline}
where $\alpha, \beta>0$ are learnable parameters.
Volume rendering can then be used to render a novel view image by using this volume density to weigh the color (RGB) value at the point $\boldsymbol{x}$, as estimated by a separate color prediction network. 

\section{Proposed Approach}
\label{sec:method}
In this section,
we introduce \method~, our novel approach for neural implicit surface representation.
\method~ reconstructs the 3D surface of a scene and can render it from any viewpoint, given a series of posed 2D images of the 3D scene. \method~ leverages our novel, frequency-stratified encoders to encode an input point in 3D space and decode it, using any (off-the-shelf~\cite{yariv_volume_2021}) decoder, to obtain the SDF value of the point as well as a feature, encoding its appearance. This appearance feature can then be decoded to obtain the view-dependent color, given the desired viewing direction.
Fig.~\ref{fig:framework} shows an overview of our proposed approach. 

\subsection{Frequency-domain Representation}
\label{sec:method_frequency}
Prior approaches for neural implicit surface representation struggle to simultaneously represent the correct shape of complex surfaces while capturing their fine details. This is primarily because they employ a single encoder network for the input point that attempts to capture all the various surface frequencies present in the scene (possibly from a very low to a very high one) simultaneously. This typically leads to a bias towards capturing the low--frequencies while ignoring the high-frequency details.
In our framework, we overcome this challenge by employing three encoders (low--frequency encoder ($\rm{Enc_L}$), middle--frequency encoder ($\rm{Enc_M}$), and high--frequency encoder ($\rm{Enc_H}$)) that convert the input to features corresponding to different frequency bands, instead of a single encoding,
to make the model more expressive and capable of representing surfaces with a wide variety of frequencies. 

To transform the spatial coordinates into the frequency domain,
we encode the input point using positional encodings (see Sec~\ref{sec:positionalencoding})  
and route it to the appropriate frequency encoder based on its associated frequency.
For instance, to distribute $6$ frequency levels (i.e., $\level~=6$) equally among the three encoders, we assign the lowest frequencies $\{2^0, 2^1\}$ to $\rm{Enc_{L}}$, the middle frequencies $\{2^2, 2^3\}$ to $\rm{Enc_M}$, while those for two highest frequencies $\{2^4, 2^5\}$ are routed to $\rm{Enc_H}$. 

Each encoder converts the positional encodings into corresponding $256$-D latent feature vectors ($\boldsymbol{f}_{\rm{L}}, \boldsymbol{f}_{\rm{M}}, \boldsymbol{f}_{\rm{H}}$). %
Such stratification of the frequency representation bolsters the model's capability to model the shape of the surface of the scene while capturing its details.

\subsection{Redundancy-aware Weighting}
\label{sec:method_attention}
\begin{figure*}
    \centering
    \includegraphics[width=0.9\linewidth]{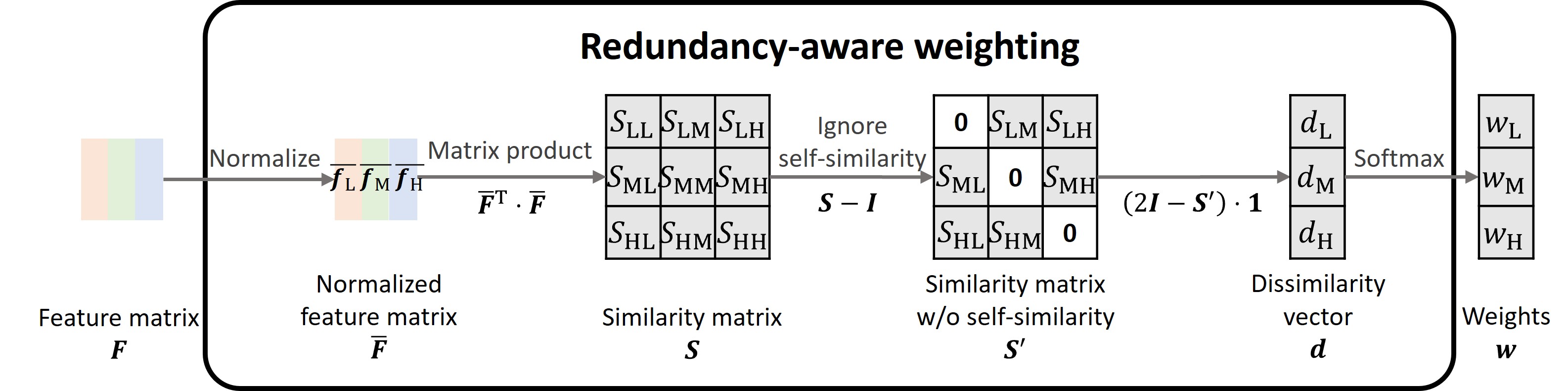}
    \caption{Redundancy-aware weighting module: 
    The redundancy-aware weighting module takes the encoded frequency features and predicts a normalized importance score, following the pipeline shown in the figure, assigning a higher weight to the frequency encoding that is least similar to the other two and vice-versa.   %
    }
    \label{fig:attentionbasedweighting}
\end{figure*}
For the encoder capacity to be maximally utilized, encouraging dissimilarity between the learned representations of the three encoders is essential. To promote such behavior and effectively combine the complementary information learned by the different encoders, we propose a novel, \emph{redundancy-aware weighting} module, as shown in Fig.~\ref{fig:attentionbasedweighting}. This module estimates normalized importance scores for each of the three different feature encodings and uses them as weights to combine the encodings to derive a unified representation. 
A higher score is assigned to the feature encoding which is the most dissimilar 
to the other two and vice-versa, promoting the learning of complementary feature encodings between the encoders.

At the outset, the module concatenates features from the three encoders into a matrix, which we denote as $\boldsymbol{F} = [\boldsymbol{f}_{\rm{L}}, \boldsymbol{f}_{\rm{M}}, \boldsymbol{f}_{\rm{H}}] \in \mathbb{R}^{256 \times 3}$, which is then normalized per column,
based on the $L_2$ norm, denoted by $\boldsymbol{\bar{F}}$.
Next, a similarity matrix $\boldsymbol{S}$ is computed by taking the matrix product of $\boldsymbol{\bar{F}}^T$ and $\boldsymbol{\bar{F}}$, %
as shown in Eq.~\ref{equ:similaritymatrix}.
\begin{equation}
\label{equ:similaritymatrix}
    \boldsymbol{S} =\boldsymbol{\bar{F}}^T\cdot\boldsymbol{\bar{F}}=
    \begin{pmatrix}
        S_{\rm{LL}} & S_{\rm{LM}} & S_{\rm{LH}}\\
        S_{\rm{ML}} & S_{\rm{MM}} & S_{\rm{MH}}\\
        S_{\rm{HL}} & S_{\rm{HM}} & S_{\rm{HH}}
    \end{pmatrix},
\end{equation}
where each entry, $\in [-1,1]$.
To compute the dissimilarity information from $\boldsymbol{S}$,
we remove the diagonal entries, which capture the self-similarity, as shown: %
\begin{equation}
    \boldsymbol{S}^\prime = \boldsymbol{S} - \boldsymbol{I}=    
    \begin{pmatrix}
        0 & S_{\rm{LM}} & S_{\rm{LH}}\\
        S_{\rm{ML}} & 0 & S_{\rm{MH}}\\
        S_{\rm{HL}} & S_{\rm{HM}} & 0
    \end{pmatrix},
\end{equation}
where $\boldsymbol{I}$ denotes the $3\times3$ identity matrix.
Next, a dissimilarity vector $\boldsymbol{d}$ is computed, as shown in Eq.~\ref{equ:dissimilarity}:
\begin{equation}
\label{equ:dissimilarity}
    \boldsymbol{d} = (2\boldsymbol{I} - \boldsymbol{S}^\prime)\cdot\boldsymbol{1},
\end{equation}
where $\boldsymbol{1}$ is $[1,1,1]^T$.
Finally, the weight vector $\boldsymbol{w}$ for $\boldsymbol{F}$ is given by Eq.~\ref{eq:weight}.

\begin{equation}
\label{eq:weight}
    \boldsymbol{w} = \rm{Softmax}\left(\frac{\boldsymbol{d}}{\tau}\right),
\end{equation}
where the $\rm{Softmax}(\cdot)$ function rescales elements in a vector to be in the range $[0, 1]$ and sum to $1$,
and $\tau$ is a temperature parameter that controls the smoothness of the softmax distribution.
The default value of $\tau$ is set to $0.5$.
The redundancy-aware encoder features are then computed by: $\boldsymbol{F}\cdot\rm{diag}(\boldsymbol{w})$. %

\subsection{Decoder}
The redundancy-weighted encoder features can be decoded to obtain the SDF value of the point and its appearance feature. This is undertaken via a decoder ($\rm{Dec}$), often instantiated by a MLP network, which takes the flattened redundancy-weighted feature vector as an input and estimates the SDF value and an appearance feature vector ($\boldsymbol{f}_{\rm{RGB}}$) as an output. $\boldsymbol{f}_{\rm{RGB}}$ is then used to derive the view-dependent RGB color for the point $\boldsymbol{x}$.
The final RGB-color value of the point is obtained by feeding $\boldsymbol{f}_{RGB}$, the point coordinates, and the viewing direction
to the color prediction network $\rm{MLP_{RGB}}$, akin to volume rendering methods discussed in Sec.~\ref{sec:neuralvolumerendering}.

\subsection{Loss Function}
\label{sec:lossfunction}
We train \method~ using the following set of  losses: %
(i) the photometric loss $\mathcal{L}_{\rm{RGB}}$ and (ii) the Eikonal loss $\mathcal{L}_{\rm{Eikonal}}$. The final loss is given by:
\begin{equation}
\label{equ:trainingloss}
    \mathcal{L} = \mathcal{L}_{\rm{RGB}} + \lambda\mathcal{L}_{\rm{Eikonal}},
\end{equation}
where $\lambda \in \mathbb{R}, \lambda > 0$. $\mathcal{L}_{\rm{RGB}}$ and $\mathcal{L}_{\rm{Eikonal}}$ in Eq.~\ref{equ:trainingloss} are defined as follows:
\begin{gather}
    \mathcal{L}_{\rm{RGB}} = ||\boldsymbol{C}_{\boldsymbol{p}} - \hat{\boldsymbol{C}}_{\boldsymbol{p}}(\boldsymbol{r})||_1,\label{eq:rgbloss}\\
    \mathcal{L}_{\rm{Eikonal}} = (||\nabla d_\Omega(\boldsymbol{z})||-1)^2.
    \label{equ:eikonalloss}
\end{gather}
In Eq.~\ref{eq:rgbloss}, $\boldsymbol{C}_{\boldsymbol{p}}$ is the ground truth color at pixel $\boldsymbol{p}$, 
and $\hat{\boldsymbol{C}}_{\boldsymbol{p}}(\boldsymbol{r})$ is the rendered color (obtained using Eq.~\ref{equ:rendering}).
In Eq.~\ref{equ:eikonalloss}, $d_\Omega(\boldsymbol{z})$  is an approximated SDF value for the sampled point $\boldsymbol{z}$.

\begin{table*}[t]
    \centering
    \resizebox{\textwidth}{!}{%
    \begin{tabular}{llccccccccc|c}
        \toprule
        & Method (no. of parameters) & Doll & Egg & Head & Angel & Bull & Robot & Dog & Bread & Camera & \textbf{Mean} \\
        \midrule
        \multirow{3}{3em}{PSNR($\uparrow$)} & VolSDF~\cite{yariv_volume_2021} (0.5M) & 25.43 & 27.23 & 26.94 & 30.28 & 26.18 & 26.39 & 28.44 & \textbf{31.18} & 22.96 & 27.23\\
        & Scaled-up VolSDF (1.4M)  & 26.07 & 27.15 & 26.62 & 30.37 & 26.08 & 25.07 & 28.32 & 29.44 & 23.02 & 26.90 \\
        & Ours (1.4M) & \textbf{26.22} & \textbf{27.48} & \textbf{27.29} & \textbf{30.52} & \textbf{26.33} & \textbf{26.69} & \textbf{28.56} & 30.22 & \textbf{23.08} & \textbf{27.38}\\
        \midrule
        \multirow{3}{3em}{SSIM($\uparrow$)} & VolSDF~\cite{yariv_volume_2021} (0.5M) & 0.911 & 0.943 & 0.959 & 0.989 & 0.970 & 0.957 & 0.950 & \textbf{0.988} & 0.928 & 0.955\\
        & Scaled-up VolSDF (1.4M) & 0.925 & 0.943 & 0.956 & \textbf{0.990} & 0.970 & 0.946 & 0.949 & 0.980 & 0.929 & 0.954\\
        & Ours (1.4M) & \textbf{0.928} & \textbf{0.946} & \textbf{0.961} & \textbf{0.990} & \textbf{0.971} & \textbf{0.962} & \textbf{0.952} & 0.983 & \textbf{0.930} & \textbf{0.958} \\ 
        \midrule
        \multirow{3}{3em}{LPIPS($\downarrow$)} & VolSDF~\cite{yariv_volume_2021} (0.5M) & 0.041 & 0.032 & 0.017 & 0.007 & 0.021 & 0.032 & 0.027 & \textbf{0.006} & 0.045 & 0.025 \\
        & Scaled-up VolSDF (1.4M) & \textbf{0.035} & 0.032  & 0.018 & \textbf{0.006} & 0.021 & 0.043 & 0.028 & 0.011 & 0.045 & 0.027\\
        & Ours (1.4M) & \textbf{0.035} &  \textbf{0.030} & \textbf{0.015} & \textbf{0.006} & \textbf{0.020} & \textbf{0.030} & \textbf{0.026} & 0.009 & \textbf{0.044} & \textbf{0.024} \\ 
        \bottomrule
    \end{tabular}
    }
    \caption{Quantitative results for 9 scenes from the BlendedMVS dataset. 
    The best score in each scene is shown in \textbf{bold}}.
    \vspace{-0.05in}
    \label{tab:BMVS}
\end{table*}

\section{Experiments}
\label{sec:experiments}
\begin{figure*}[t]
    \centering
    \resizebox{0.75\textwidth}{!}{%
    \begin{minipage}[c]{0.04\textwidth}
        \centering
        \rotatebox[origin=c]{90}{\textbf{Doll}}
    \end{minipage}
    \hfill
    \begin{subfigure}[c]{0.225\textwidth}
        \includegraphics[width=\columnwidth]{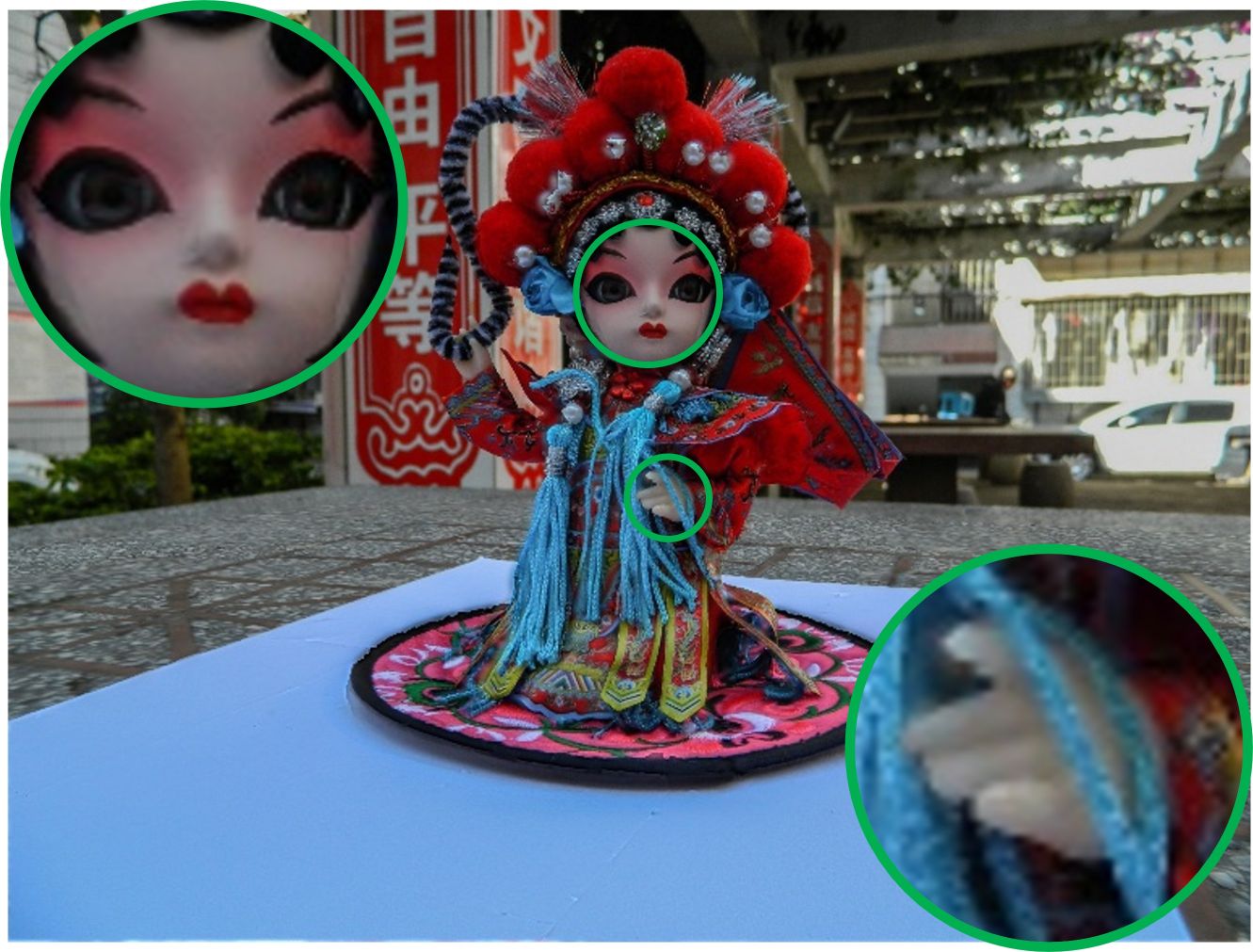}
    \end{subfigure}
    \begin{subfigure}[c]{0.225\textwidth}
        \includegraphics[width=\columnwidth]{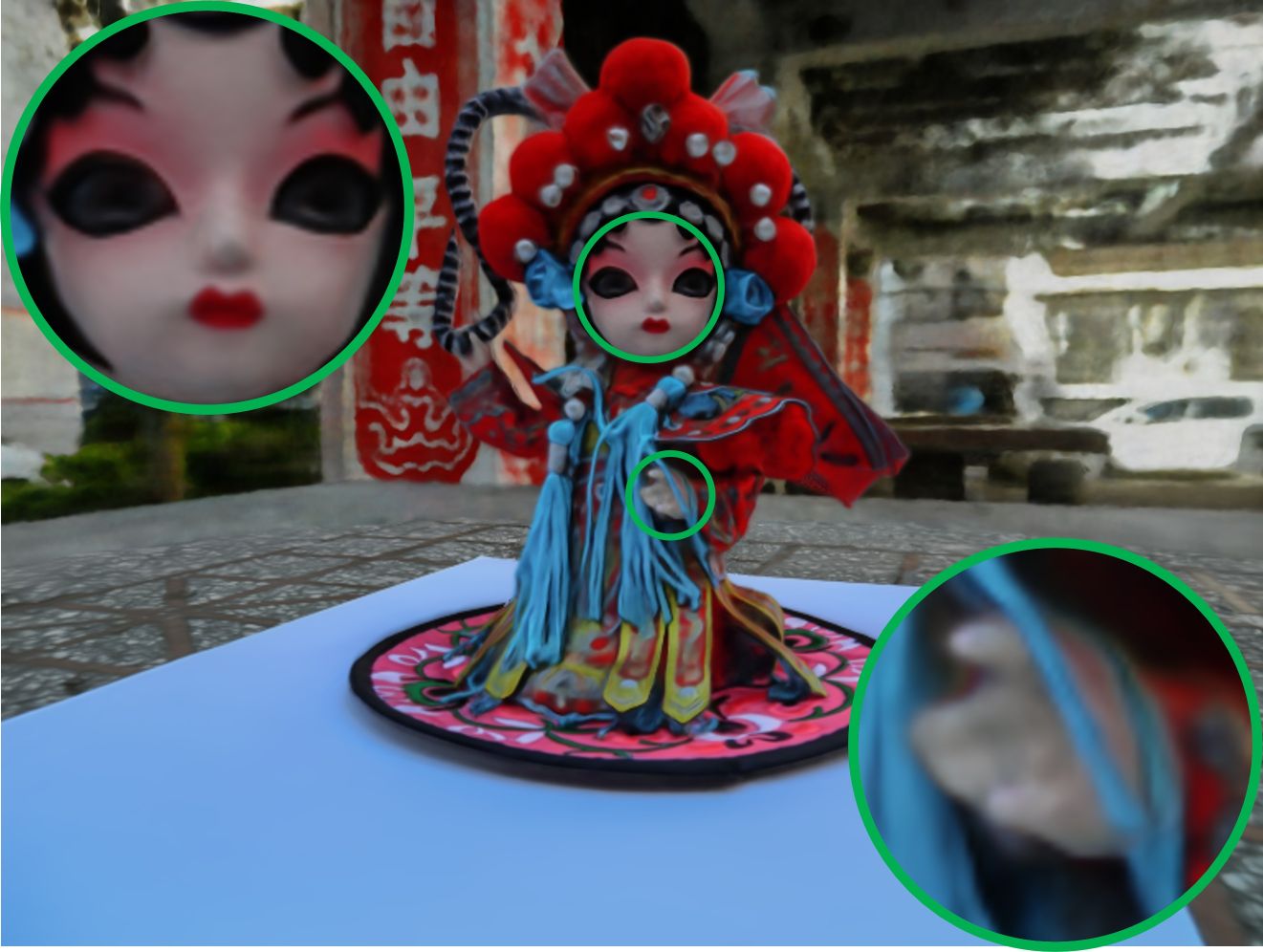}
    \end{subfigure}
    \begin{subfigure}[c]{0.225\textwidth}
        \includegraphics[width=\columnwidth]{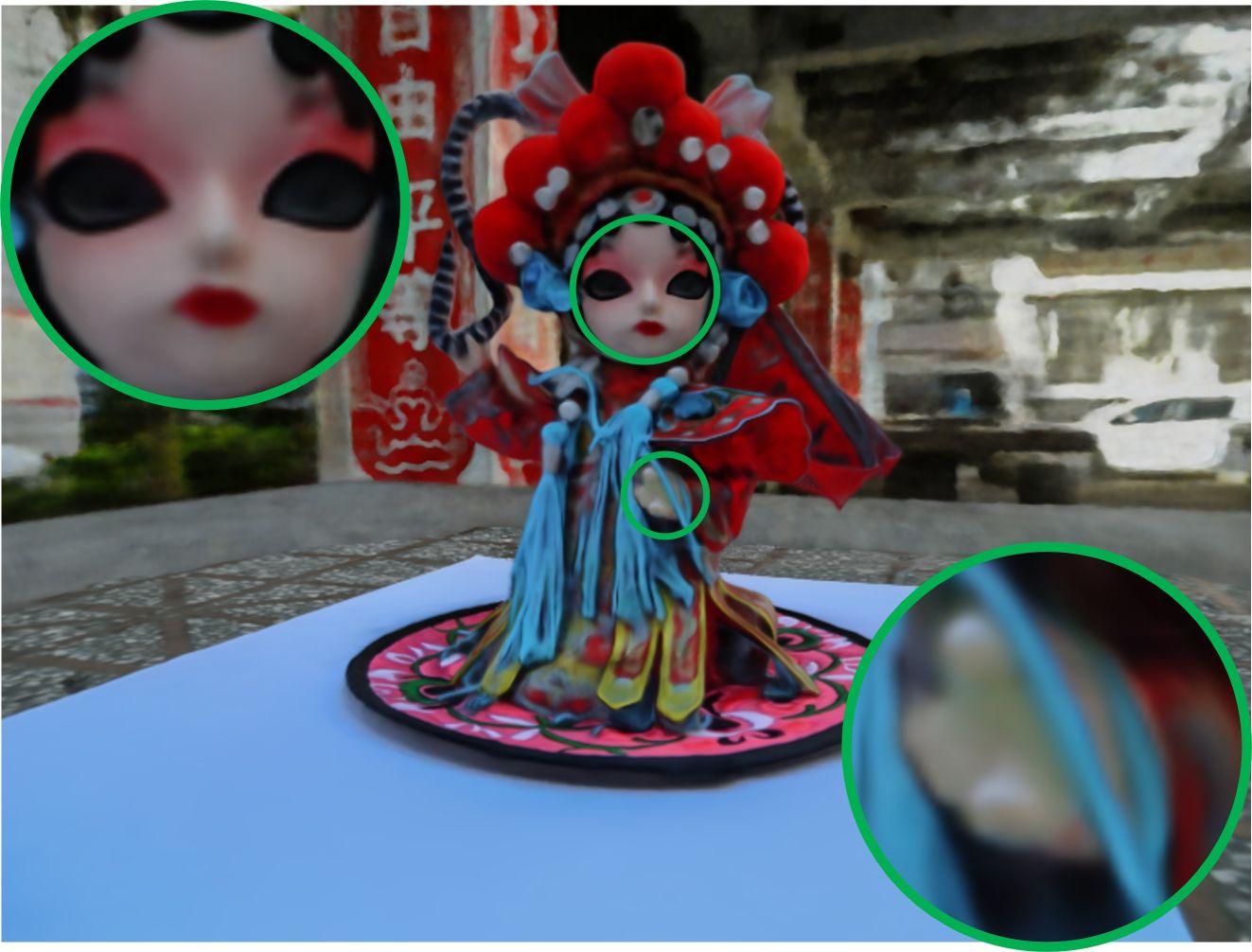}
    \end{subfigure}
    \begin{subfigure}[c]{0.225\textwidth}
        \includegraphics[width=\columnwidth]{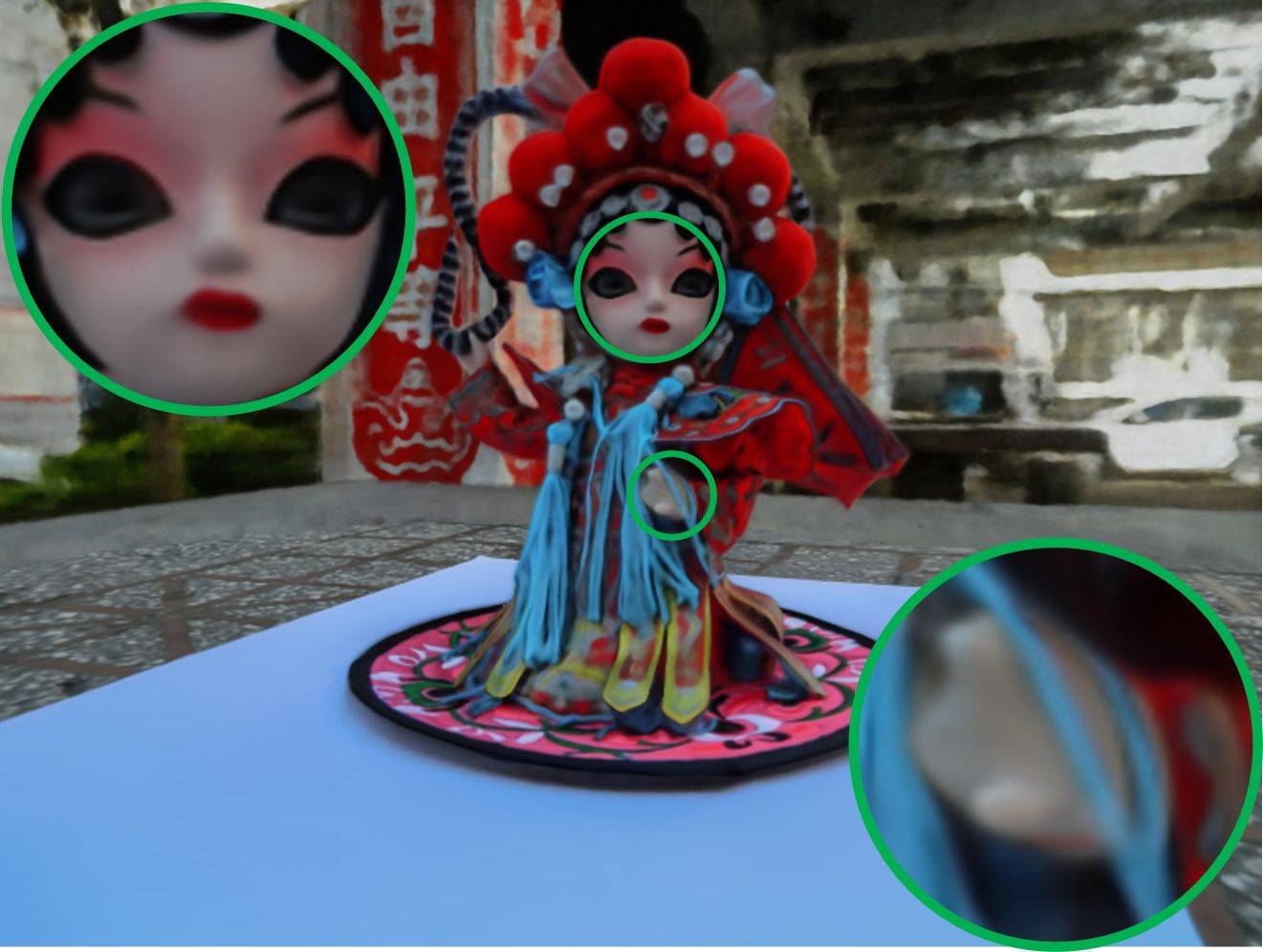}
    \end{subfigure}
    }
    \resizebox{0.75\textwidth}{!}{%
    \begin{minipage}[c]{0.04\textwidth}
        \centering
        \rotatebox[origin=c]{90}{\textbf{Bull}}
    \end{minipage}
    \hfill
    \begin{subfigure}[c]{0.225\textwidth}
        \includegraphics[width=\columnwidth]{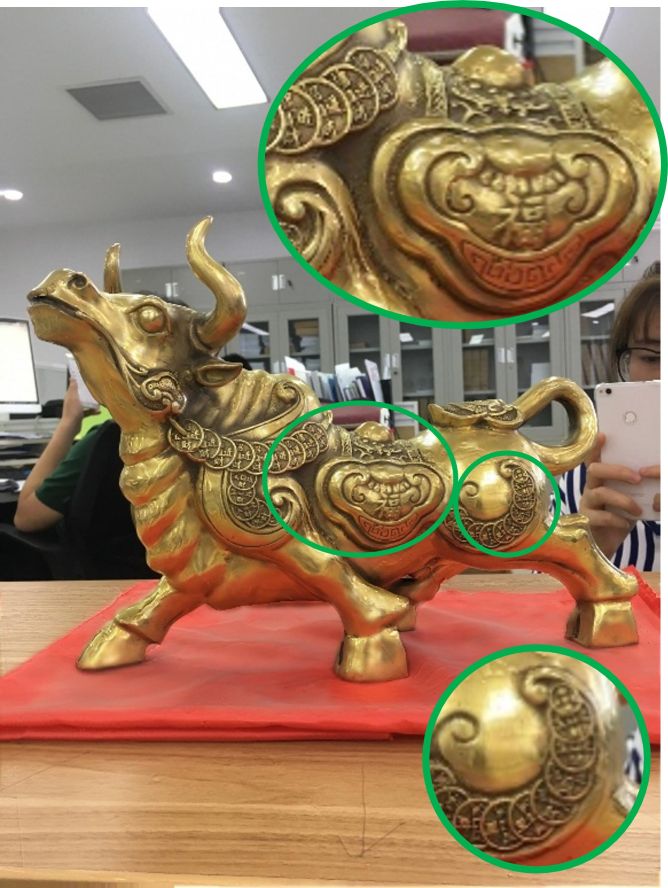}
    \end{subfigure}
    \begin{subfigure}[c]{0.225\textwidth}
        \includegraphics[width=\columnwidth]{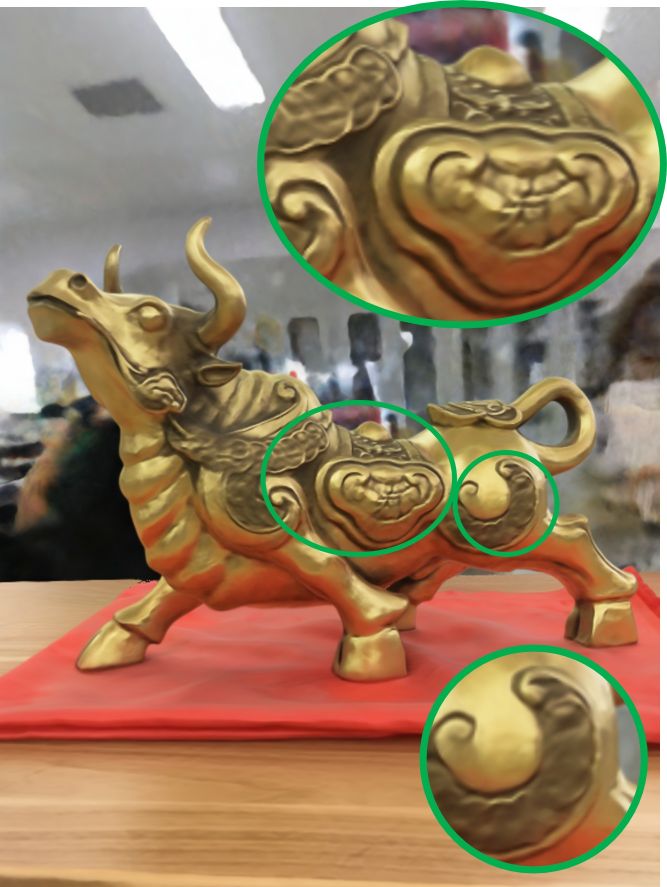}
    \end{subfigure}
    \begin{subfigure}[c]{0.225\textwidth}
        \includegraphics[width=\columnwidth]{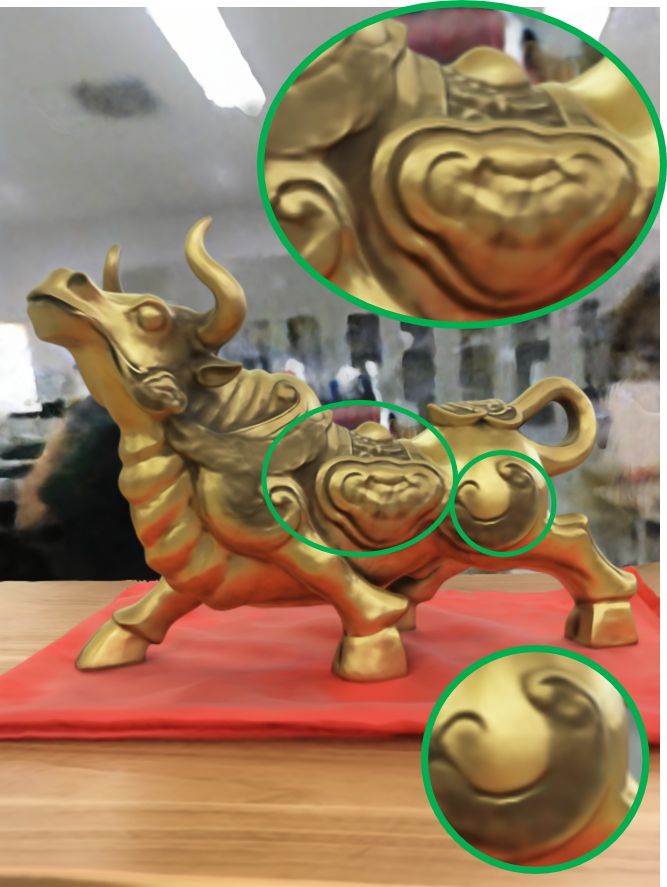}
    \end{subfigure}
    \begin{subfigure}[c]{0.225\textwidth}
        \includegraphics[width=\columnwidth]{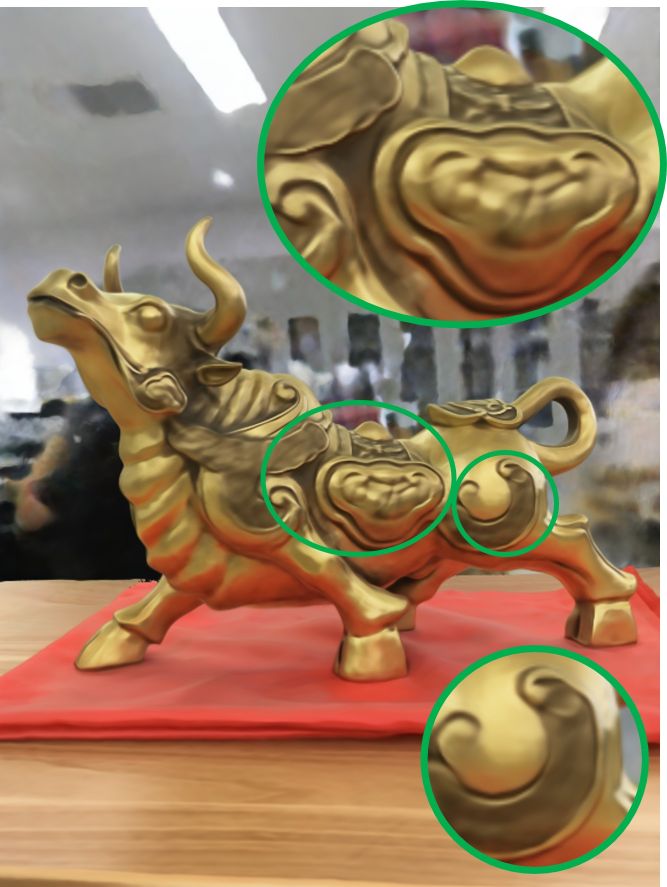}
    \end{subfigure}
    }
    \resizebox{0.75\textwidth}{!}{%
    \begin{minipage}[c]{0.04\textwidth}
        \centering
        \rotatebox[origin=c]{90}{\textbf{Robot}}
    \end{minipage}
    \hfill
    \begin{subfigure}[c]{0.225\textwidth}
        \includegraphics[width=\columnwidth]{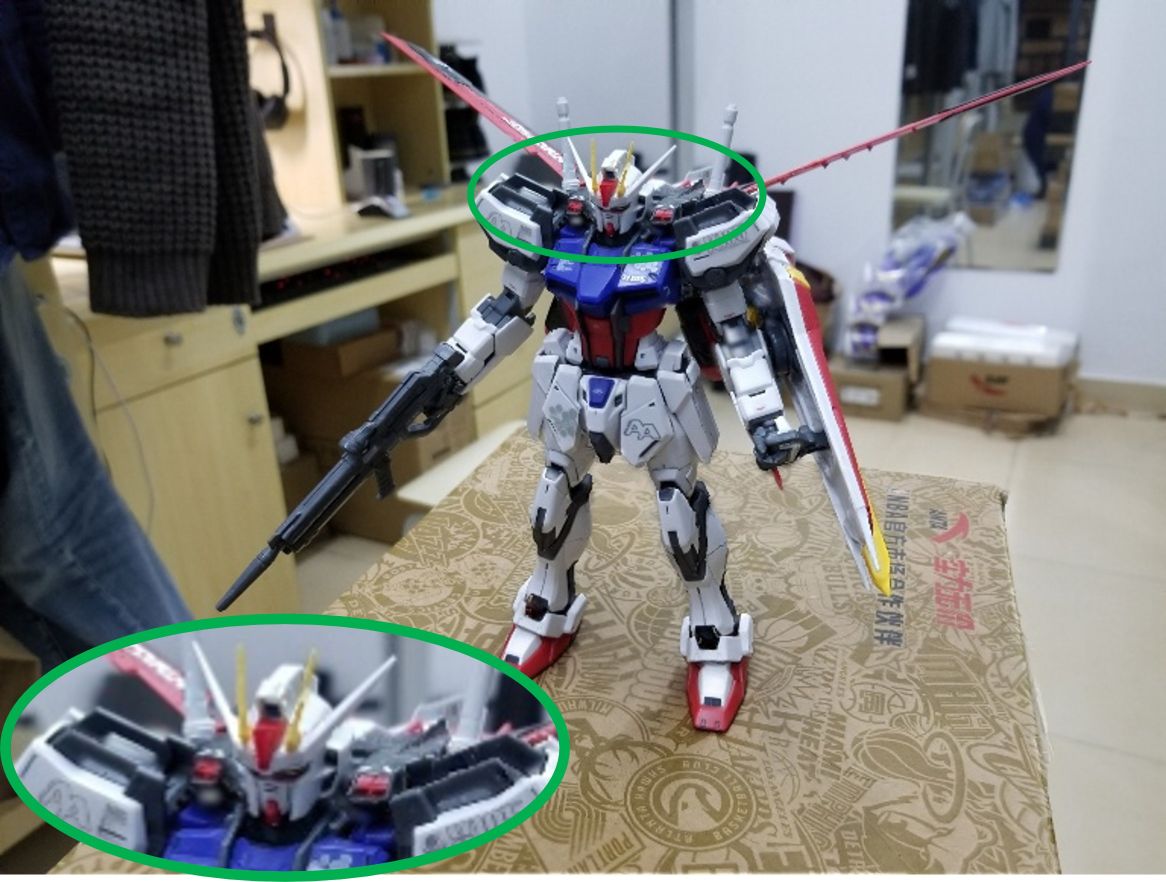}
        \subcaption{Reference image}
        \label{fig:RGB_gt}
    \end{subfigure}
    \begin{subfigure}[c]{0.225\textwidth}
        \includegraphics[width=\columnwidth]{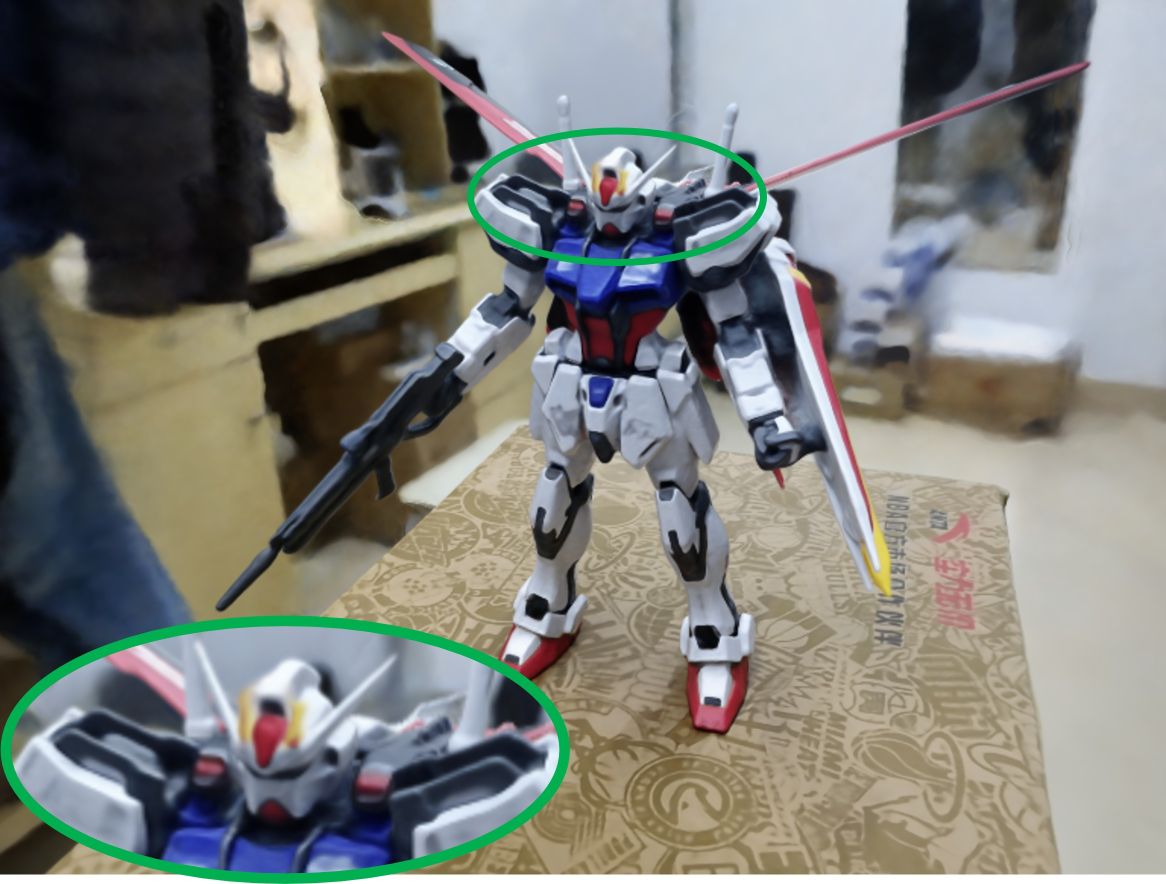}
        \caption{Ours}
        \label{fig:RGB_ours}
    \end{subfigure}
    \begin{subfigure}[c]{0.225\textwidth}
        \includegraphics[width=\columnwidth]{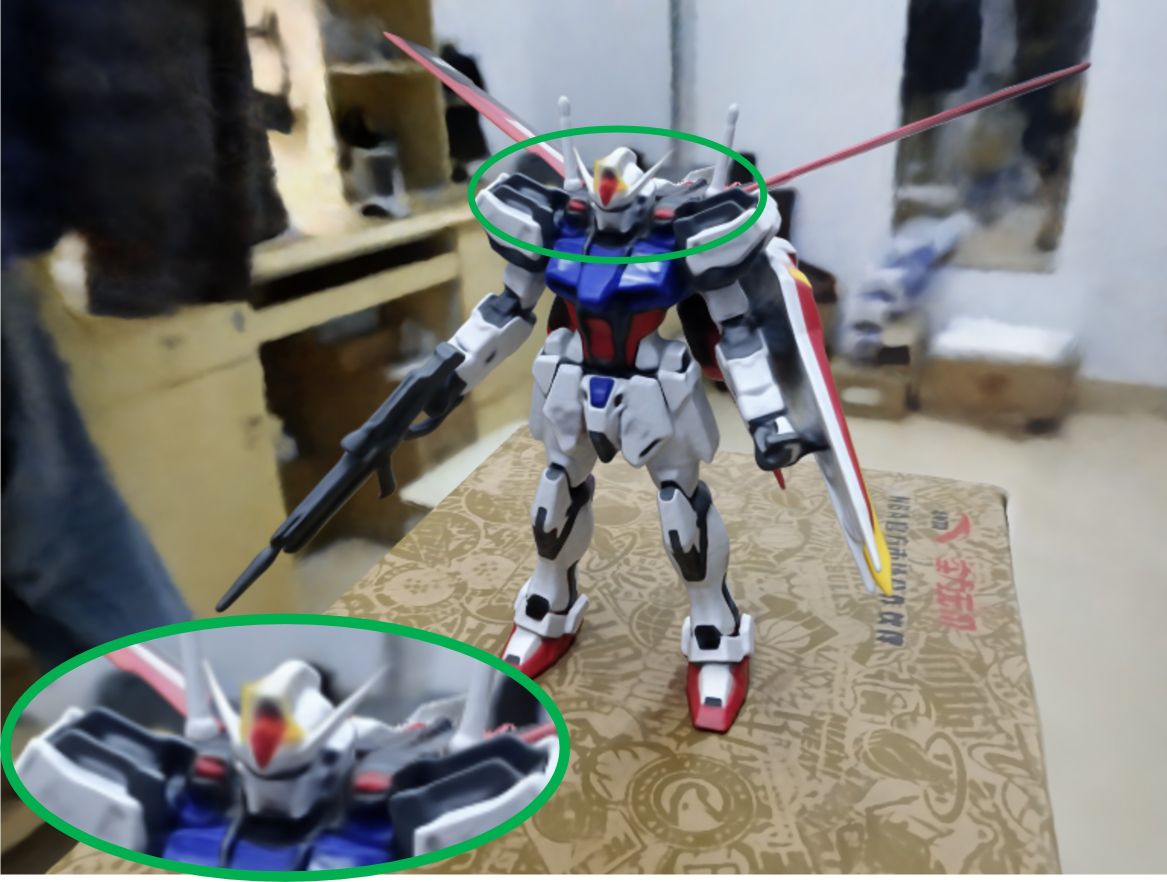}
        \caption{VolSDF}
        \label{fig:RGB_vanilla}
    \end{subfigure}
    \begin{subfigure}[c]{0.225\textwidth}
        \includegraphics[width=\columnwidth]{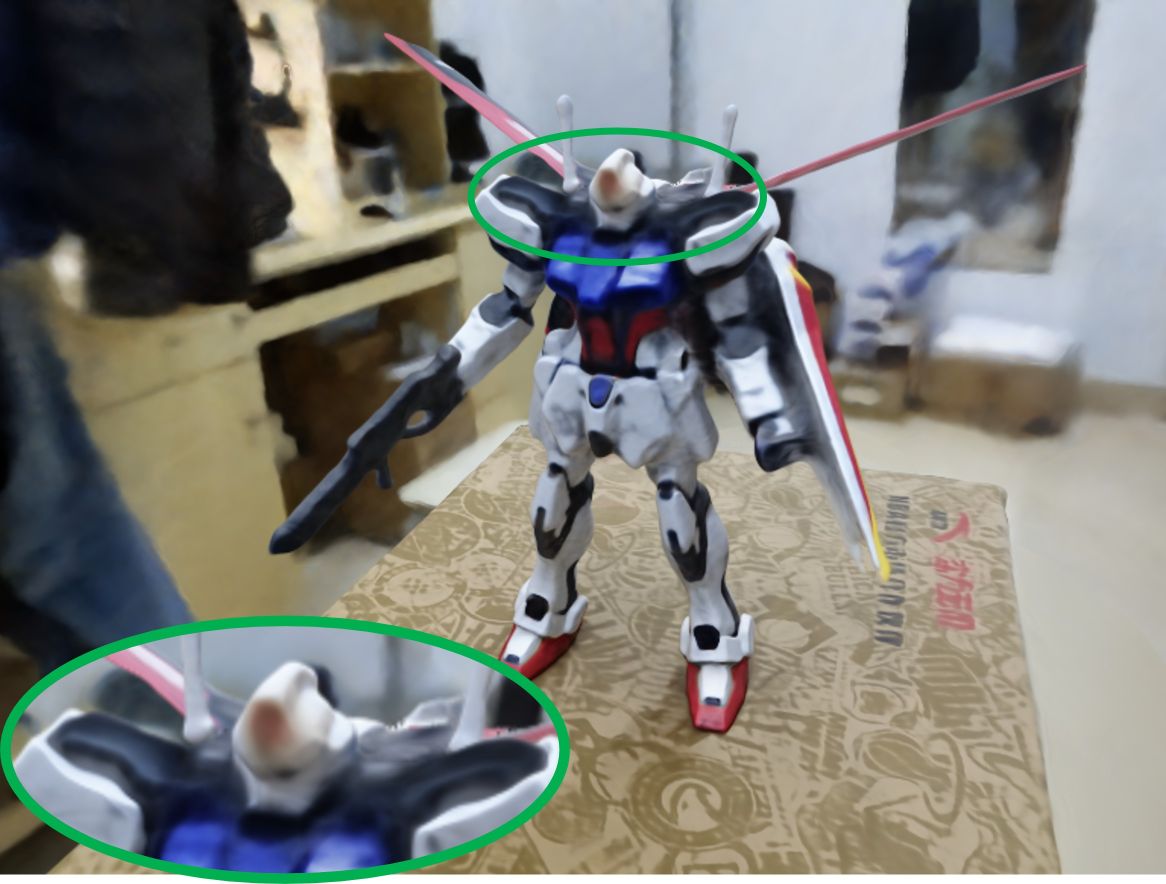}
        \caption{Scaled-up VolSDF}
        \label{fig:RGB_scaleup}
    \end{subfigure}
    }
    \caption{Qualitative comparison of viewpoint-based scene rendering on the BlendedMVS dataset. } %
    \label{fig:RGBresults}
\end{figure*}

\begin{figure*}[t]
    \centering
    \resizebox{0.8\textwidth}{!}{%
    \begin{minipage}[c]{0.07\textwidth}
        \centering
        \rotatebox[origin=c]{90}{\textbf{Doll}}
    \end{minipage}
    \hfill
    \begin{subfigure}[c]{0.225\textwidth}
        \includegraphics[width=\columnwidth]{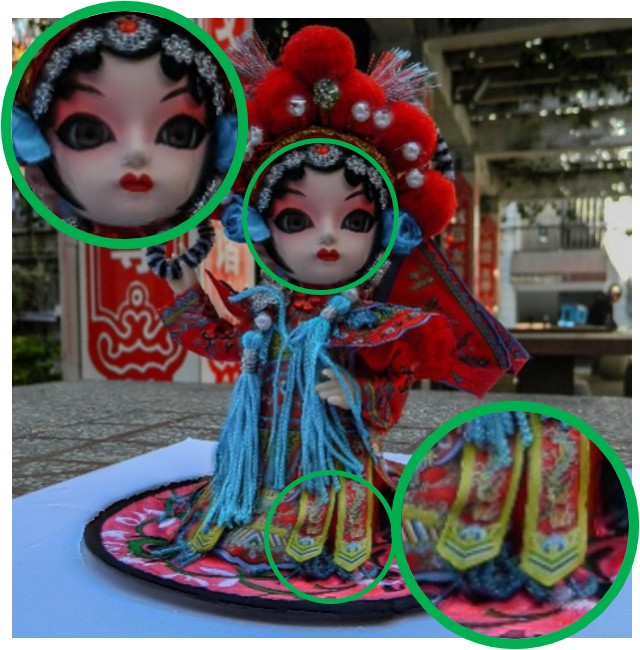}
    \end{subfigure}
    \begin{subfigure}[c]{0.225\textwidth}
        \includegraphics[width=\columnwidth]{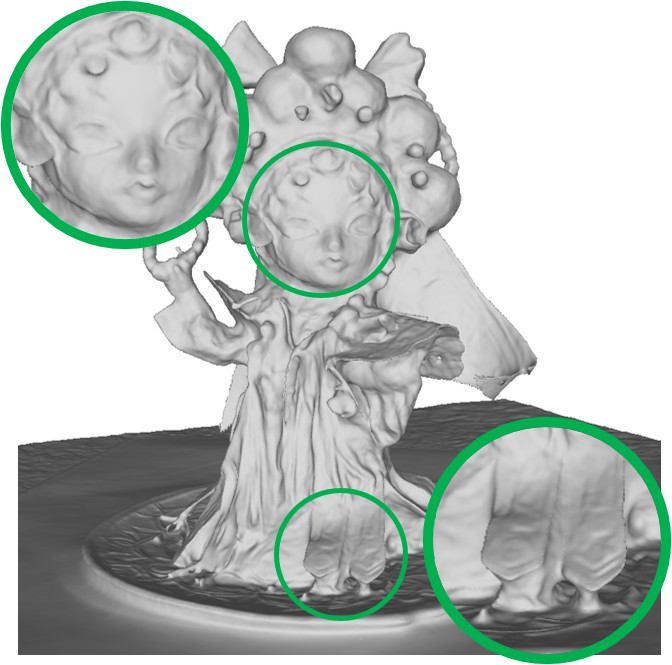}
    \end{subfigure}
    \begin{subfigure}[c]{0.225\textwidth}
        \includegraphics[width=\columnwidth]{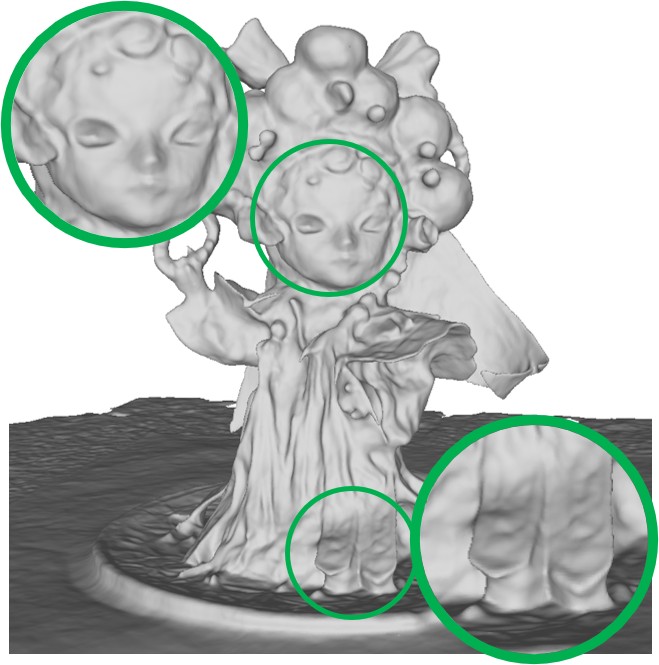}
    \end{subfigure}
    \begin{subfigure}[c]{0.225\textwidth}
        \includegraphics[width=\columnwidth]{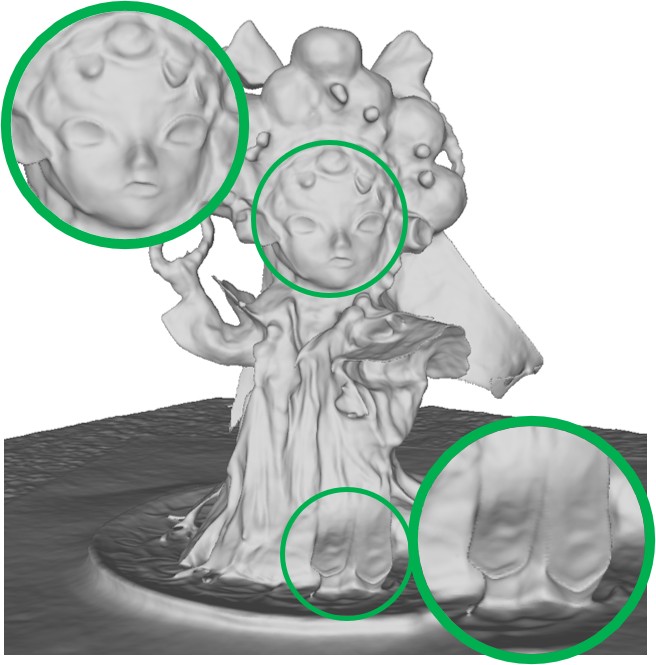}
    \end{subfigure}
    }
    \resizebox{0.8\textwidth}{!}{%
    \begin{minipage}[c]{0.07\textwidth}
        \centering
        \rotatebox[origin=c]{90}{\textbf{Bull}}
    \end{minipage}
    \hfill
    \begin{subfigure}[c]{0.225\textwidth}
        \includegraphics[width=\columnwidth]{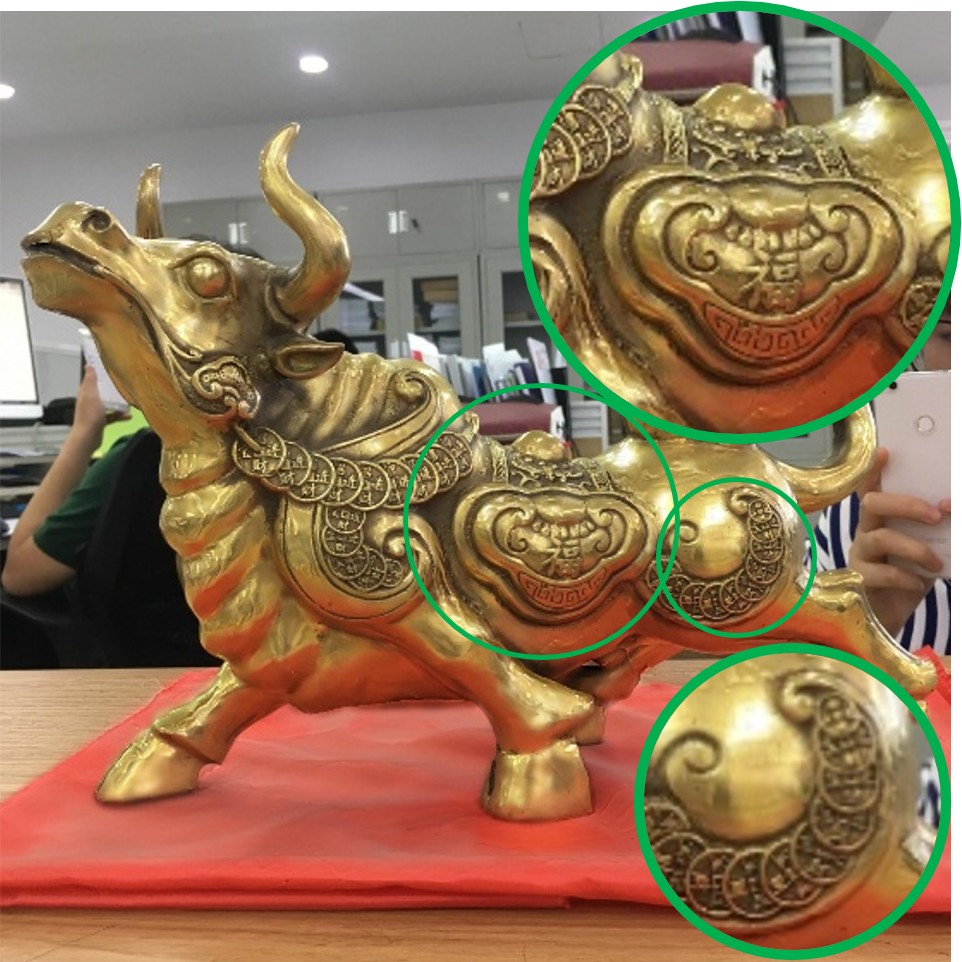}
    \end{subfigure}
    \begin{subfigure}[c]{0.225\textwidth}
        \includegraphics[width=\columnwidth]{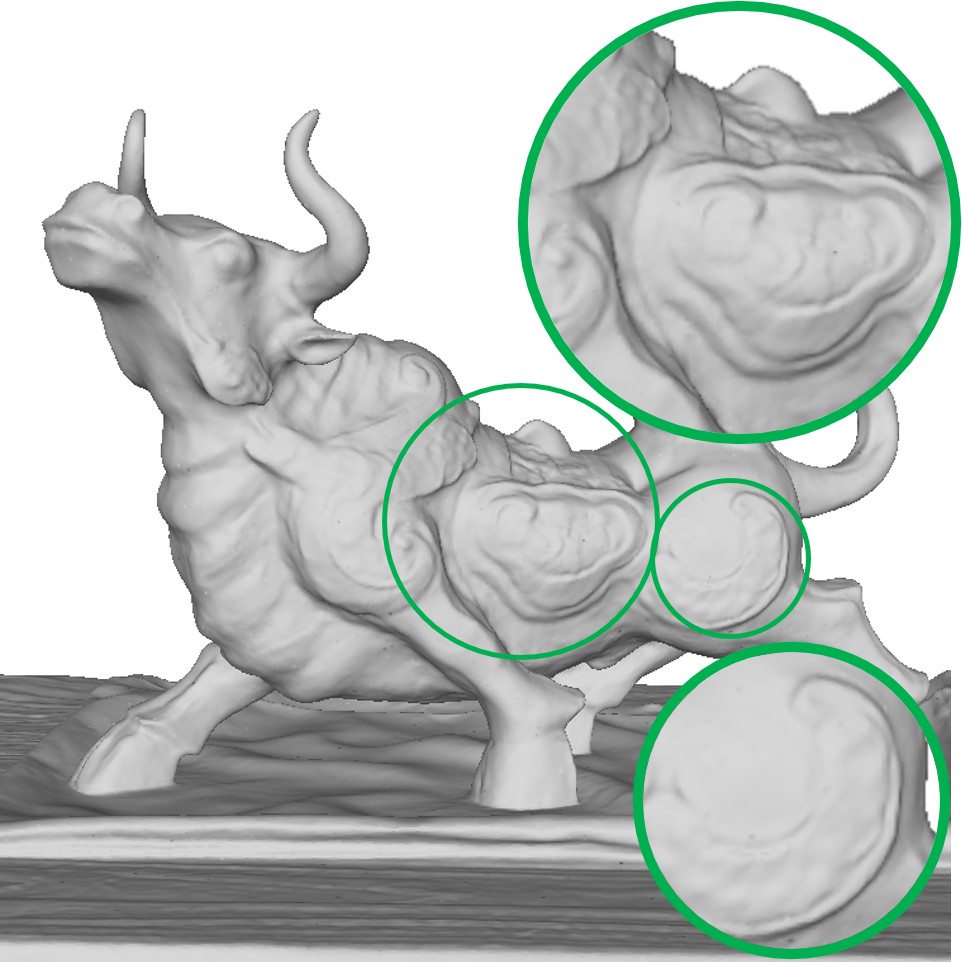}
    \end{subfigure}
    \begin{subfigure}[c]{0.225\textwidth}
        \includegraphics[width=\columnwidth]{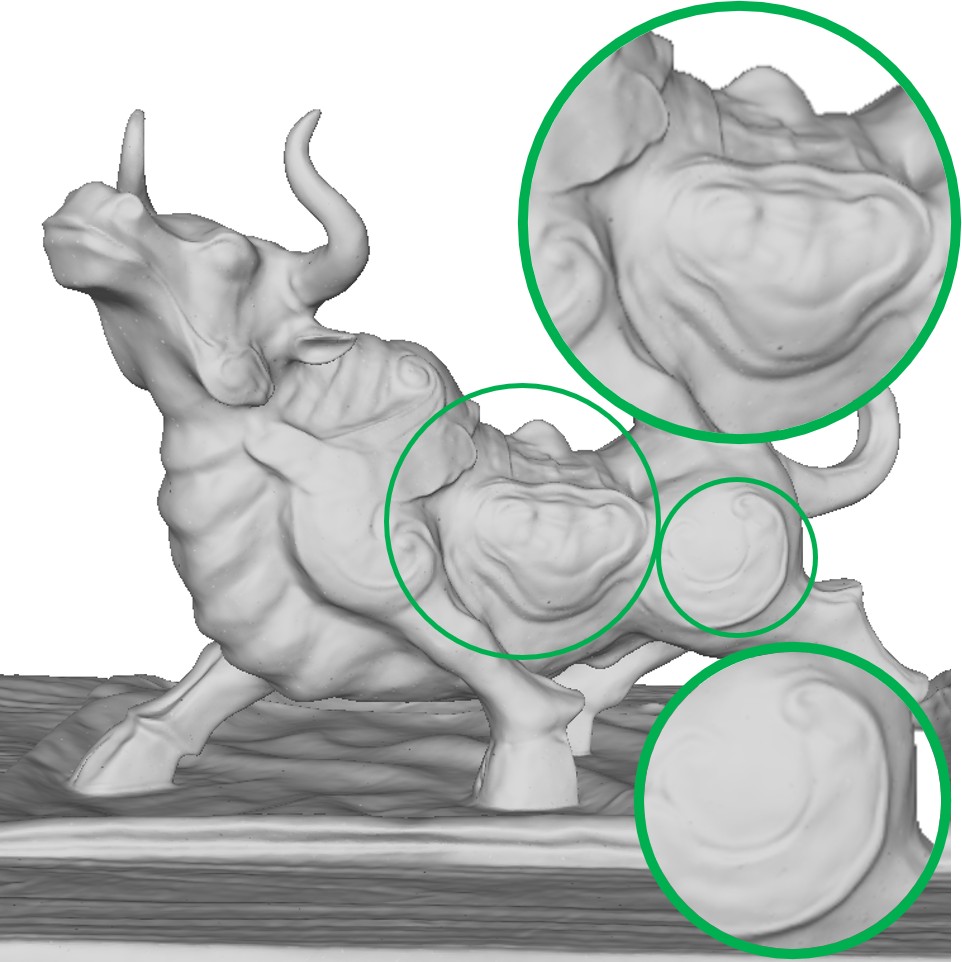}
    \end{subfigure}
    \begin{subfigure}[c]{0.225\textwidth}
        \includegraphics[width=\columnwidth]{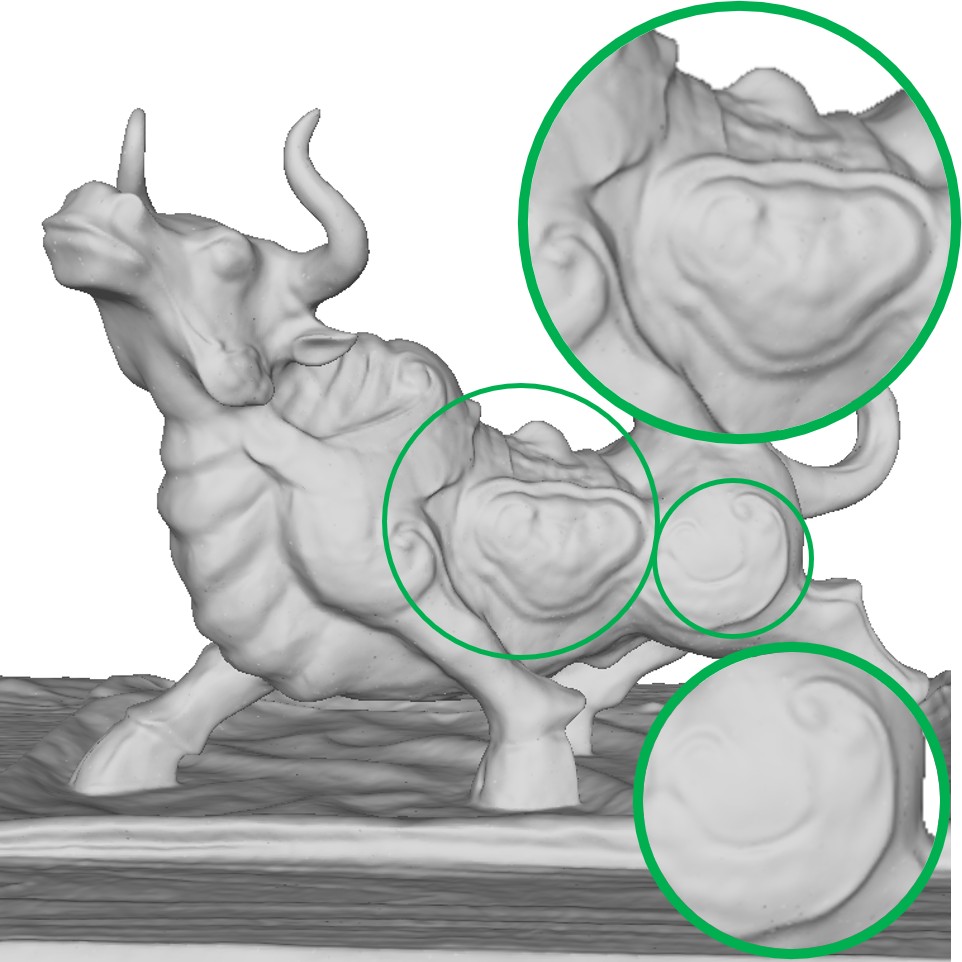}
    \end{subfigure}
    }
    \resizebox{0.8\textwidth}{!}{%
    \begin{minipage}[c]{0.07\textwidth}
        \centering
        \rotatebox[origin=c]{90}{\textbf{Robot}}
    \end{minipage}
    \hfill
    \begin{subfigure}[c]{0.225\textwidth}
        \includegraphics[width=\columnwidth]{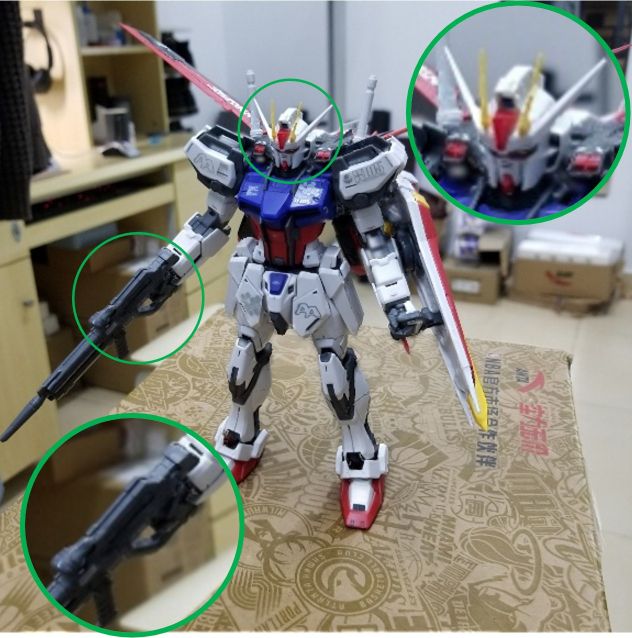}
        \subcaption{Reference image}
        \label{fig:gt_annotation}
    \end{subfigure}
    \begin{subfigure}[c]{0.225\textwidth}
        \includegraphics[width=\columnwidth]{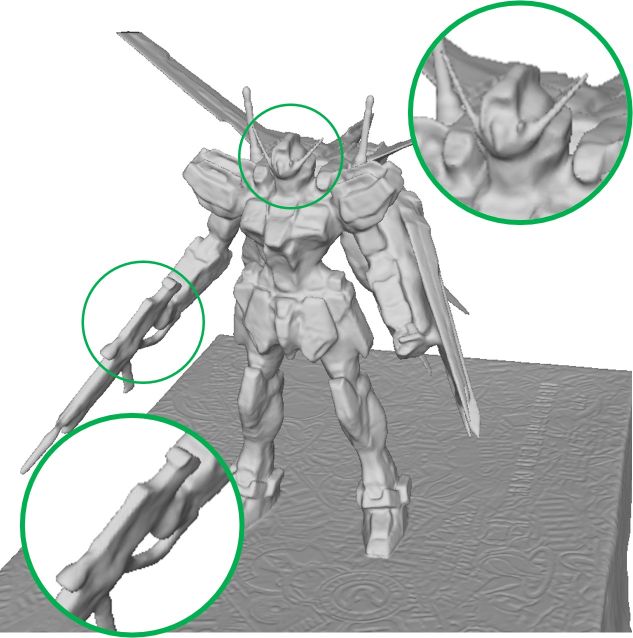}
        \caption{Ours}
        \label{fig:mesh_ours}
    \end{subfigure}
    \begin{subfigure}[c]{0.225\textwidth}
        \includegraphics[width=\columnwidth]{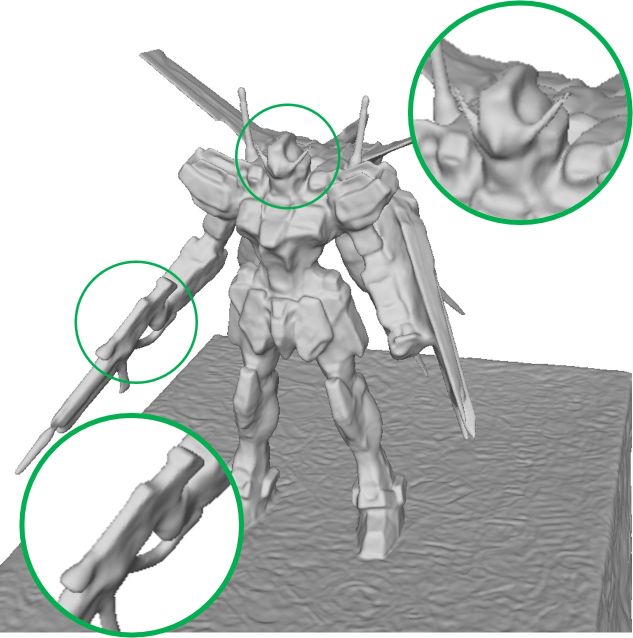}
        \caption{VolSDF}
        \label{fig:mesh_vanilla}
    \end{subfigure}
    \begin{subfigure}[c]{0.225\textwidth}
        \includegraphics[width=\columnwidth]{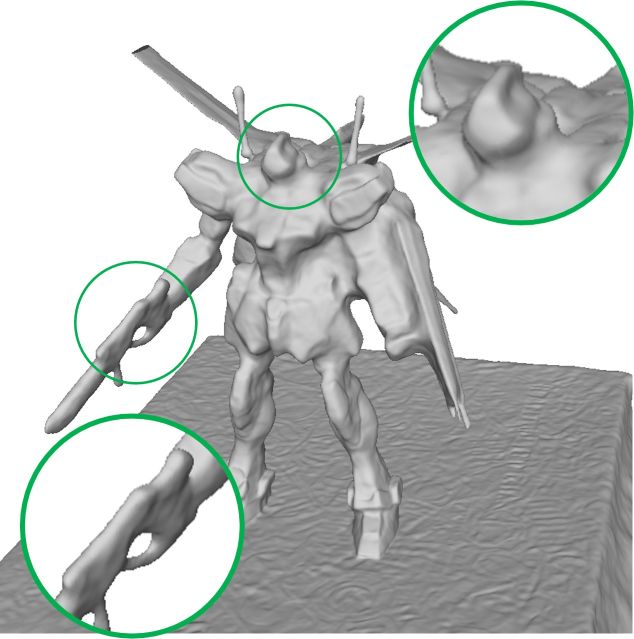}
        \caption{Scaled-up VolSDF}
        \label{fig:mesh_scaleup}
    \end{subfigure}
    }
    \caption{Qualitative comparison of surface reconstruction quality for the BlendedMVS dataset.}
    \label{fig:Meshresults}
\end{figure*}
\begin{figure}[t]
    \centering
    \includegraphics[width=0.7\linewidth]{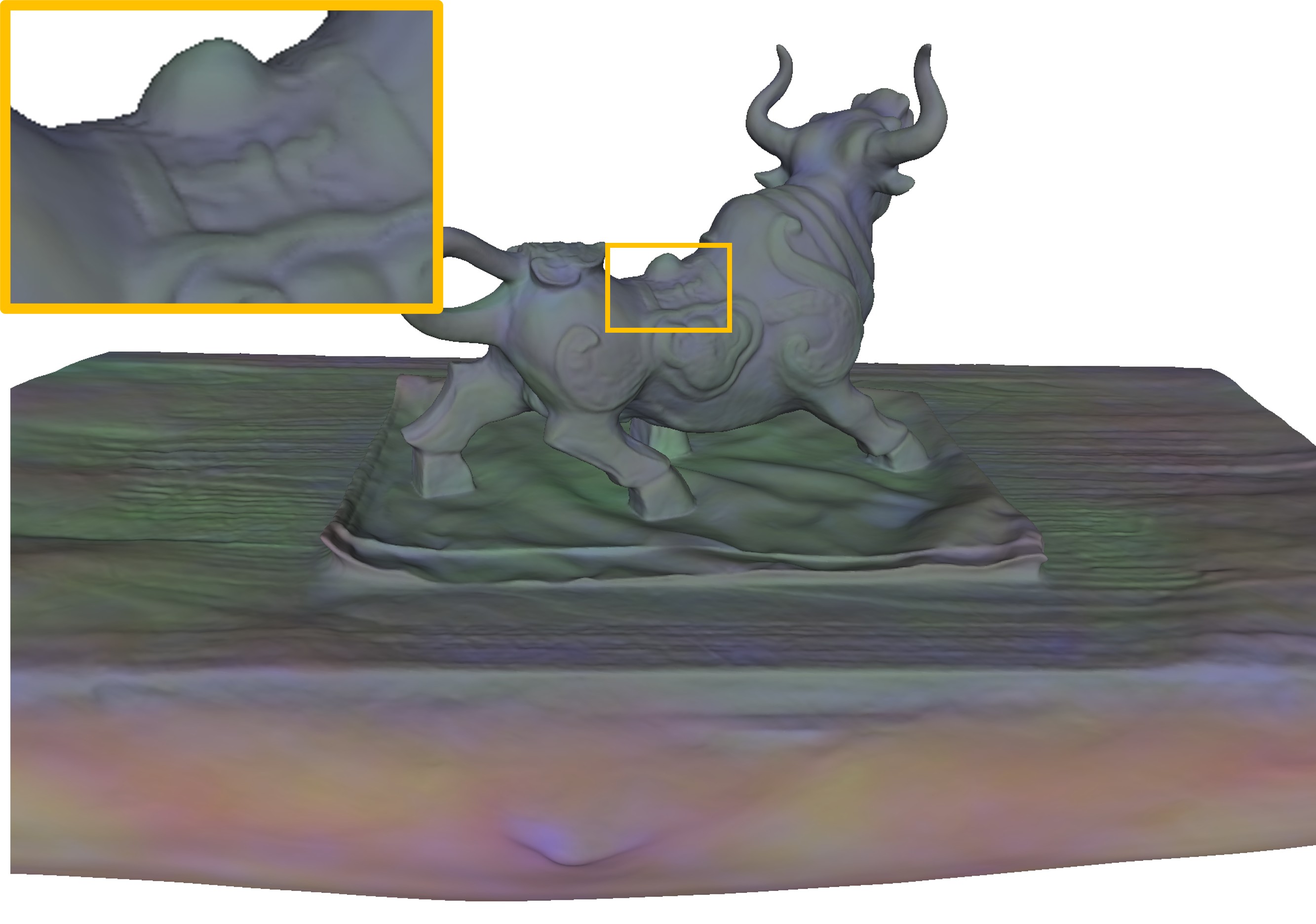}
    \caption{Visualization of norms of weighted feature vectors, $\boldsymbol{F}\cdot\rm{diag}(\boldsymbol{w})$. 
    The norms of low--, middle--, and high--frequency features are visualized as \textcolor{red}{red}, \textcolor{teal}{green}, and \textcolor{blue}{blue} channels, respectively.}
    \label{fig:norm}
\end{figure}
\begin{figure}[t]
    \centering
    \begin{subfigure}[c]{0.3\columnwidth}
        \includegraphics[width=\columnwidth]{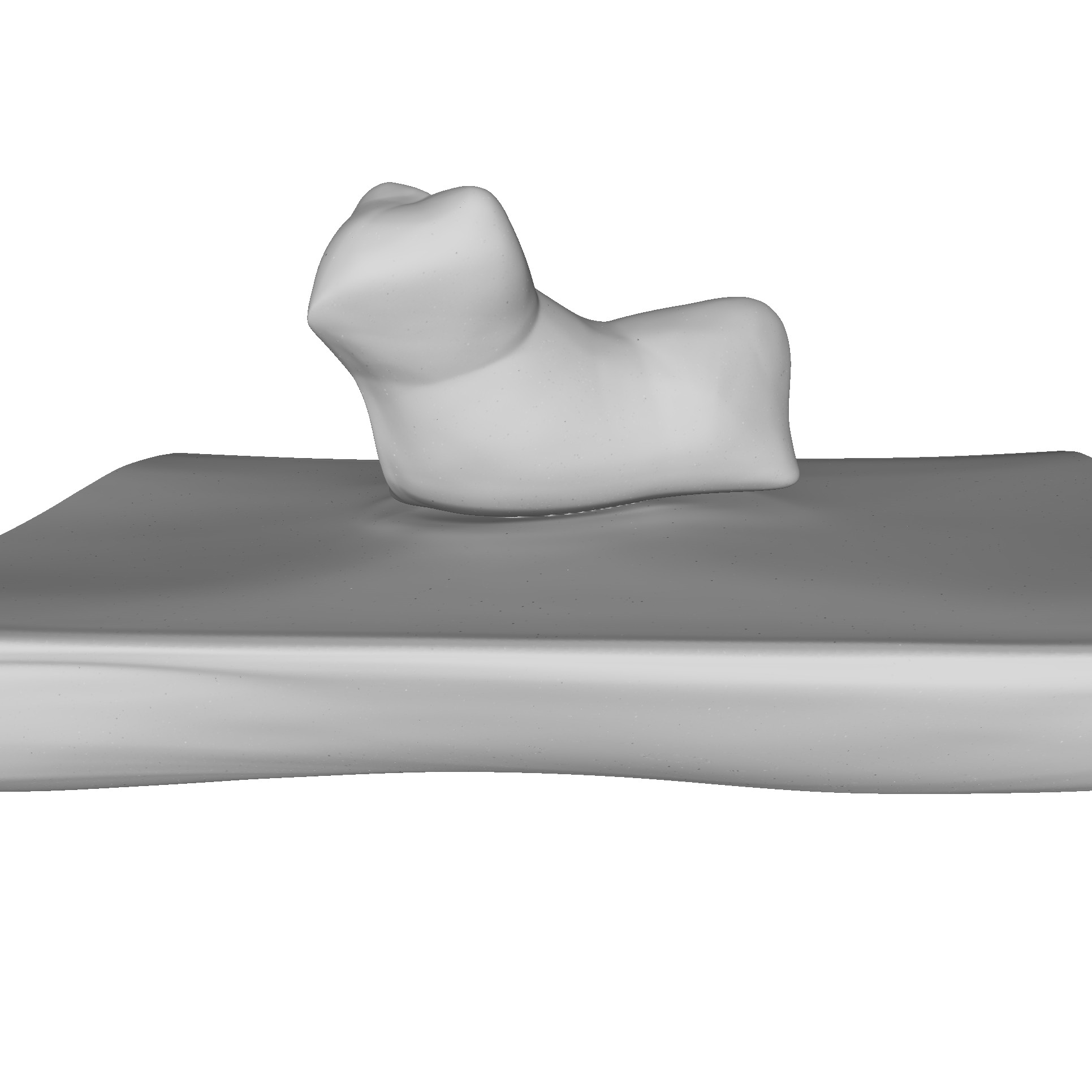}
        \caption{Low frequency}
        \label{fig:lowfreqmesh}
    \end{subfigure}
    \begin{subfigure}[c]{0.3\columnwidth}
        \includegraphics[width=\columnwidth]{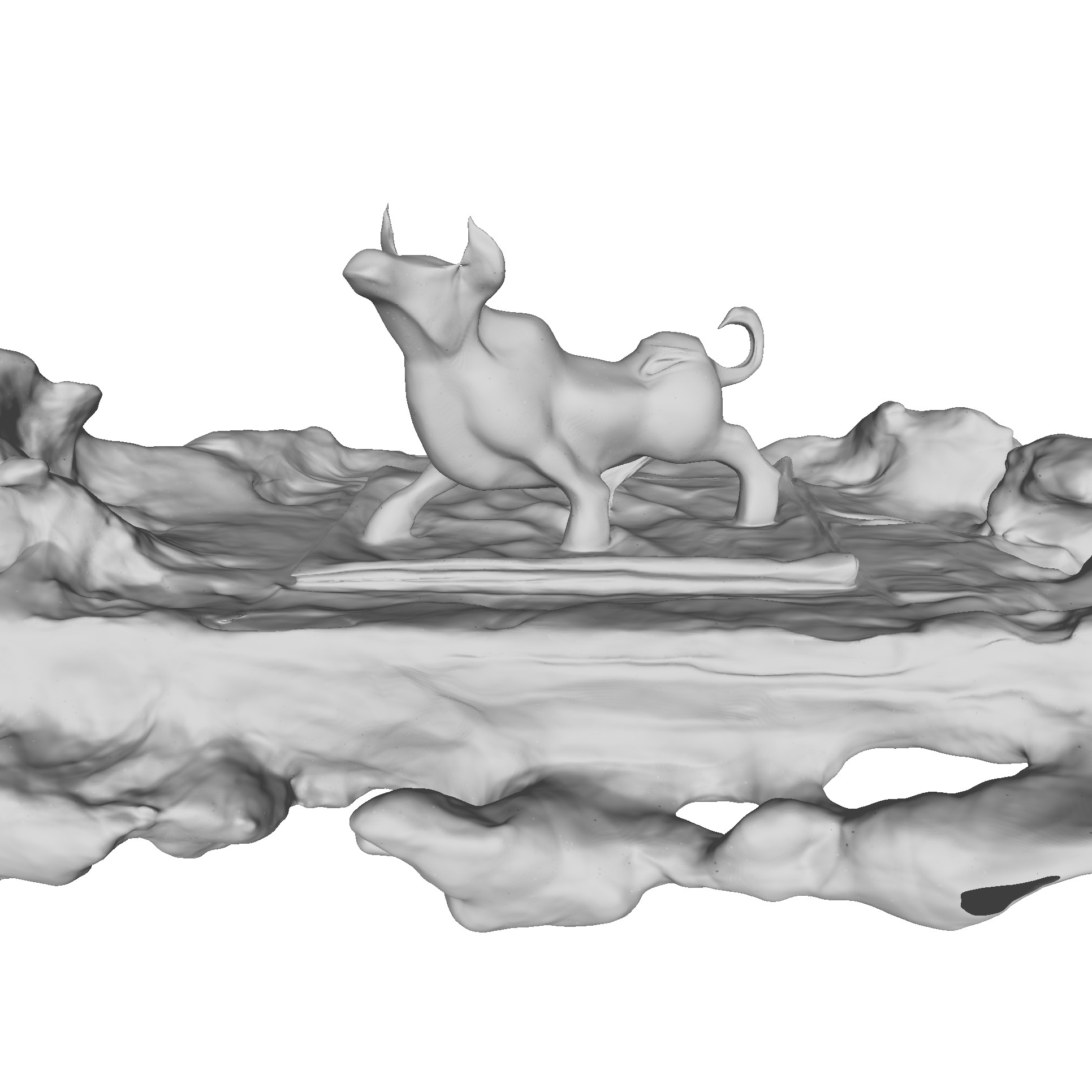}
        \caption{Middle frequency}
        \label{fig:middlefreqmesh}
    \end{subfigure}
    \begin{subfigure}[c]{0.3\columnwidth}
        \includegraphics[width=\columnwidth]{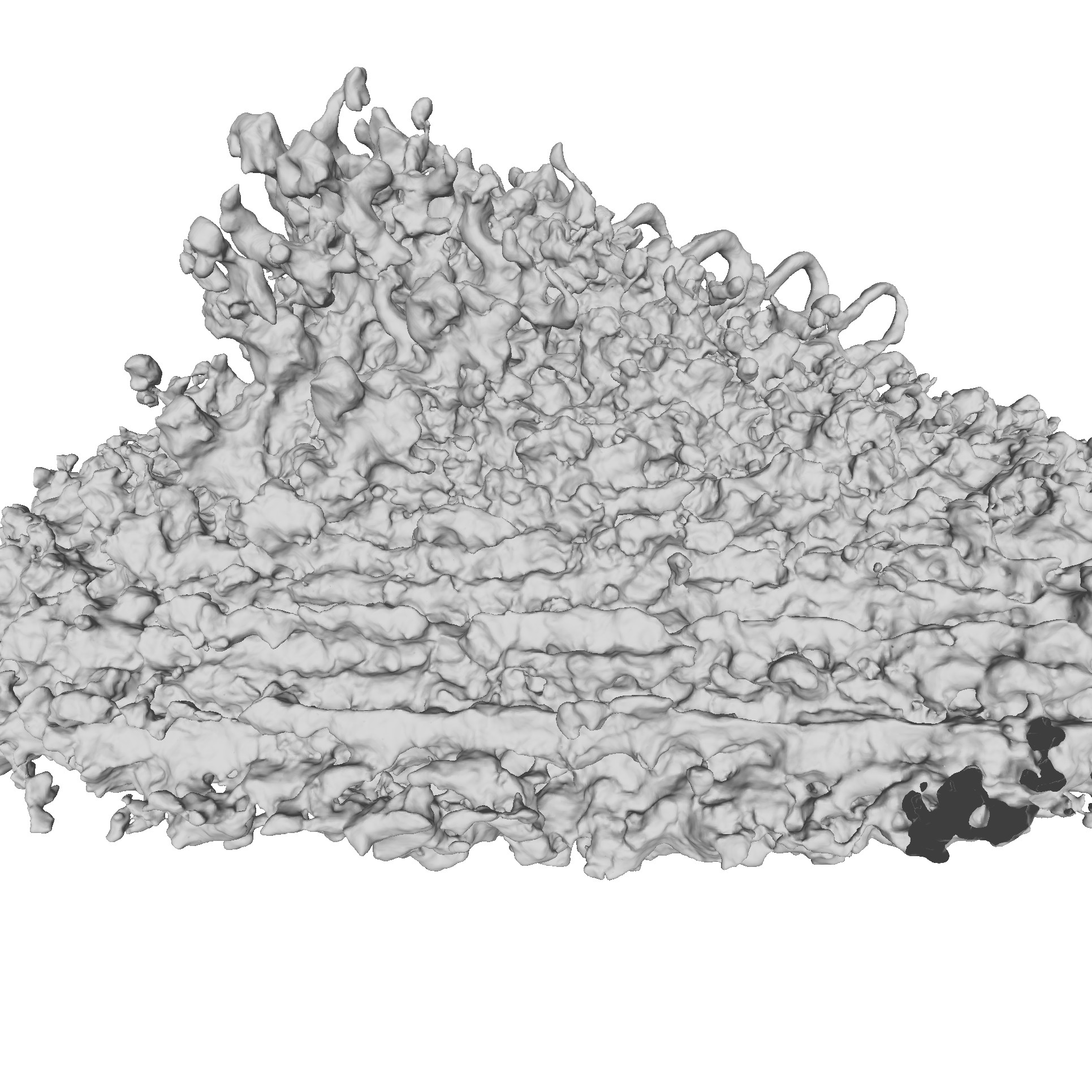}
        \caption{High frequency}
        \label{fig:highfreqmesh}
    \end{subfigure}
    \caption{Reconstructed meshes for each frequency band.}
    \label{fig:mesheachfreq}
\end{figure}

\begin{table}[tb]
    \centering
    \resizebox{\columnwidth}{!}{%
    \begin{tabular}{c|ccccccccc}
        \toprule
        $N_{\rm{L}},N_{\rm{M}},N_{\rm{H}}$ & Doll & Egg & Head & Angel & Bull & Robot & Dog & Bread & Camera \\
        \midrule
        2,2,2 & 26.22 & 27.48 & \textbf{27.29} & 30.52 & 26.33 & 26.69 & 28.56 & 30.22 & 23.08 \\
        1,2,3 & 26.01 & 27.38 & 27.00 & 30.37 & 26.35 & 26.58 & 28.47 & 30.85 & 22.95 \\
        1,3,2 & \textbf{26.23} & 27.44 & 27.07 & 30.44 & 26.25 & 26.76 & 28.42 & 29.73 & \textbf{23.11} \\
        2,1,3 & 25.96 & \textbf{27.50} & 26.95 & 29.97 & 26.28 & 26.67 & 28.50 & 29.46 & 22.92 \\
        2,3,1 & 26.00 & 27.03 & 27.15 & 30.35 & 26.04 & 26.51 & 28.75 & \textbf{31.80} & 23.09 \\
        3,1,2 & 26.02 & 27.34 & 27.05 & 30.50 & \textbf{26.38} & \textbf{26.84} & 28.67 & 30.17 & 23.01 \\
        3,2,1 & 26.18 & 26.85 & 27.16 & \textbf{30.56} & 26.14 & 26.61 & \textbf{28.82} & 31.73 & 23.01 \\
        \bottomrule
    \end{tabular}
    }
    \caption{Quantitative comparison of scene rendering performance of various assignments of frequencies to each encoder, in terms of PSNR.
    \textbf{Bold} texts denote the best score in each scene.}
    \label{tab:ablation_freqassignment}
\end{table}

We evaluate the performance of \method~ for the tasks of viewpoint-based scene rendering 
and 3D surface reconstruction across various complex, real-world scenes, comparing it against appropriate baselines.

\subsection{Experimental Setup}
\label{sec:experimentalsetup}
\noindent\textbf{Implementation details:}
We implemented \method~ in Pytorch~\cite{paszke2019pytorchimperativestylehighperformance}
and performed experiments on an NVIDIA A40 GPU with 48GB RAM.
All three encoders of \method~ have 6 layers with 256 dimensions per layer,
while the decoder has 2 layers with 256 dimensions per layer. 
We set the frequency level $\level~=6$ and distribute it evenly to each encoder,
i.e., each encoder deals with two frequencies. Additionally, we also concatenate the original input point coordinates to the input of each frequency encoder.
The training loss (Eq.~\ref{equ:similaritymatrix}) is computed with $\lambda=0.1$.
We set the initial learning rate to $0.005$ for all the parameters in the model,
which are optimized by the Adam optimizer~\cite{kingma2017adammethodstochasticoptimization}. 
The color network $\rm{MLP_{RGB}}$ has 4 layers with 256 dimensions.

\noindent\textbf{Dataset:}
We evaluate our method quantitatively and qualitatively on the BlendedMVS dataset~\cite{yao_blendedmvs_2020},
which consists of various object-centric real-world scenes with backgrounds.
Following the protocol of prior work~\cite{yariv_volume_2021}, we selected the same 9 scenes for evaluation. 
Each scene is composed of 31 to 144 multi-view images with a resolution of $768\times576$.

\noindent\textbf{Evaluation metrics:}
We evaluate the performance of competing methods for the task of view-dependent scene rendering 
using standard metrics such as peak signal-to-noise ratio (PSNR) measured in dB, structural similarity index measure (SSIM)~\cite{WangSSIM2004}, and learned perceptual image patch similarity (LPIPS)~\cite{zhang2018unreasonableeffectivenessdeepfeatures}. Besides, we also qualitatively evaluate the quality of the reconstructed 3D mesh (since the ground truth mesh is not available for this dataset).

\noindent\textbf{Baselines:}
\method~ is flexible in design and can work with any off-the-shelf decoder. For our experiments, we use the popular VolSDF decoder~\cite{yariv_volume_2021}.  %
Given this setup, to evaluate the effectiveness of our method, we compare our approach against VolSDF~\cite{yariv_volume_2021} and a customized, challenging baseline called Scaled-up VolSDF.
The Scaled-up VolSDF is an adaptation of VolSDF, where the number of parameters is increased to be roughly the same as Ours, for fair comparison. This baseline has a surface encoding network with 8 layers with 427 dimensions each, instead of the typical 256 dimensions in VolSDF. %

\subsection{Results}
\label{sec:exper_results}

Table~\ref{tab:BMVS} summarizes the results of quantitative comparisons of our method against VolSDF and Scaled-up VolSDF.
\method~ achieves the highest PSNR and SSIM and the lowest LPIPS score for all scenes in the dataset, except for the simpler, less textured scene -- the \textit{Bread} scene, registering gains of up to $2\%$ on SSIM over the Scaled-up VolSDF baseline on an overall assessment.

Qualitative comparisons of rendered images on the \textit{Doll}, \textit{Bull}, and \textit{Robot} scenes are presented in Fig.~\ref{fig:RGBresults}.
As illustrated in these images, \method~ considerably improves rendering quality, especially the fine details of objects.
Qualitative comparisons on the reconstructed meshes are presented in Fig.~\ref{fig:Meshresults}.
In particular, the reconstructed surfaces by \method~ have higher fidelity and are better at preserving the details, 
e.g., bands on the \textit{Doll}'s cloth, the \textit{Bull}'s saddle, the \textit{Robot}'s gun and face.
Moreover, we also notice that the eyeballs of the \textit{Doll} are inappropriately reconstructed as concave surfaces by VolSDF and Scaled-up VolSDF, while \method~ does a better job of the reconstruction. 
We see that \method~ outperforms VolSDF and Scaled-up VolSDF both in terms of scene rendering and surface reconstruction quality.
These results attest to the effectiveness of our method and show that the gain in performance cannot simply be attributed to scaling up the number of parameters. More visualizations are present in the supplementary.

To verify that appropriate frequency bands are used in each region and that the encoders learn complementary features,
we visualize the norms of weighted features ($\boldsymbol{F}\cdot\rm{diag}(\boldsymbol{w})$), that are redundancy-aware, for each frequency band and the quality of meshes obtained for each frequency band.

\noindent\textbf{Norms of weighted features for each frequency-band:}
Fig.~\ref{fig:norm} shows the reconstructed mesh of the \textit{Bull} scene,
where the vertex color denotes the norm of weighted features. 
For this visualization, the low--, middle--, and high--frequency features are mapped to red, green, and blue channels, respectively.
Note also that the norm is scaled to $[0.4, 1.0]$ for visibility.
We see that high--frequency information (blue) is more dominant in regions with finer details, e.g., decorative carving,
whereas low--frequency information (red) is mainly used for unobserved and interpolated areas where details are missing.
Our encoders successfully distinguish between smooth and rough surface regions
and model them with different frequency bands.

\noindent\textbf{Surface reconstructions for each frequency domain:}
To examine whether each encoder learns complementary features, we decode the output of each frequency encoder independently and visualize the results. 
Figs.~\ref{fig:lowfreqmesh}, ~\ref{fig:middlefreqmesh}, and \ref{fig:highfreqmesh} are meshes reconstructed from feature vectors $\boldsymbol{f}_{\rm{L}}$, $\boldsymbol{f}_{\rm{M}}$, $\boldsymbol{f}_{\rm{H}}$, respectively, for the \textit{Bull} scene.
As shown in Fig.~\ref{fig:mesheachfreq},
the low--frequency mesh captures the global structure of the scene well,
the middle--frequency mesh gets the rough shape of objects and some details,
while the high--frequency mesh captures the fine details.
These results show %
that the encoders successfully learn complementary, frequency-dependent features. 

\subsection{Ablation Study}
\label{sec:exper_ablation}
\begin{table}[t]
    \centering
    \resizebox{\columnwidth}{!}{%
    \begin{tabular}{cccc}
        \toprule
        & (a) Scaled-up VolSDF & (b) \makecell{Ours w/o \\ redundancy-aware weighting} & (c) Ours \\
        \midrule
        PSNR ($\uparrow$)& 28.32 & 28.31 & \textbf{28.56}\\
        \midrule
        SSIM ($\uparrow$)& 0.949 & 0.950 & \textbf{0.952}\\
        \midrule
        LPIPS ($\downarrow$) & 0.028 & 0.027 & \textbf{0.026}\\
        \bottomrule
    \end{tabular}
    }
    \caption{Ablation of the redundancy-aware weighting module: We show quantitative results for the \textit{Dog} scene using the Scaled-up VolSDF, Ours without redundancy-aware weighting, and Ours.}
    \label{tab:ablaiton_redundancy-aware}
\end{table}
\noindent\textbf{Ablation of the redundancy-aware weighting:}
To evaluate the effect of our redundancy-aware weighting module,
we take the average of features from the different encoders instead of applying the redundancy-aware weighting.
As seen in Table~\ref{tab:ablaiton_redundancy-aware}, the Scaled-up VolSDF and Ours without redundancy-aware weighting perform worse than our proposed \method~, attesting to its efficacy.

\noindent\textbf{Assignment of frequency-bands to each encoder:}
Unlike the experiments in Sec.~\ref{sec:exper_results}, we construct variants of our model where we assign frequency levels to the encoders unevenly.
Note that the total number of frequency levels $\level~$ is set to 6.
Table~\ref{tab:ablation_freqassignment} shows quantitative results 
for different configurations, under this setup.
Though the optimal assignment of frequency domains seems to vary depending on the scene,
the even distribution ($(N_{\rm{L}},N_{\rm{M}},N_{\rm{H}})=(2,2,2)$) performs most stably across various scenes.

\section{Conclusions}
\label{sec:conclusions}
In this work, we proposed \method~, a novel approach for neural implicit surface representation.
\method~ stratifies the scene into multiple frequency levels according to the surface frequencies and leverages a novel \emph{redundancy-aware weighting} module to effectively capture complementary information by promoting mutual dissimilarity of the encoded features.
Empirical results show that coupling \method~ encoders with the VolSDF decoder improves the qualities of reconstructed mesh as well as their viewpoint-based surface renderings.

Going forward, we plan to evaluate \method~ on other datasets and backbones. 
Combining \method~ with object-compositional frameworks, such as ObjectSDF~\cite{wu_object-compositional_2022} and RICO~\cite{li_rico_2023}, should allow us to reconstruct more complex scenes with multiple objects,
which can be leveraged for higher fidelity complex 3D simulation, and 3D content generation for AR/VR.

{
    \small
    \bibliographystyle{ieeenat_fullname}
    \bibliography{references}
}

\clearpage
\setcounter{page}{1}
\maketitlesupplementary

The following summarizes the supplementary materials we present:
 \begin{enumerate}
     \item Ablation study of the \emph{redundancy-aware weighting} module.
     \item Comparative study of the number of frequency levels.
     \item Comparative study of encoder architecture variants.
 \end{enumerate}

\begin{table*}[tbh]
    \centering
    \begin{tabular}{cccc}
        \toprule
        & (a) Scaled-up VolSDF & (b) \makecell{Ours w/o \\ redundancy-aware weighting} & (c) Ours \\
        \midrule
        PSNR ($\uparrow$)& 28.32 & 28.31 & \textbf{28.56}\\
        \midrule
        SSIM ($\uparrow$)& 0.949 & 0.950 & \textbf{0.952}\\
        \midrule
        LPIPS ($\downarrow$) & 0.028 & 0.027 & \textbf{0.026}\\
        \bottomrule
    \end{tabular}
    \caption{Ablation of the redundancy-aware weighting module: We show quantitative results for the \textit{Dog} scene using the Scaled-up VolSDF, Ours without redundancy-aware weighting, and Ours.}
    \label{tab:ablaiton_redundancy-aware_supple}
\end{table*}

\begin{table*}[tbh]
    \centering
\resizebox{\textwidth}{!}{%
\begin{tabular}{lc|ccccccccc|c}
    \toprule
    Method & Frequency level ($\level~$) & Doll & Egg & Head & Angel & Bull & Robot & Dog & Bread & Camera & \textbf{Mean} \\
    \midrule
    Scaled-up VolSDF & 6 & 26.07 & 27.15 & 26.62 & 30.37 & 26.08 & 25.07 & 28.32 & 29.44 & 23.02 & 26.90\\
    Ours & 6 & 26.22 & 27.48 & 27.29 & 30.52 & 26.33 & 26.69 & 28.56 & 30.22 & 23.08 & 27.38\\
    \midrule
    Scaled-up VolSDF & 9 & 25.69 & 26.66 & 26.94 & 28.59 & 26.02 & 22.67 & 26.78 & 32.62 & 23.45 & 26.60\\
    Ours & 9 & 26.10 & 27.47 & 27.24 & 30.56 & 25.78 & 26.85 & 28.88 & 30.08 & 23.28 & 27.36\\
    \midrule
    Scaled-up VolSDF & 12 & -- & -- & -- & -- & -- & -- & 24.86 & -- & 19.59 & --\\
    Ours & 12 & 26.02 & 27.54 & 25.81 & 30.56 & 26.89 & 26.66 & 28.62 & 30.18 & 30.26 & 27.21\\
    \bottomrule
\end{tabular}
}
\caption{Comparison of viewpoint-based rendering performance with a varying number of frequencies, as measured by PSNR.
-- denotes that the method failed to construct a mesh during training.} %
\label{table:ablation_frequencylevel}
\end{table*}

\begin{table*}[tbh]
    \centering
    \begin{tabular}{c|ccccccccc|c}
        \toprule
        $N_{\rm{L}},N_{\rm{M}},N_{\rm{H}}$ & Doll & Egg & Head & Angel & Bull & Robot & Dog & Bread & Camera & \textbf{Mean} \\
        \midrule
        $6,6,6$ & 26.22 & 27.48 & 27.29 & 30.52 & 26.33 & 26.69 & 28.56 & 30.22 & 23.08 & 27.38\\%$\pm$2.13\\
        $5,5,5$ & 26.18 & 27.47 & 27.14 & 30.42 & 26.37 & 26.62 & 28.55 & 30.20 & 23.10 & 27.34\\%$\pm$2.11\\
        $4,4,4$ & 26.25 & 27.51 & 26.96 & 30.49 & 26.37 & 26.51 & 28.18 & 31.12 & 23.17 & 27.39\\%$\pm$2.24\\
        $4,5,6$ & 26.18 & 27.45 & 27.13 & 30.50 & 26.38 & 26.64 & 28.60 & 30.16 & 23.19 & 27.36\\%$\pm$2.10\\
        $2,4,6$ & 26.26 & 27.47 & 24.45 & 30.44 & 25.95 & 26.67 & 28.74 & 31.60 & 23.21 & 27.19\\%$\pm$2.55\\
        \bottomrule
    \end{tabular}
    \caption{Performance comparison of variants of \method~ with varying number of encoder layers, as measured by PSNR.}
    \label{table:ablation_layernumber}
\end{table*}

\begin{figure*}[tbh]
    \centering
    \begin{minipage}[c]{0.07\textwidth}
        \centering
        \rotatebox[origin=c]{90}{\textbf{Scaled-up VolSDF}}
    \end{minipage}
    \hfill
    \begin{subfigure}[c]{0.3\textwidth}
        \includegraphics[width=\columnwidth]{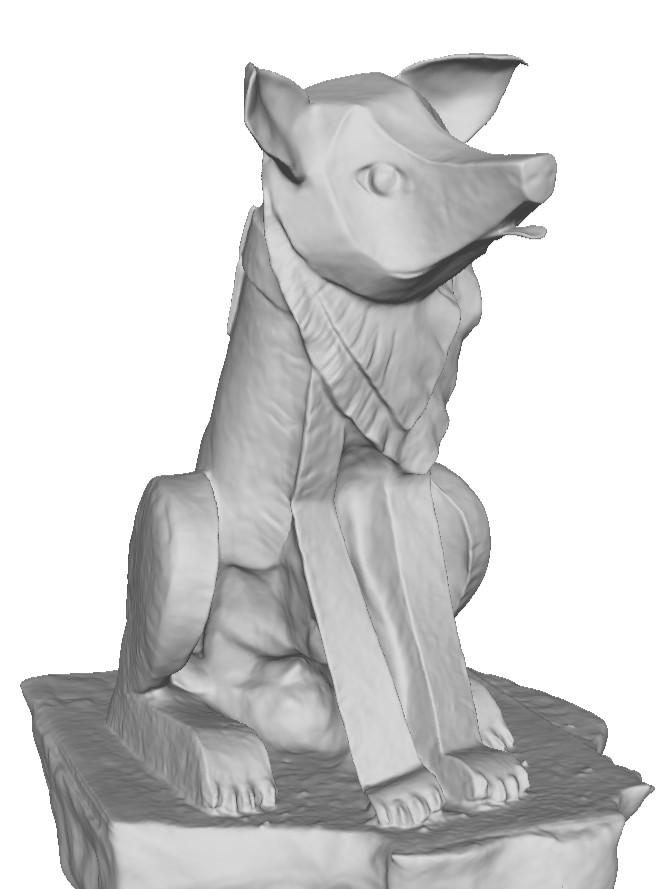}
    \end{subfigure}
    \begin{subfigure}[c]{0.3\textwidth}
        \includegraphics[width=\columnwidth]{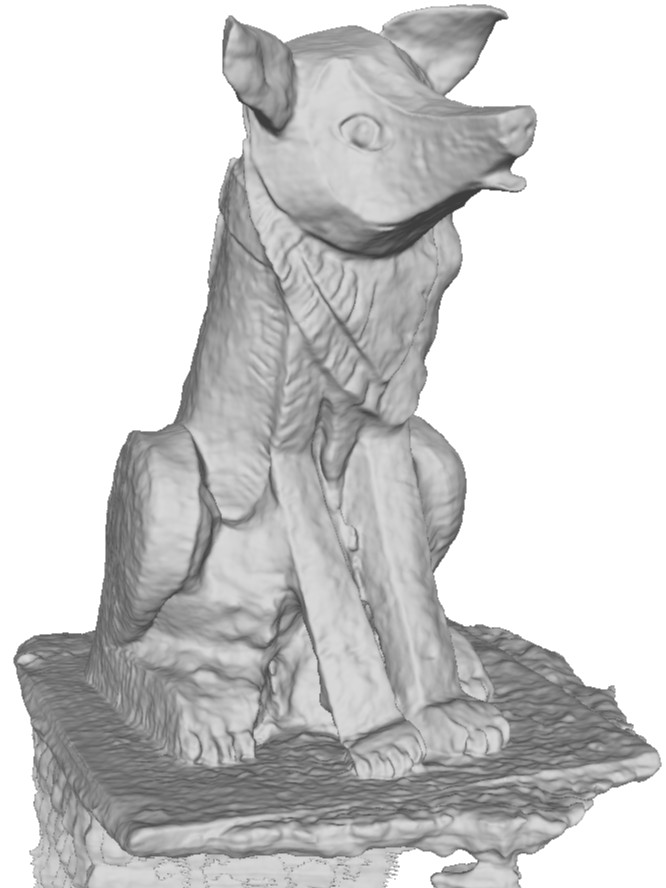}
    \end{subfigure}
    \begin{subfigure}[c]{0.3\textwidth}
        \includegraphics[width=\columnwidth]{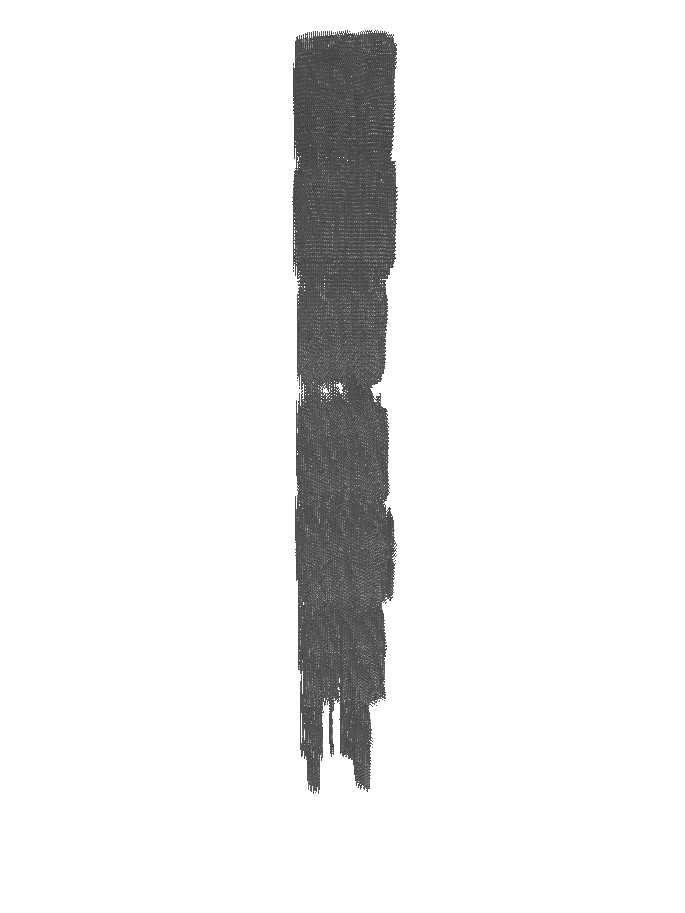}
    \end{subfigure}
    \begin{minipage}[c]{0.07\textwidth}
        \centering
        \rotatebox[origin=c]{90}{\textbf{Ours}}
    \end{minipage}
    \hfill
    \begin{subfigure}[c]{0.3\textwidth}
        \includegraphics[width=\columnwidth]{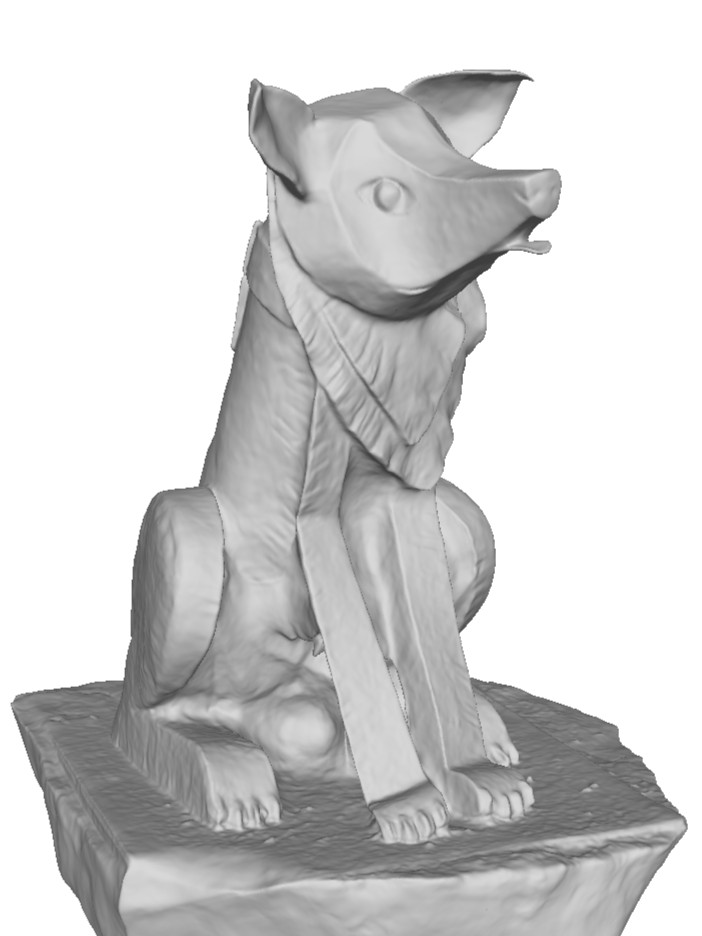}
        \caption{$\level~=6$}
    \end{subfigure}
    \begin{subfigure}[c]{0.3\textwidth}
        \includegraphics[width=\columnwidth]{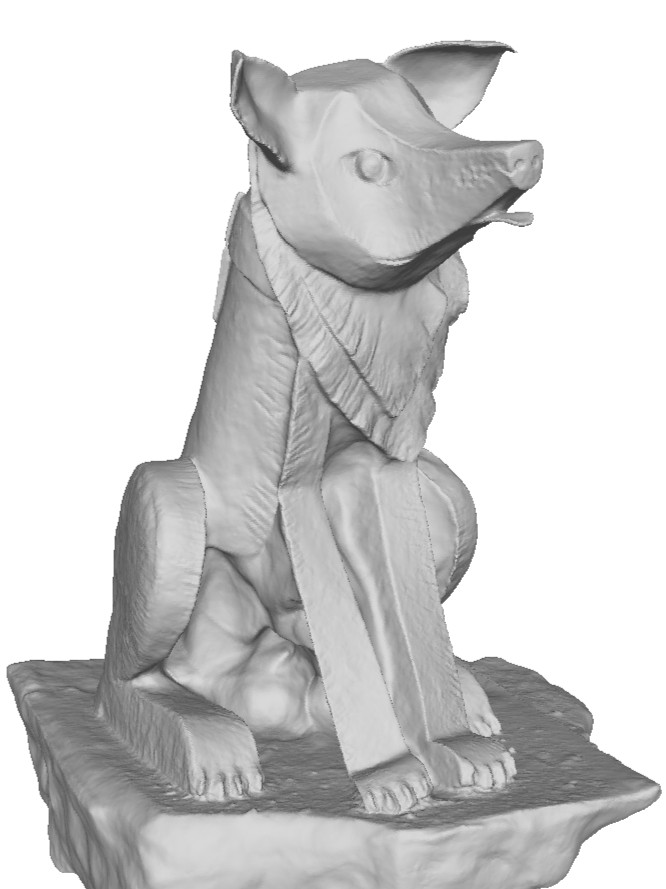}
        \caption{$\level~=9$}
    \end{subfigure}
    \begin{subfigure}[c]{0.3\textwidth}
        \includegraphics[width=\columnwidth]{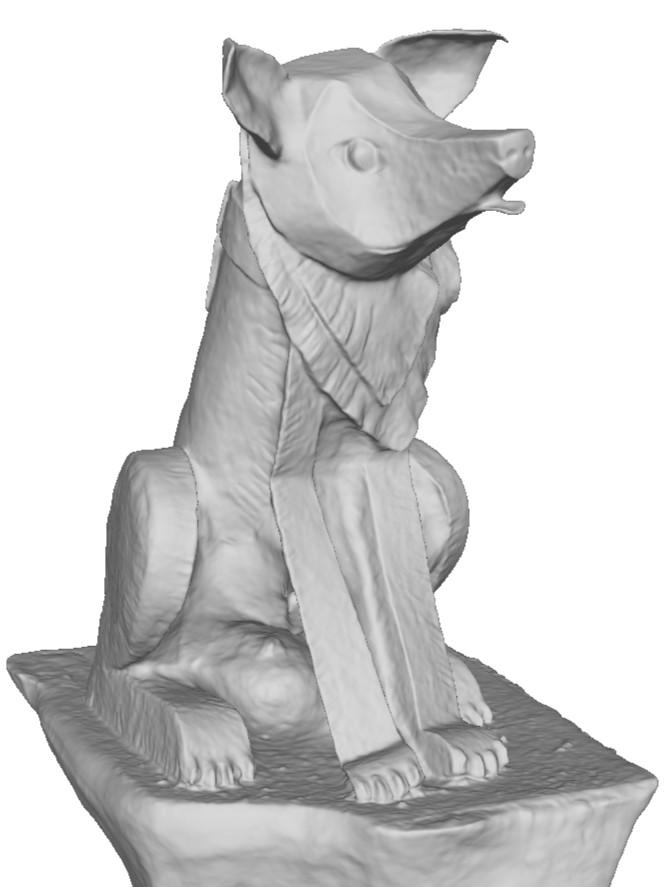}
        \caption{$\level~=12$}
    \end{subfigure}
    \caption{Qualitative comparison on the capability to deal with higher frequencies.}
    \label{fig:comparisonfrequency}
\end{figure*}

\begin{figure*}[tb]
    \centering
    \begin{minipage}[c]{0.07\textwidth}
        \centering
        \rotatebox[origin=c]{90}{\textbf{Doll}}
    \end{minipage}
    \hfill
    \begin{subfigure}[c]{0.3\textwidth}
        \includegraphics[width=\columnwidth]{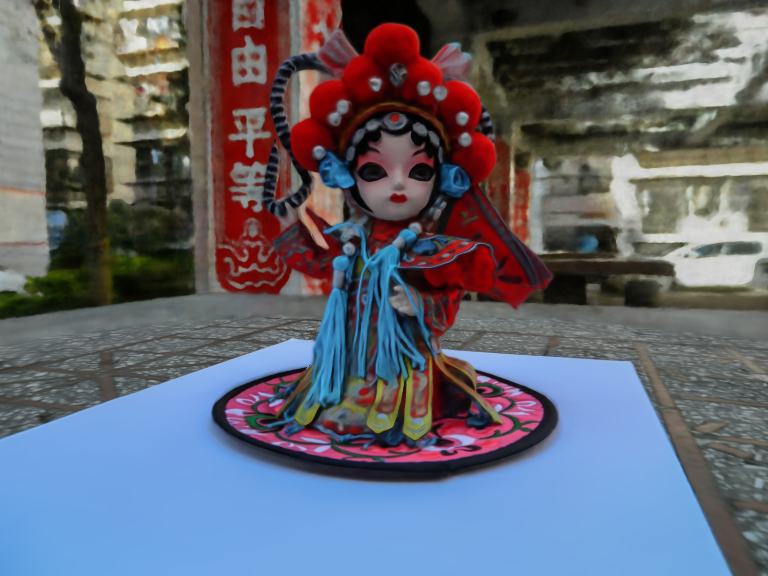}
    \end{subfigure}
    \begin{subfigure}[c]{0.3\textwidth}
        \includegraphics[width=\columnwidth]{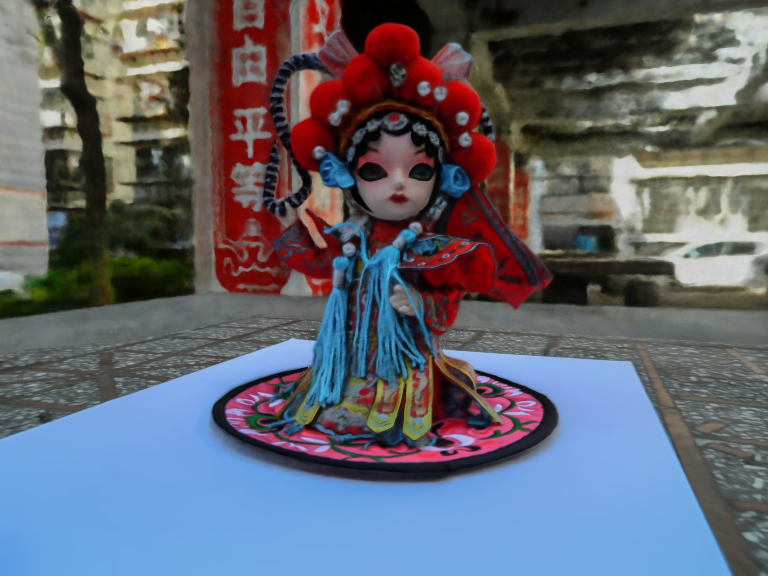}
    \end{subfigure}
    \begin{subfigure}[c]{0.3\textwidth}
        \includegraphics[width=\columnwidth]{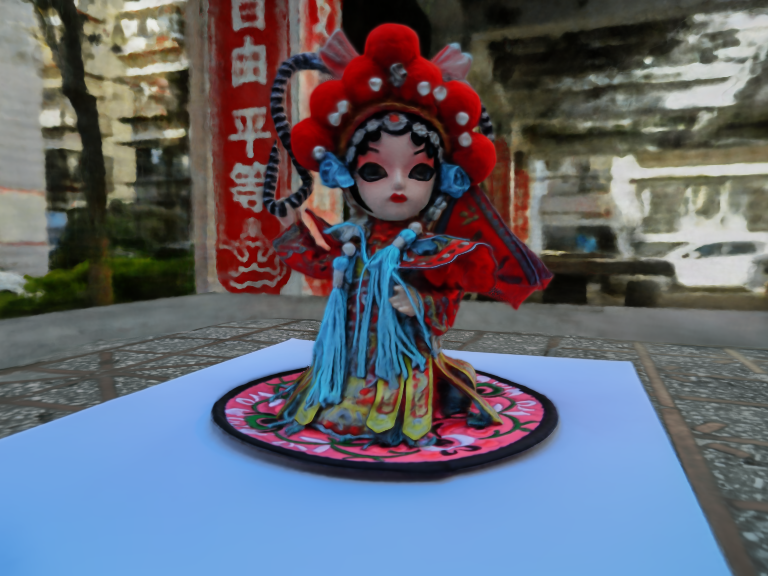}
    \end{subfigure}
    \begin{minipage}[c]{0.07\textwidth}
        \centering
        \rotatebox[origin=c]{90}{\textbf{Bull}}
    \end{minipage}
    \hfill
    \begin{subfigure}[c]{0.3\textwidth}
        \includegraphics[width=\columnwidth]{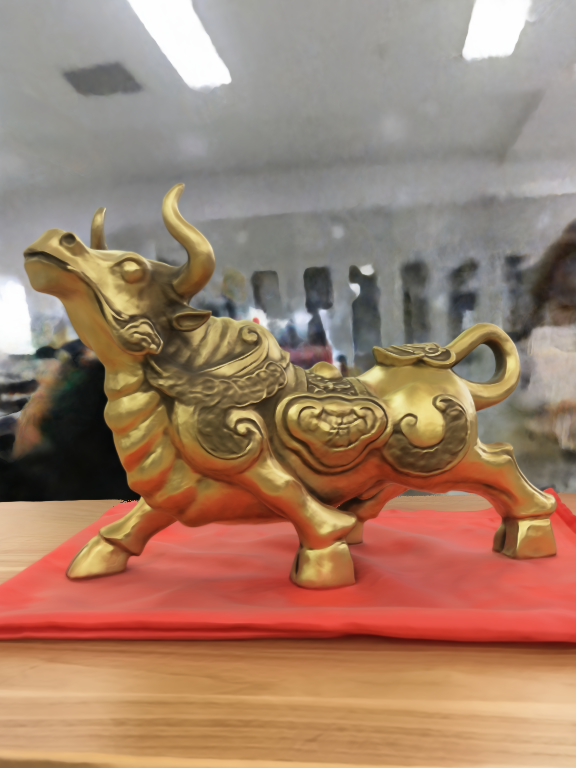}
    \end{subfigure}
    \begin{subfigure}[c]{0.3\textwidth}
        \includegraphics[width=\columnwidth]{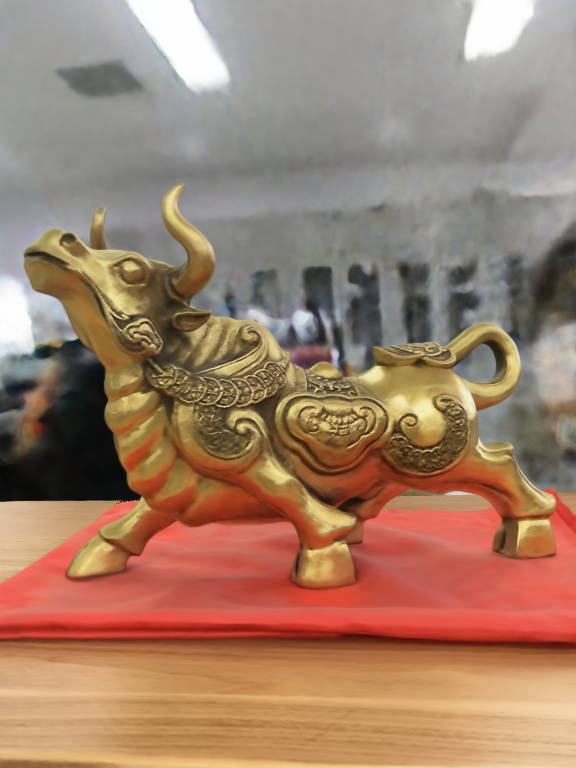}
    \end{subfigure}
    \begin{subfigure}[c]{0.3\textwidth}
        \includegraphics[width=\columnwidth]{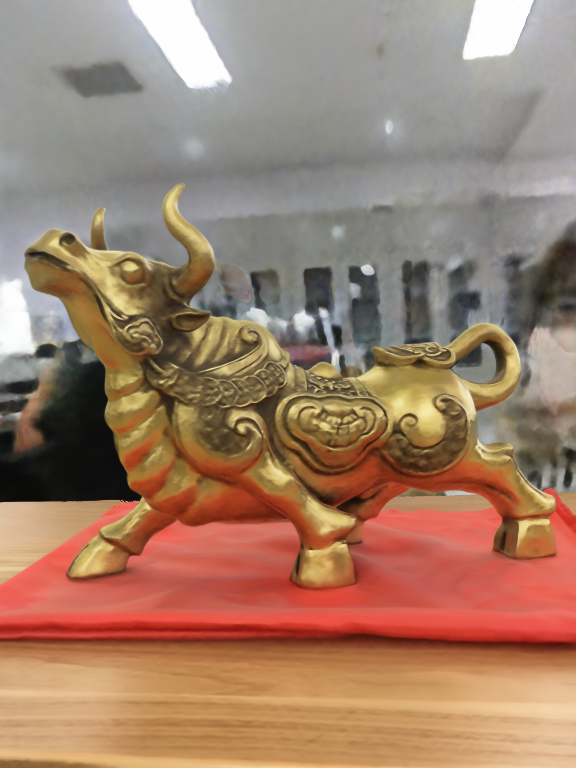}
    \end{subfigure}
    \begin{minipage}[c]{0.07\textwidth}
        \centering
        \rotatebox[origin=c]{90}{\textbf{Robot}}
    \end{minipage}
    \hfill
    \begin{subfigure}[c]{0.3\textwidth}
        \includegraphics[width=\columnwidth]{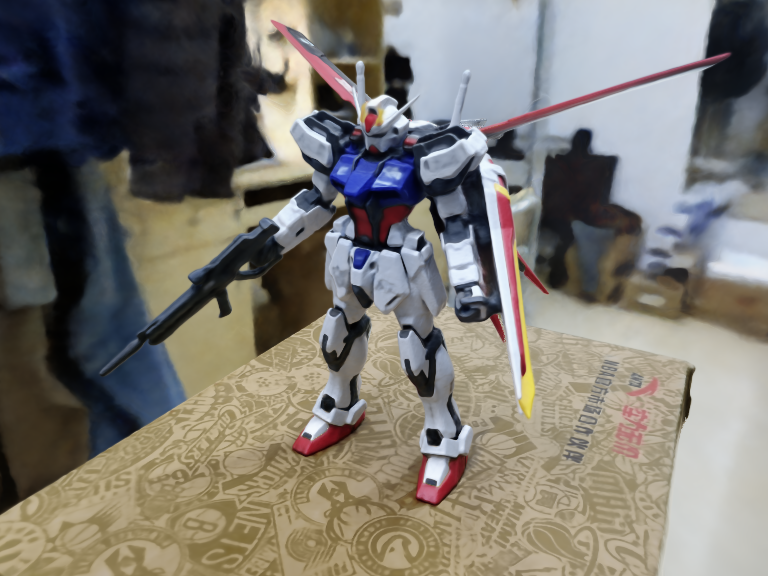}
        \caption{$\level~=6$}
        \label{fig:6freq_render}
    \end{subfigure}
    \begin{subfigure}[c]{0.3\textwidth}
        \includegraphics[width=\columnwidth]{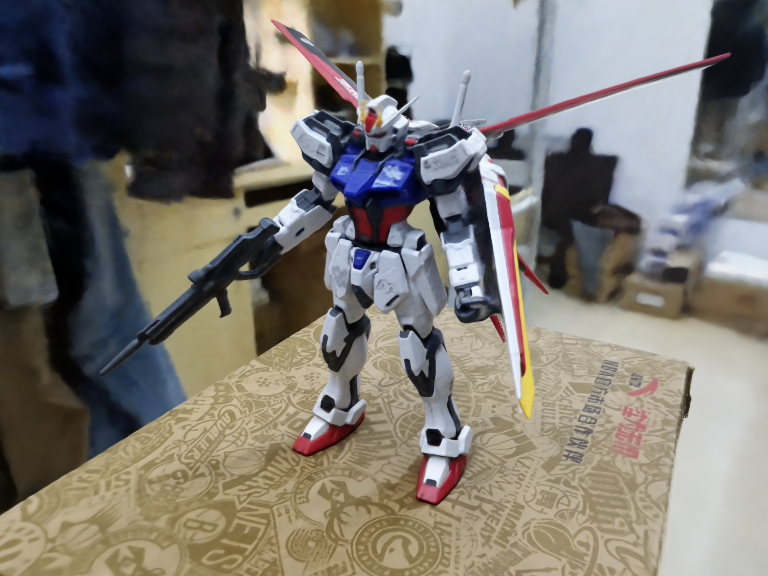}
        \caption{$\level~=9$}
        \label{fig:9freq_render}
    \end{subfigure}
    \begin{subfigure}[c]{0.3\textwidth}
        \includegraphics[width=\columnwidth]{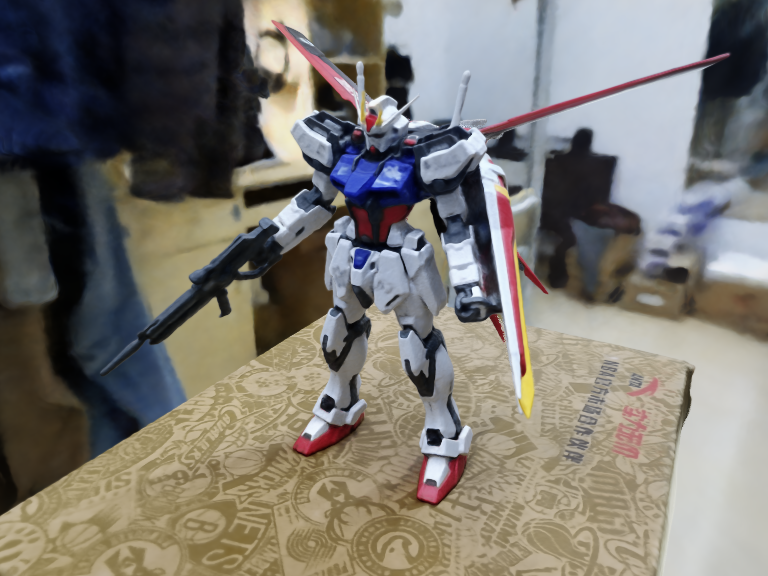}
        \caption{$\level~=12$}
        \label{fig:12freq_render}
    \end{subfigure}
    
    \caption{Qualitative comparison of viewpoint-based scene rendering with varying number of frequencies.}
    \label{fig:freq_render}
\end{figure*}

\begin{figure*}[tb]
    \centering
    \begin{minipage}[c]{0.07\textwidth}
        \centering
        \rotatebox[origin=c]{90}{\textbf{Doll}}
    \end{minipage}
    \hfill
    \begin{subfigure}[c]{0.3\textwidth}
        \includegraphics[width=\columnwidth]{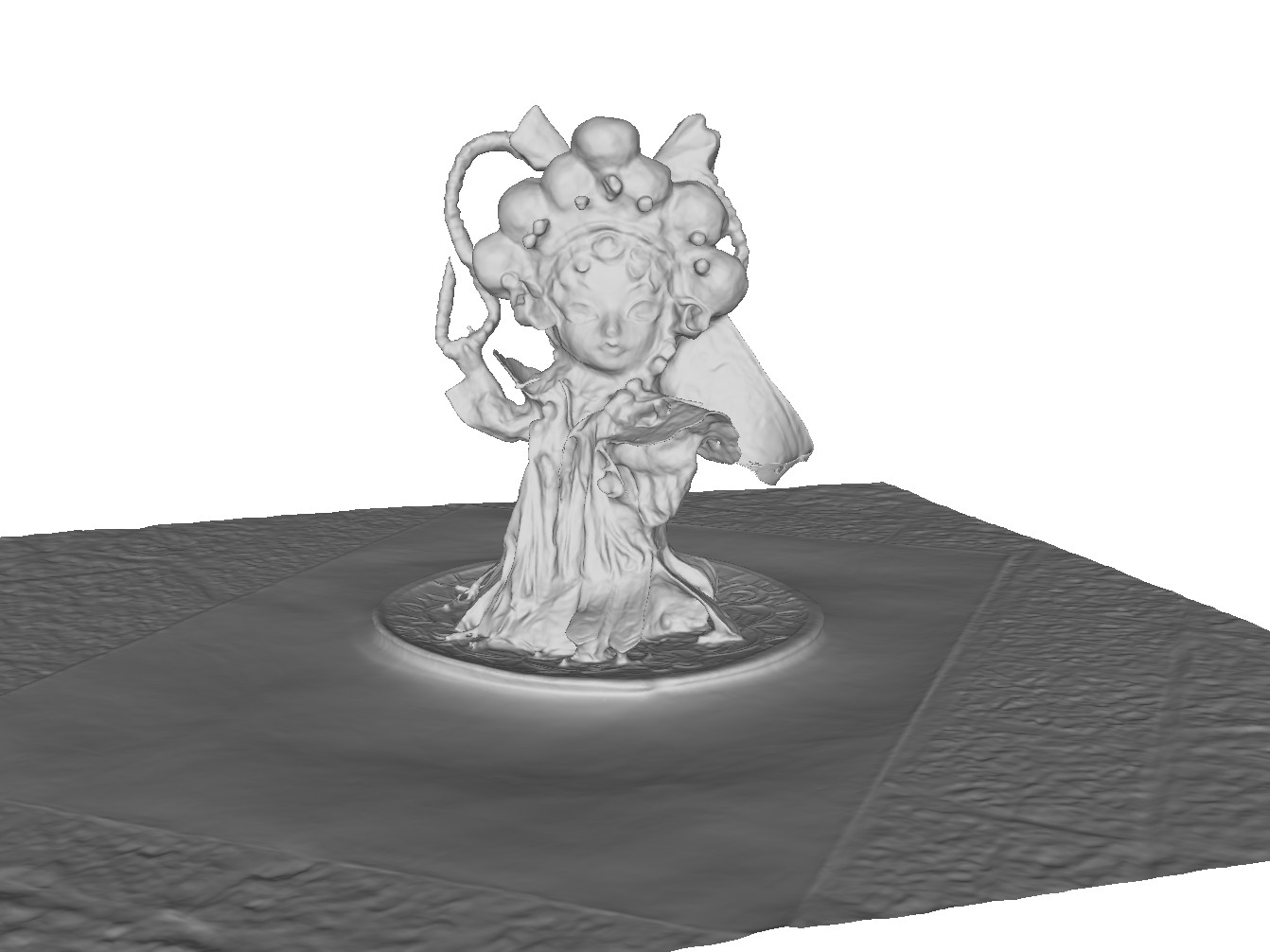}
    \end{subfigure}
    \begin{subfigure}[c]{0.3\textwidth}
        \includegraphics[width=\columnwidth]{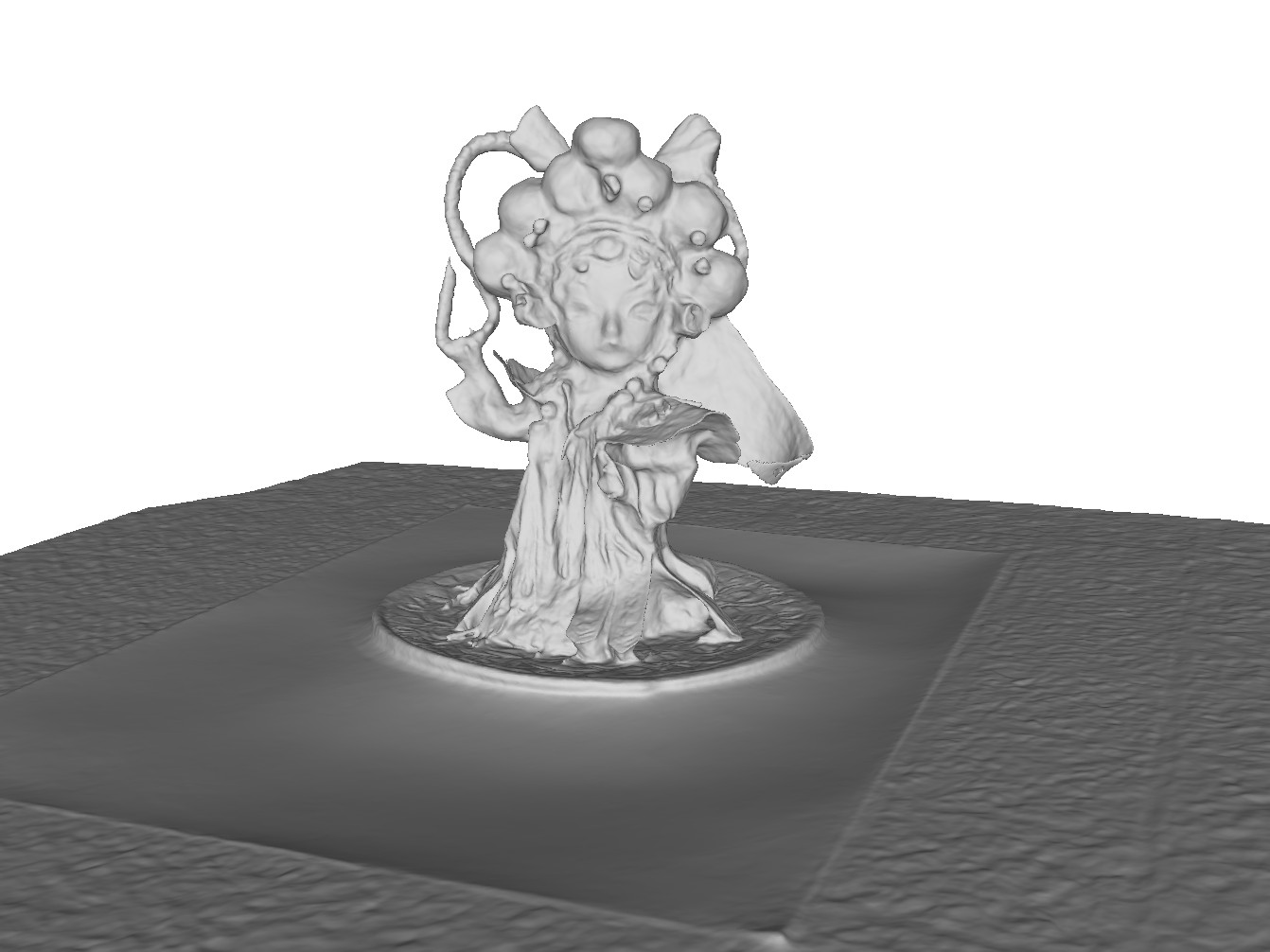}
    \end{subfigure}
    \begin{subfigure}[c]{0.3\textwidth}
        \includegraphics[width=\columnwidth]{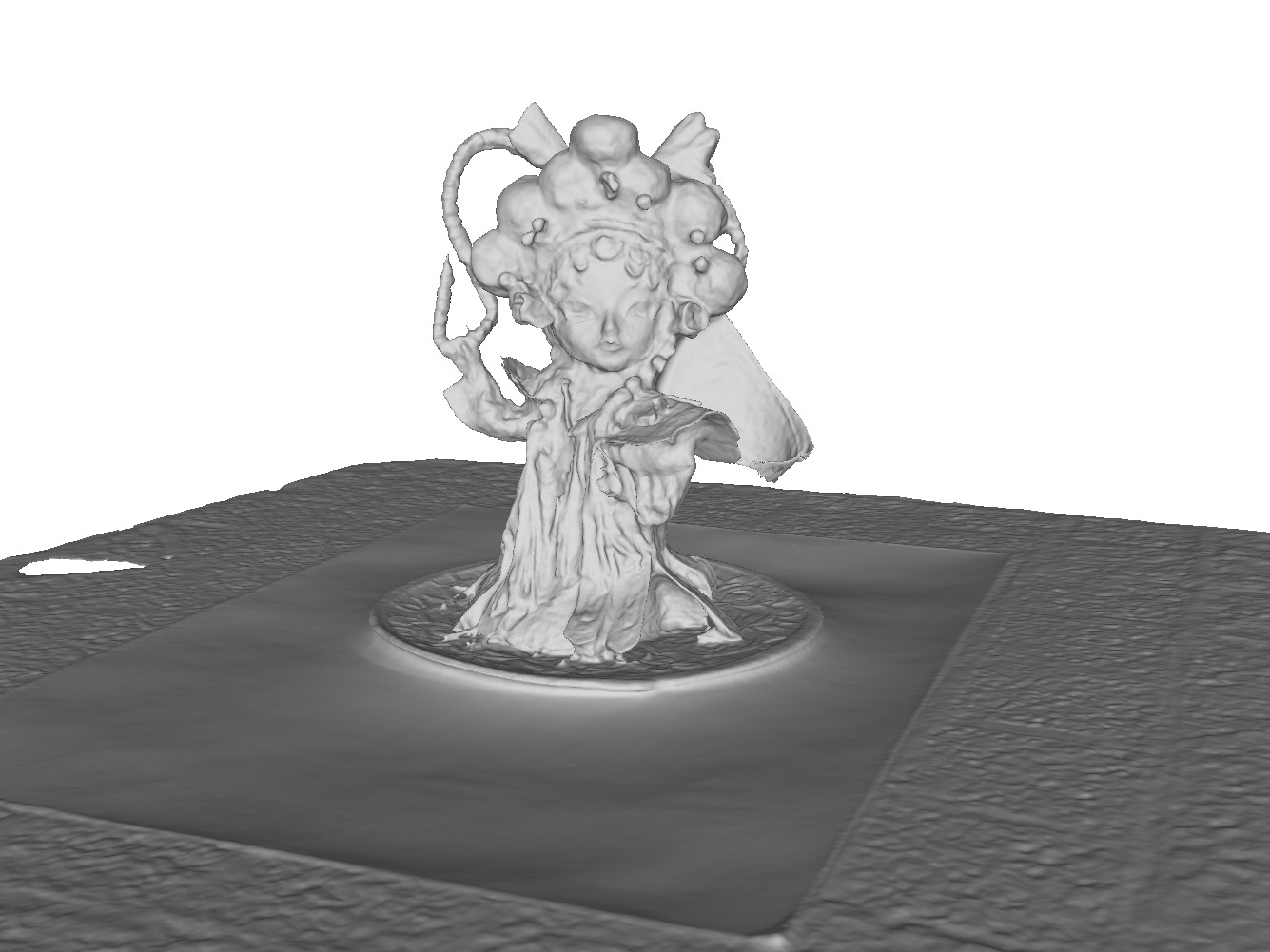}
    \end{subfigure}
    \begin{minipage}[c]{0.07\textwidth}
        \centering
        \rotatebox[origin=c]{90}{\textbf{Bull}}
    \end{minipage}
    \hfill
    \begin{subfigure}[c]{0.3\textwidth}
        \includegraphics[width=\columnwidth]{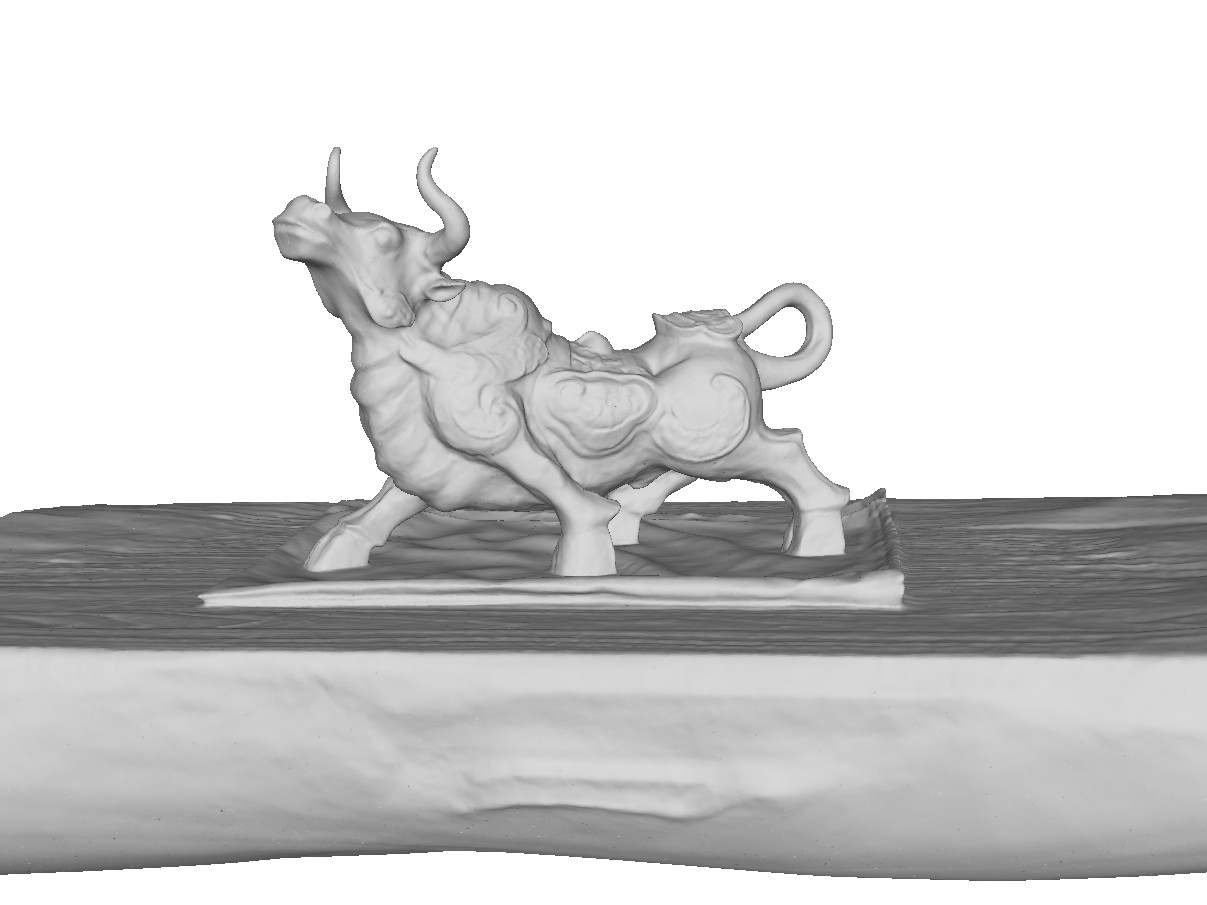}
    \end{subfigure}
    \begin{subfigure}[c]{0.3\textwidth}
        \includegraphics[width=\columnwidth]{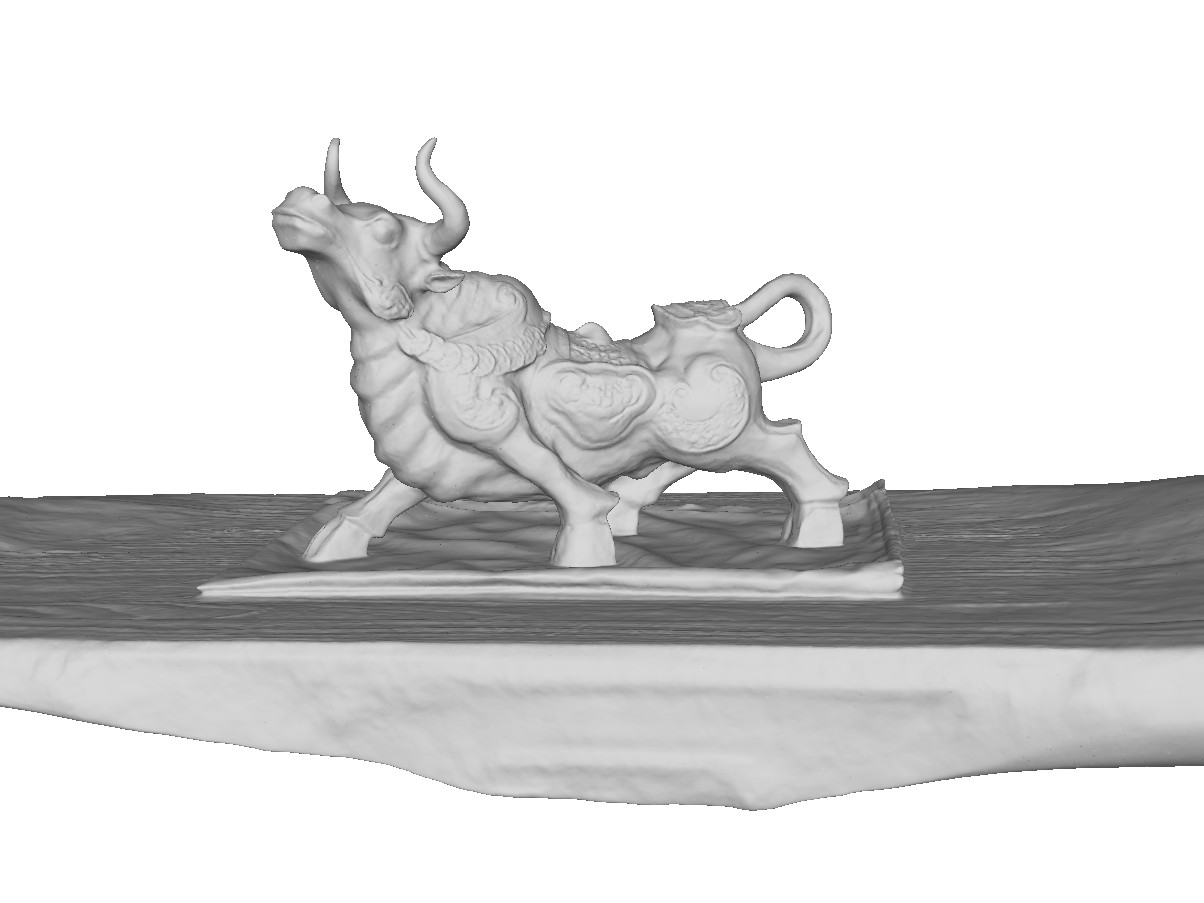}
    \end{subfigure}
    \begin{subfigure}[c]{0.3\textwidth}
        \includegraphics[width=\columnwidth]{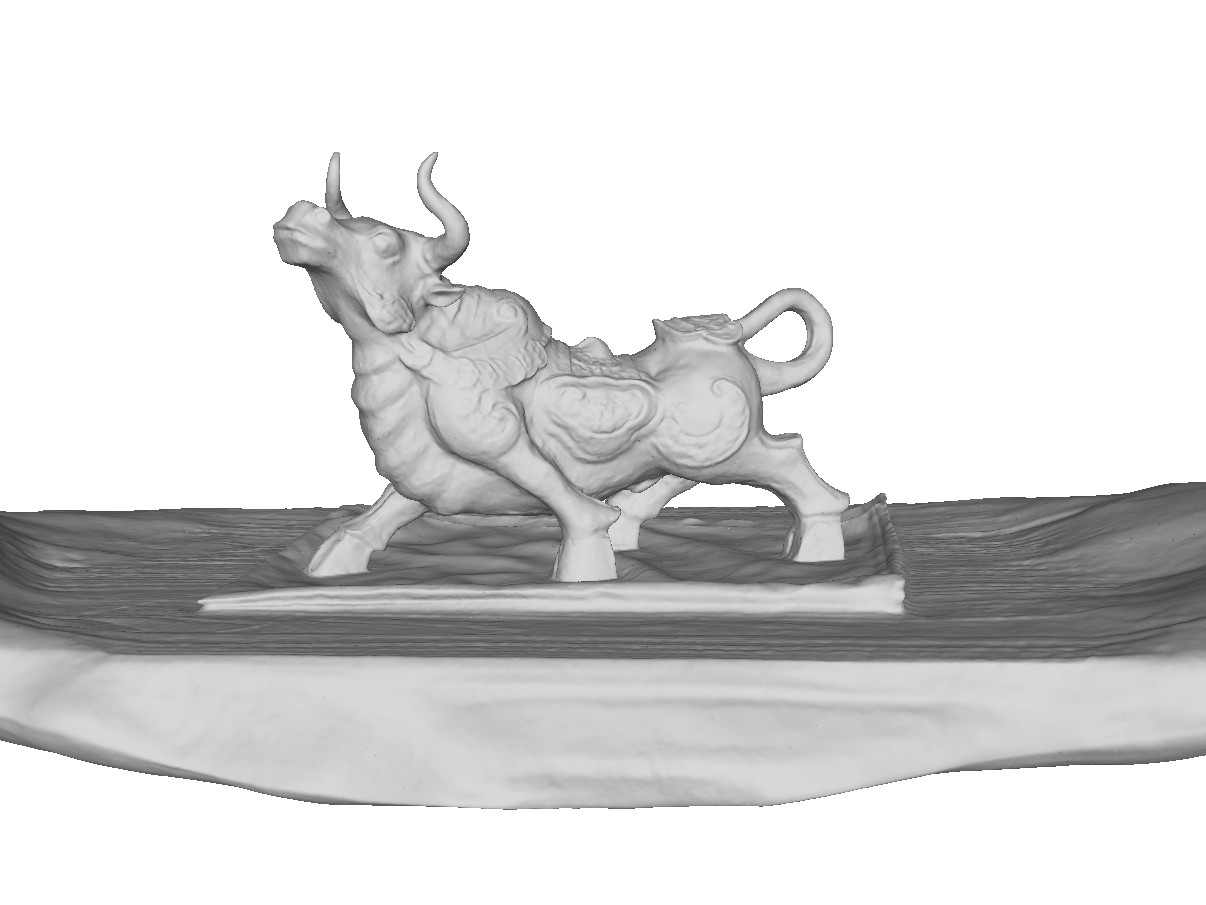}
    \end{subfigure}
    \begin{minipage}[c]{0.07\textwidth}
        \centering
        \rotatebox[origin=c]{90}{\textbf{Robot}}
    \end{minipage}
    \hfill
    \begin{subfigure}[c]{0.3\textwidth}
        \includegraphics[width=\columnwidth]{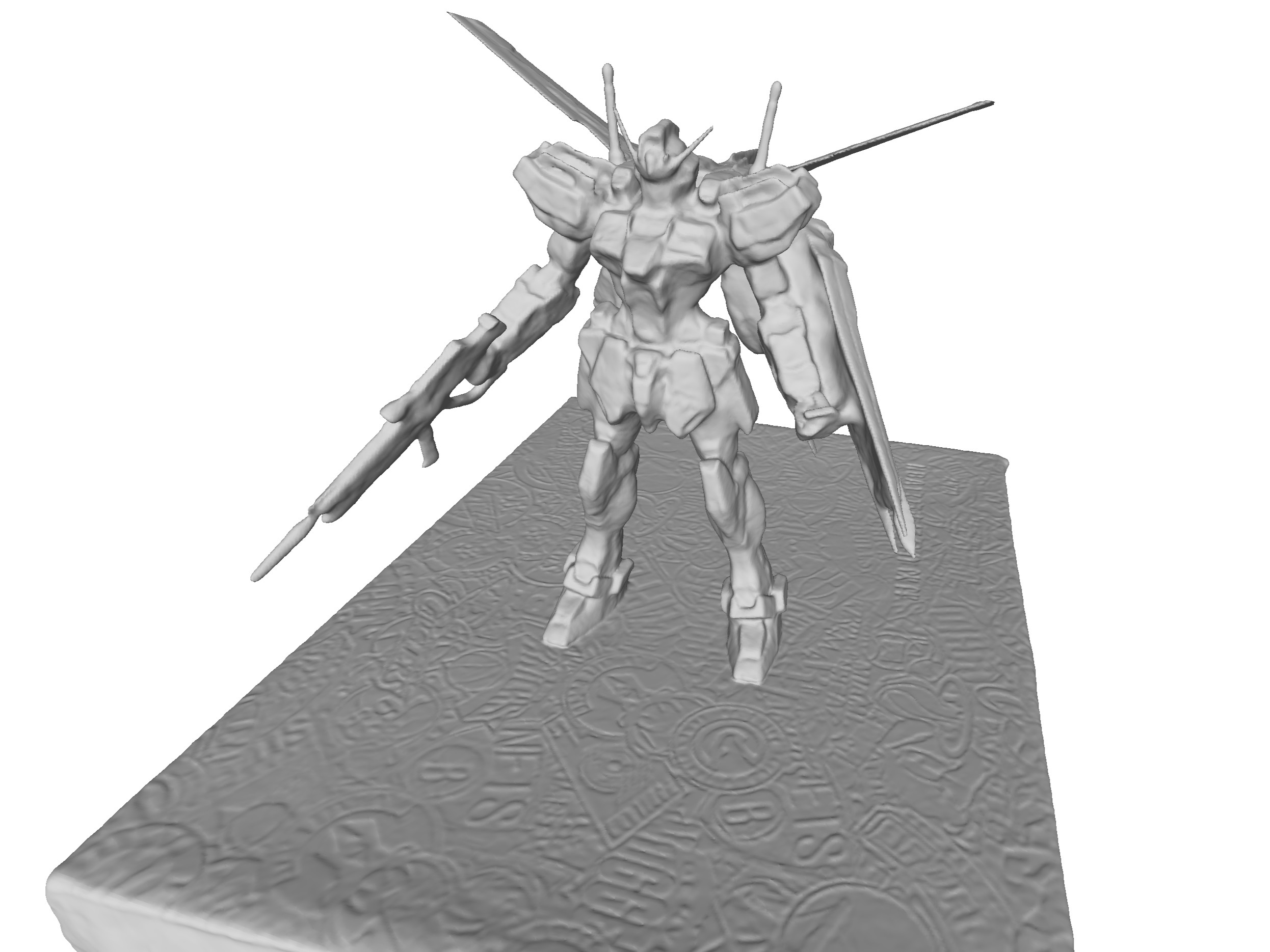}
        \caption{$\level~=6$}
        \label{fig:6freq_mesh}
    \end{subfigure}
    \begin{subfigure}[c]{0.3\textwidth}
        \includegraphics[width=\columnwidth]{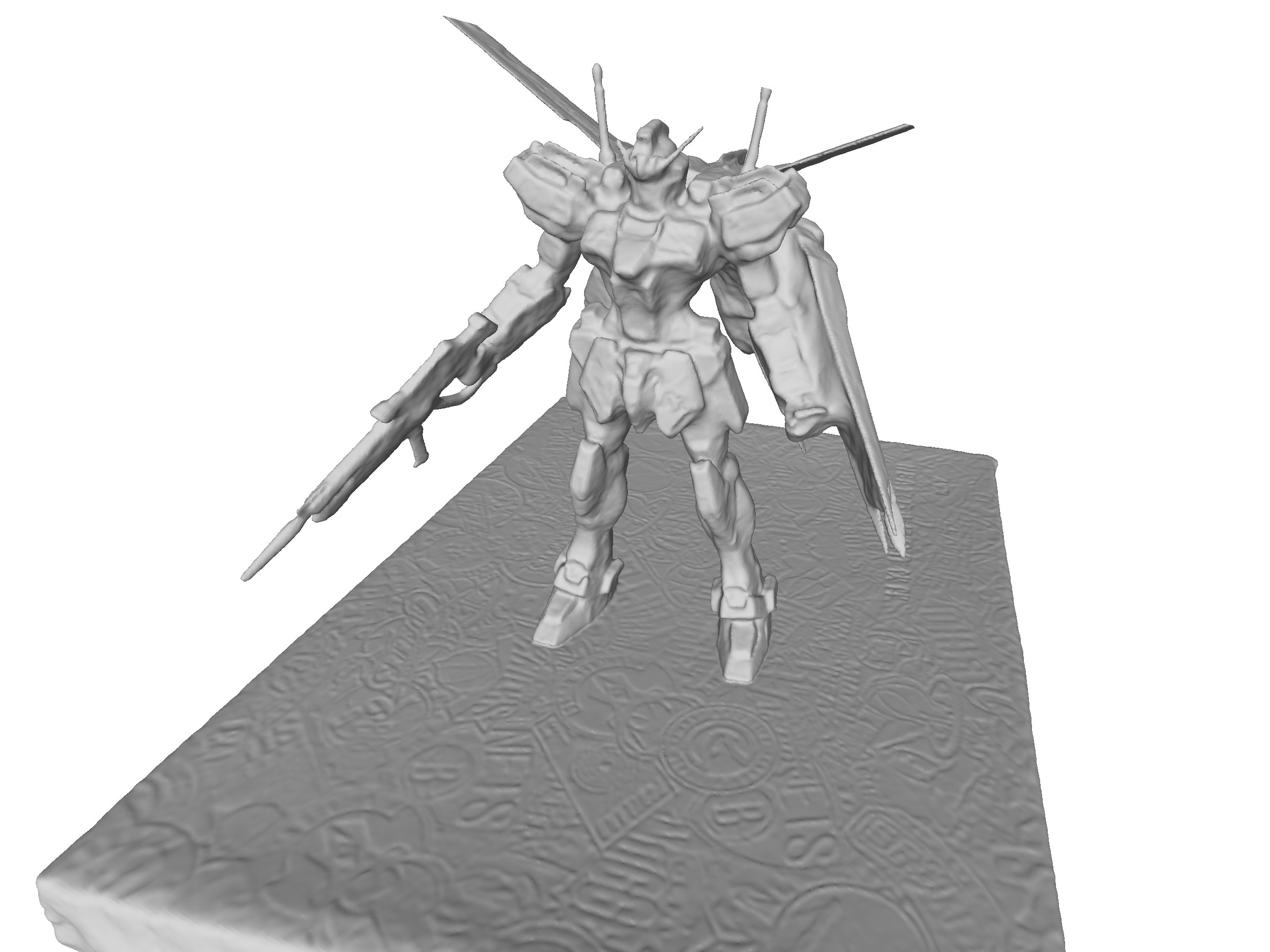}
        \caption{$\level~=9$}
        \label{fig:9freq_mesh}
    \end{subfigure}
    \begin{subfigure}[c]{0.3\textwidth}
        \includegraphics[width=\columnwidth]{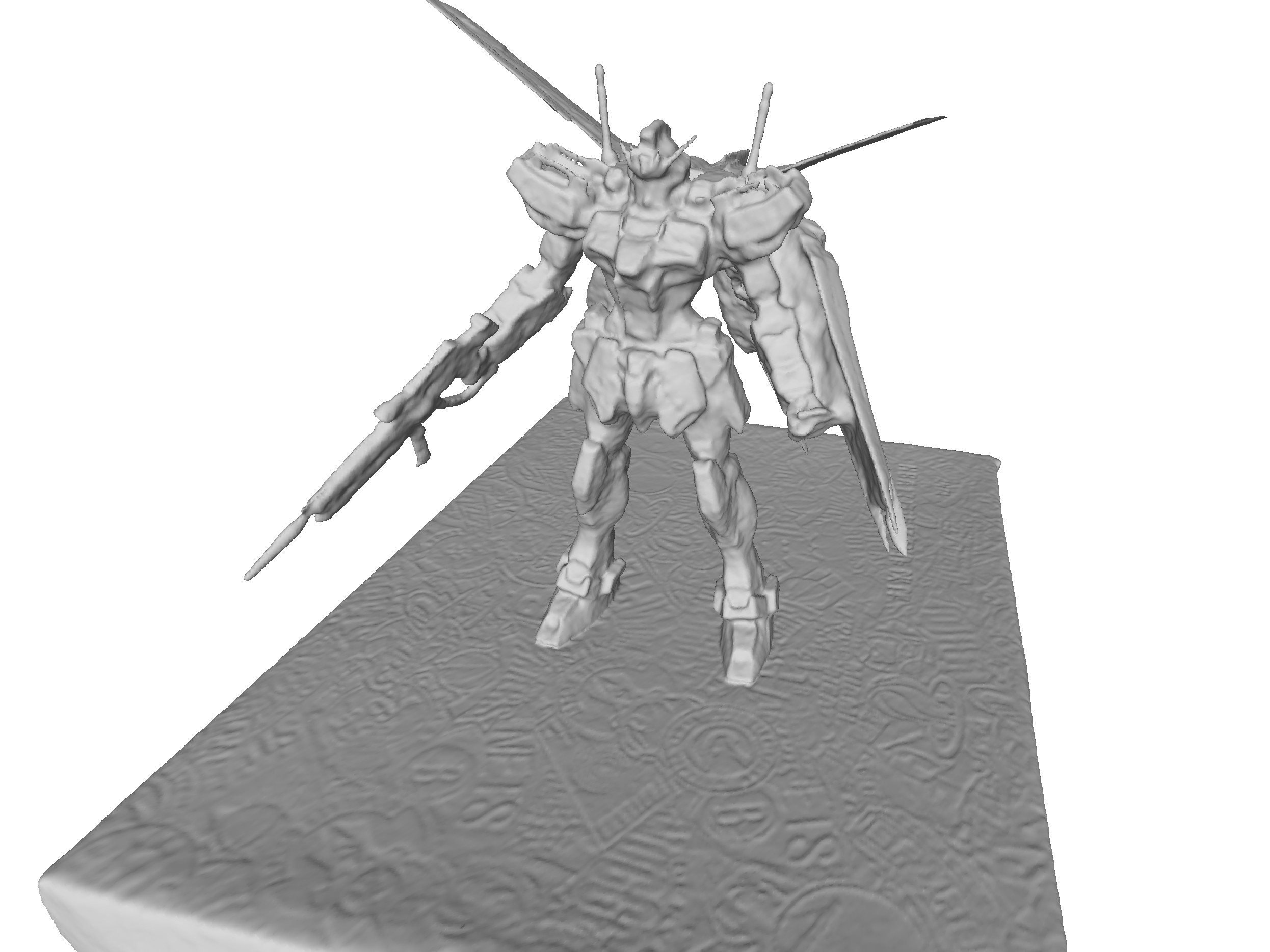}
        \caption{$\level~=12$}
        \label{fig:12freq_mesh}
    \end{subfigure}
    
    \caption{Qualitative comparison on surface reconstruction with a different number of frequencies.}
    \label{fig:freq_mesh}
\end{figure*}

\begin{figure*}[tb]
    \centering
    \begin{minipage}[c]{0.07\textwidth}
        \centering
        \rotatebox[origin=c]{90}{\textbf{Doll}}
    \end{minipage}
    \hfill
    \begin{subfigure}[c]{0.18\textwidth}
        \includegraphics[width=\columnwidth]{fig/doll_render_6freq.png}
    \end{subfigure}
    \begin{subfigure}[c]{0.18\textwidth}
        \includegraphics[width=\columnwidth]{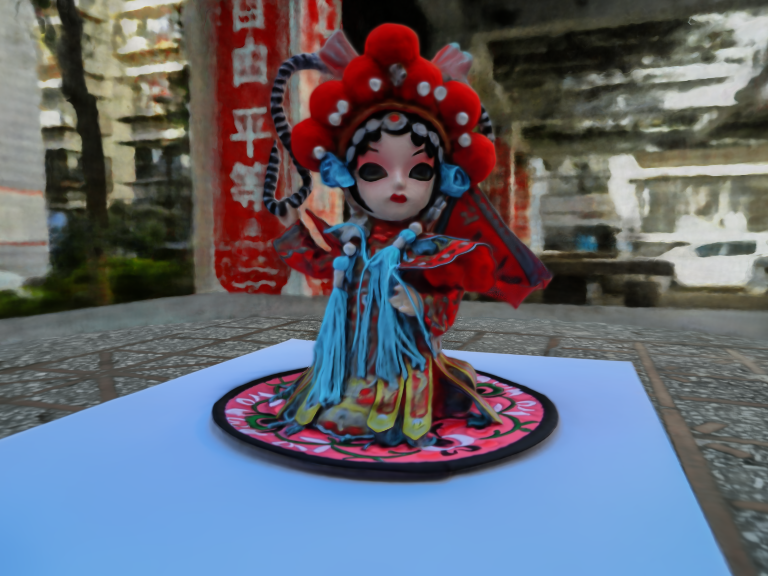}
    \end{subfigure}
    \begin{subfigure}[c]{0.18\textwidth}
        \includegraphics[width=\columnwidth]{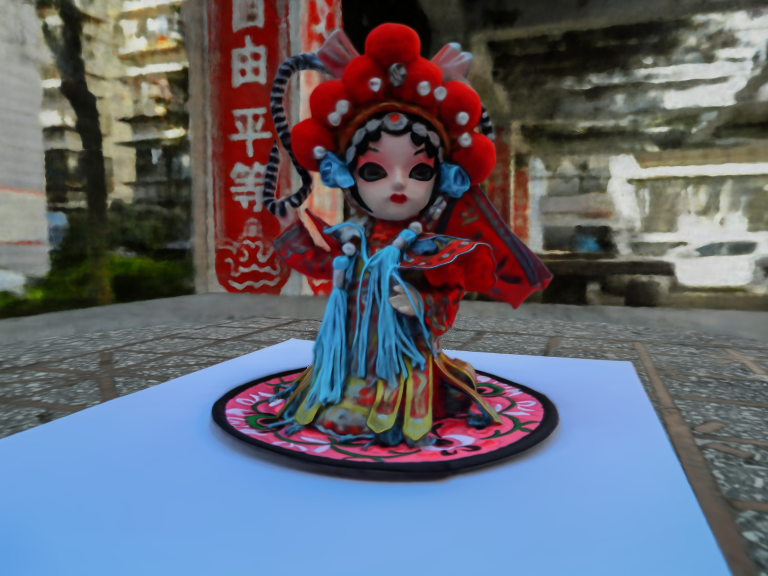}
    \end{subfigure}
    \begin{subfigure}[c]{0.18\textwidth}
        \includegraphics[width=\columnwidth]{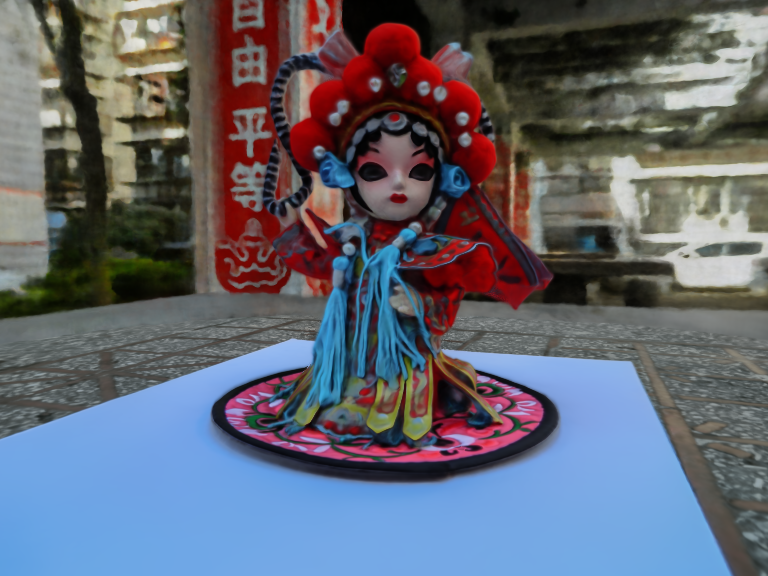}
    \end{subfigure}
    \begin{subfigure}[c]{0.18\textwidth}
        \includegraphics[width=\columnwidth]{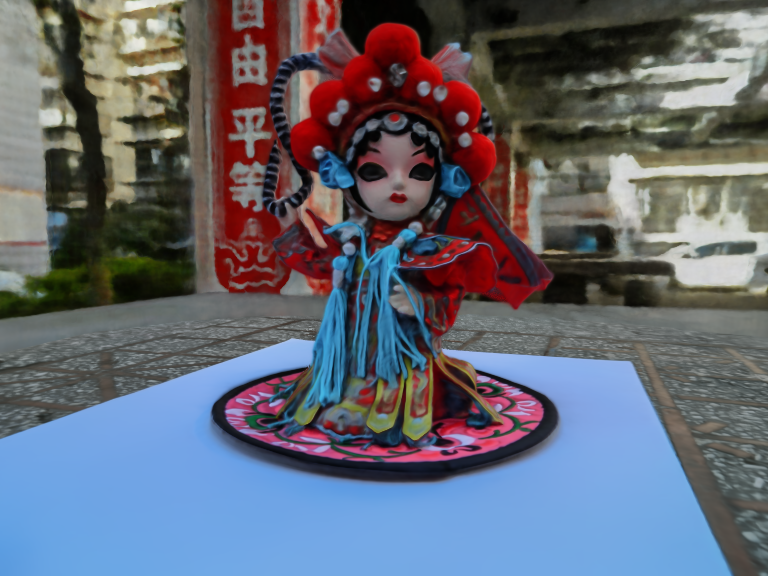}
    \end{subfigure}
    \begin{minipage}[c]{0.07\textwidth}
        \centering
        \rotatebox[origin=c]{90}{\textbf{Bull}}
    \end{minipage}
    \hfill
    \begin{subfigure}[c]{0.18\textwidth}
        \includegraphics[width=\columnwidth]{fig/bull_render_6freq.png}
    \end{subfigure}
    \begin{subfigure}[c]{0.18\textwidth}
        \includegraphics[width=\columnwidth]{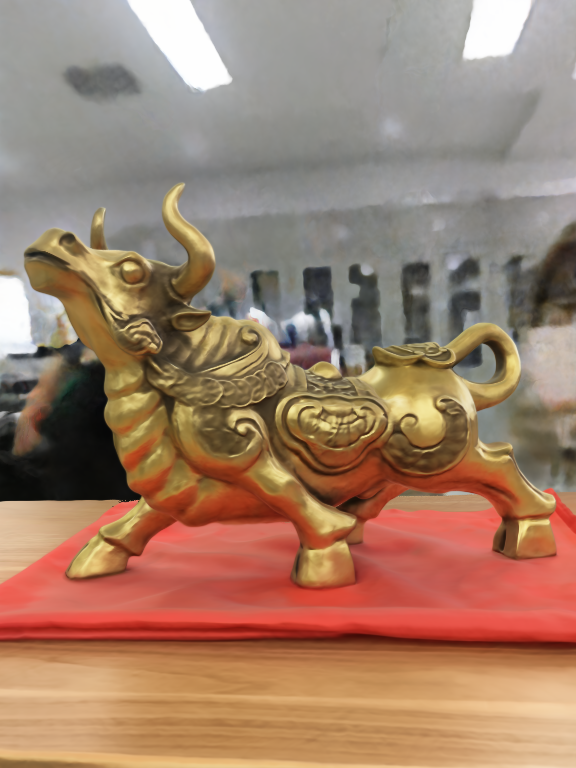}
    \end{subfigure}
    \begin{subfigure}[c]{0.18\textwidth}
        \includegraphics[width=\columnwidth]{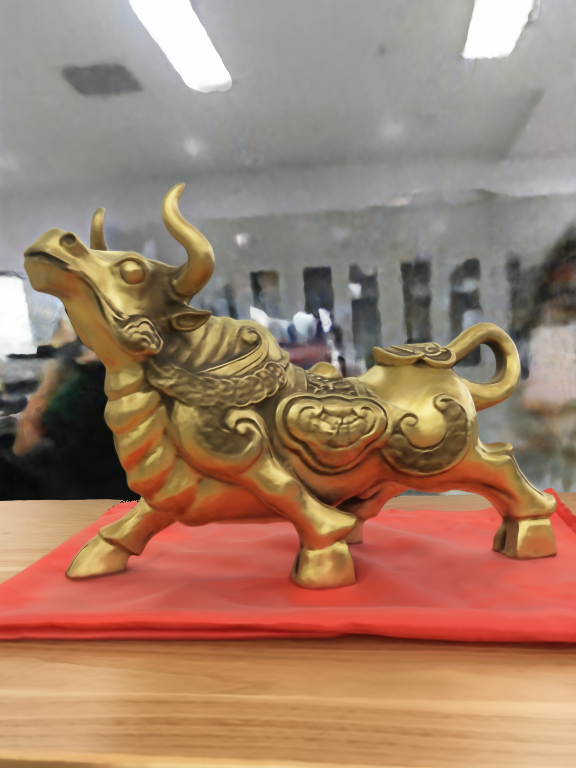}
    \end{subfigure}
    \begin{subfigure}[c]{0.18\textwidth}
        \includegraphics[width=\columnwidth]{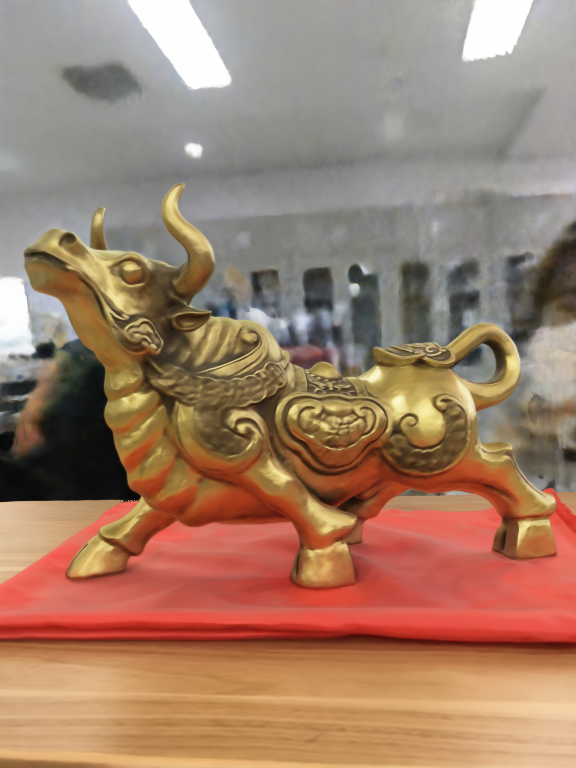}
    \end{subfigure}
    \begin{subfigure}[c]{0.18\textwidth}
        \includegraphics[width=\columnwidth]{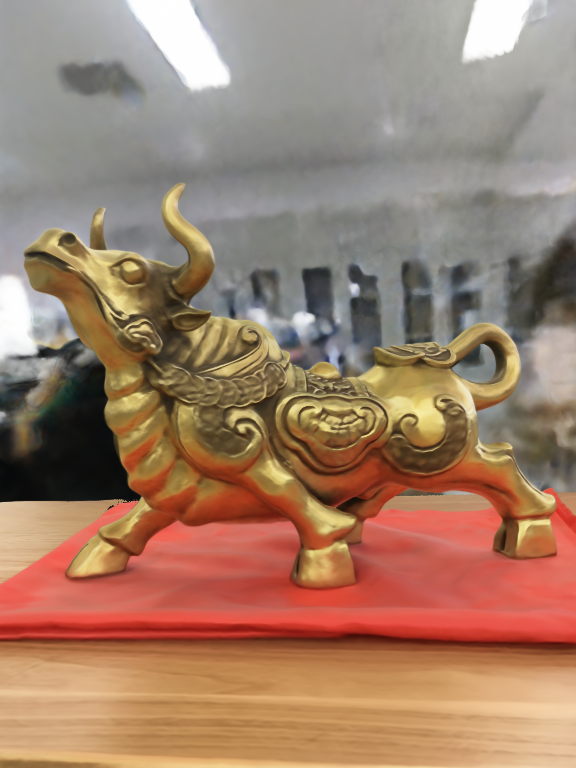}
    \end{subfigure}
    \begin{minipage}[c]{0.07\textwidth}
        \centering
        \rotatebox[origin=c]{90}{\textbf{Robot}}
    \end{minipage}
    \hfill
    \begin{subfigure}[c]{0.18\textwidth}
        \includegraphics[width=\columnwidth]{fig/robot_render_6freq.png}
        \caption{$(\rm{L,M,H})=(6,6,6)$}
        \label{fig:6layers}
    \end{subfigure}
    \begin{subfigure}[c]{0.18\textwidth}
        \includegraphics[width=\columnwidth]{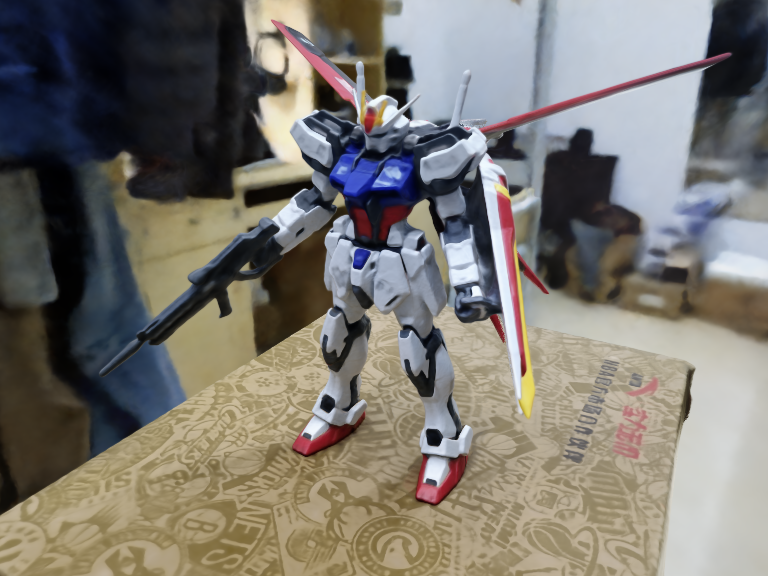}
        \caption{$(\rm{L,M,H})=(5,5,5)$}
        \label{fig:5layers}
    \end{subfigure}
    \begin{subfigure}[c]{0.18\textwidth}
        \includegraphics[width=\columnwidth]{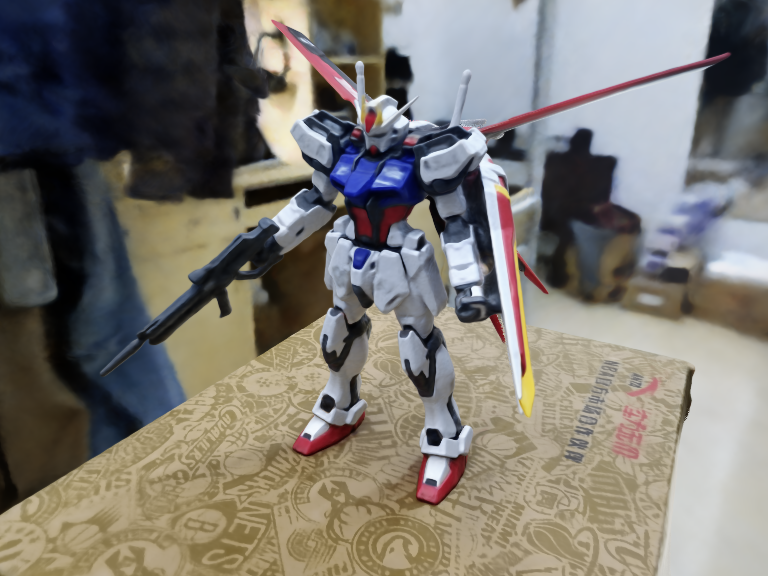}
        \caption{$(\rm{L,M,H})=(4,4,4)$}
        \label{fig:4layers}
    \end{subfigure}
    \begin{subfigure}[c]{0.18\textwidth}
        \includegraphics[width=\columnwidth]{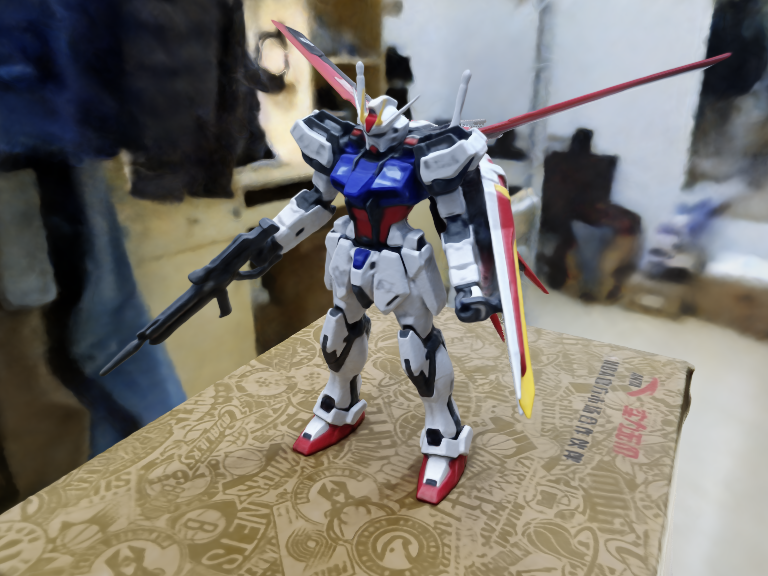}
        \caption{$(\rm{L,M,H})=(4,5,6)$}
        \label{fig:4_5_6layers}
    \end{subfigure}
    \begin{subfigure}[c]{0.18\textwidth}
        \includegraphics[width=\columnwidth]{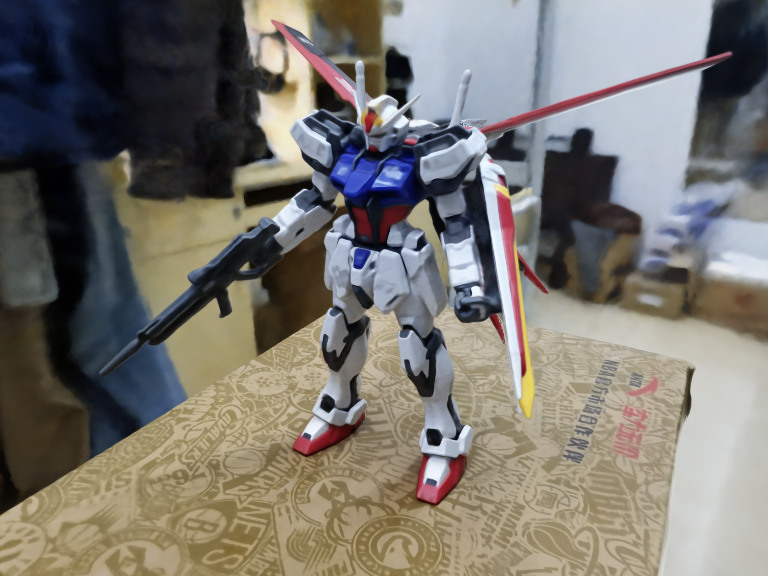}
        \caption{$(\rm{L,M,H})=(2,4,6)$}
        \label{fig:2_4_6layers}
    \end{subfigure}
    
    \caption{Qualitative comparison on viewpoint-based scene rendering using \method~, obtained by varying the number of encoder layers.}
    \label{fig:layers_render}
\end{figure*}

\begin{figure*}[tb]
    \centering
    \begin{minipage}[c]{0.07\textwidth}
        \centering
        \rotatebox[origin=c]{90}{\textbf{Doll}}
    \end{minipage}
    \hfill
    \begin{subfigure}[c]{0.18\textwidth}
        \includegraphics[width=\columnwidth]{fig/doll_mesh_6freq.jpg}
    \end{subfigure}
    \begin{subfigure}[c]{0.18\textwidth}
        \includegraphics[width=\columnwidth]{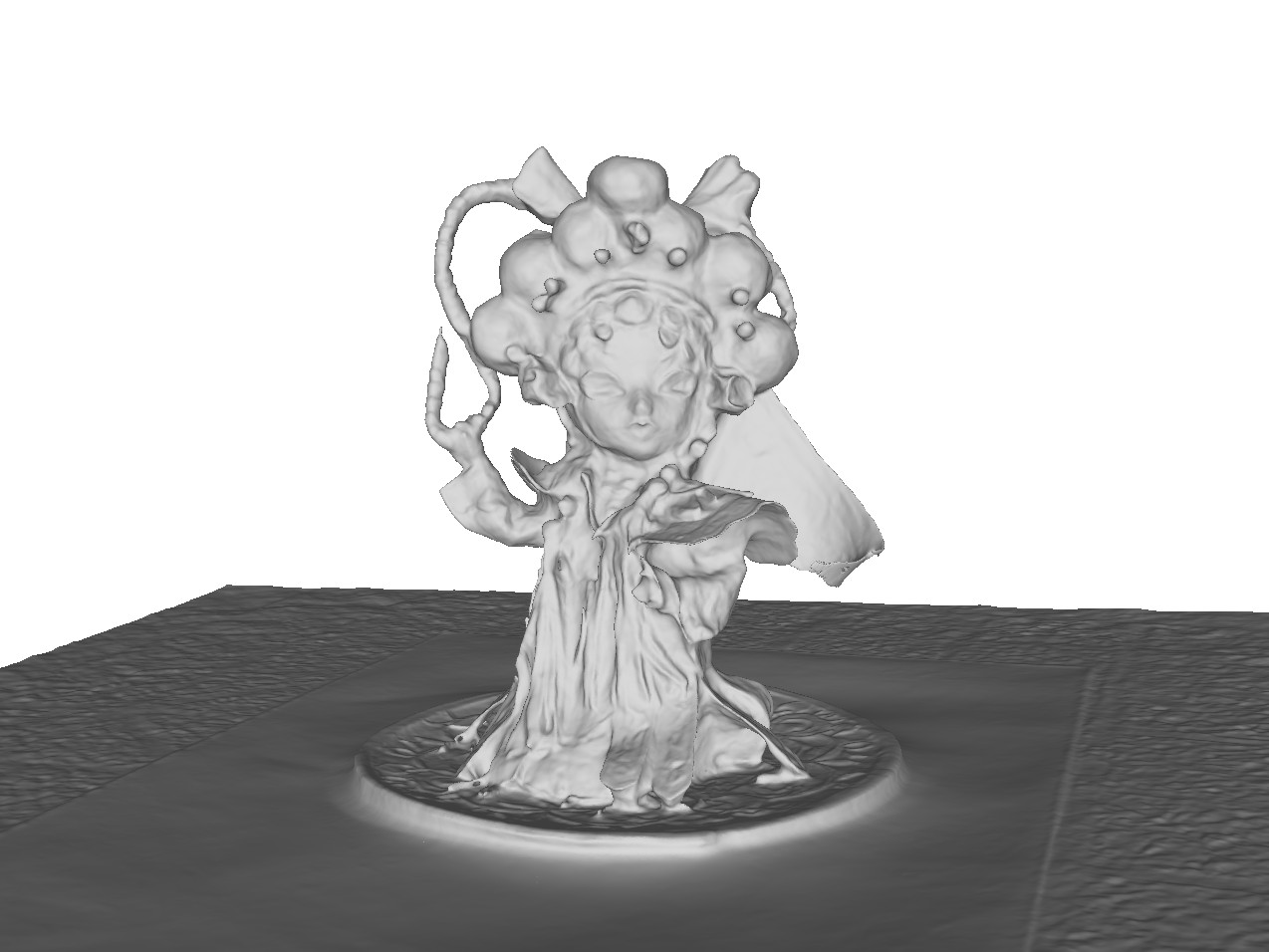}
    \end{subfigure}
    \begin{subfigure}[c]{0.18\textwidth}
        \includegraphics[width=\columnwidth]{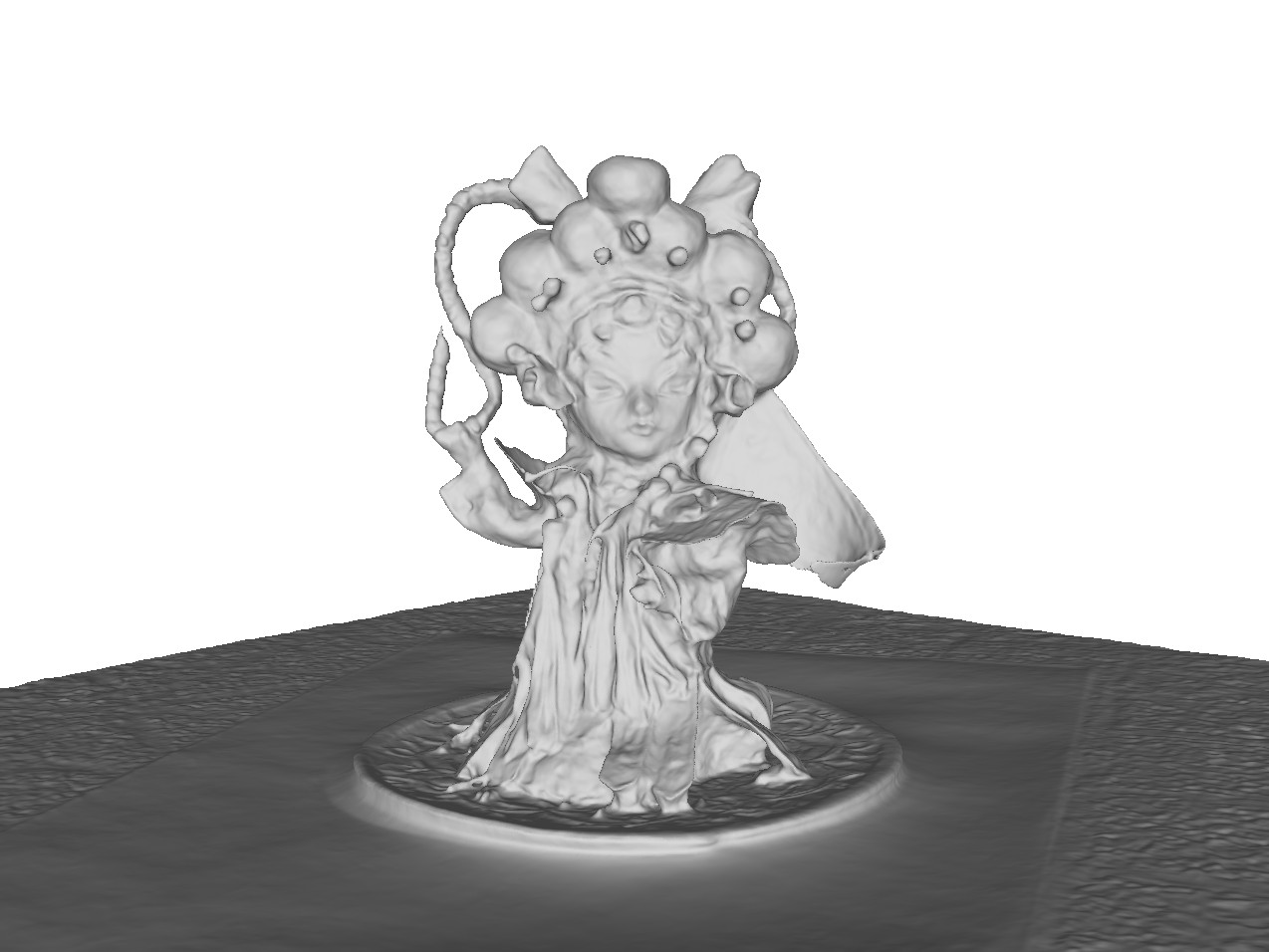}
    \end{subfigure}
    \begin{subfigure}[c]{0.18\textwidth}
        \includegraphics[width=\columnwidth]{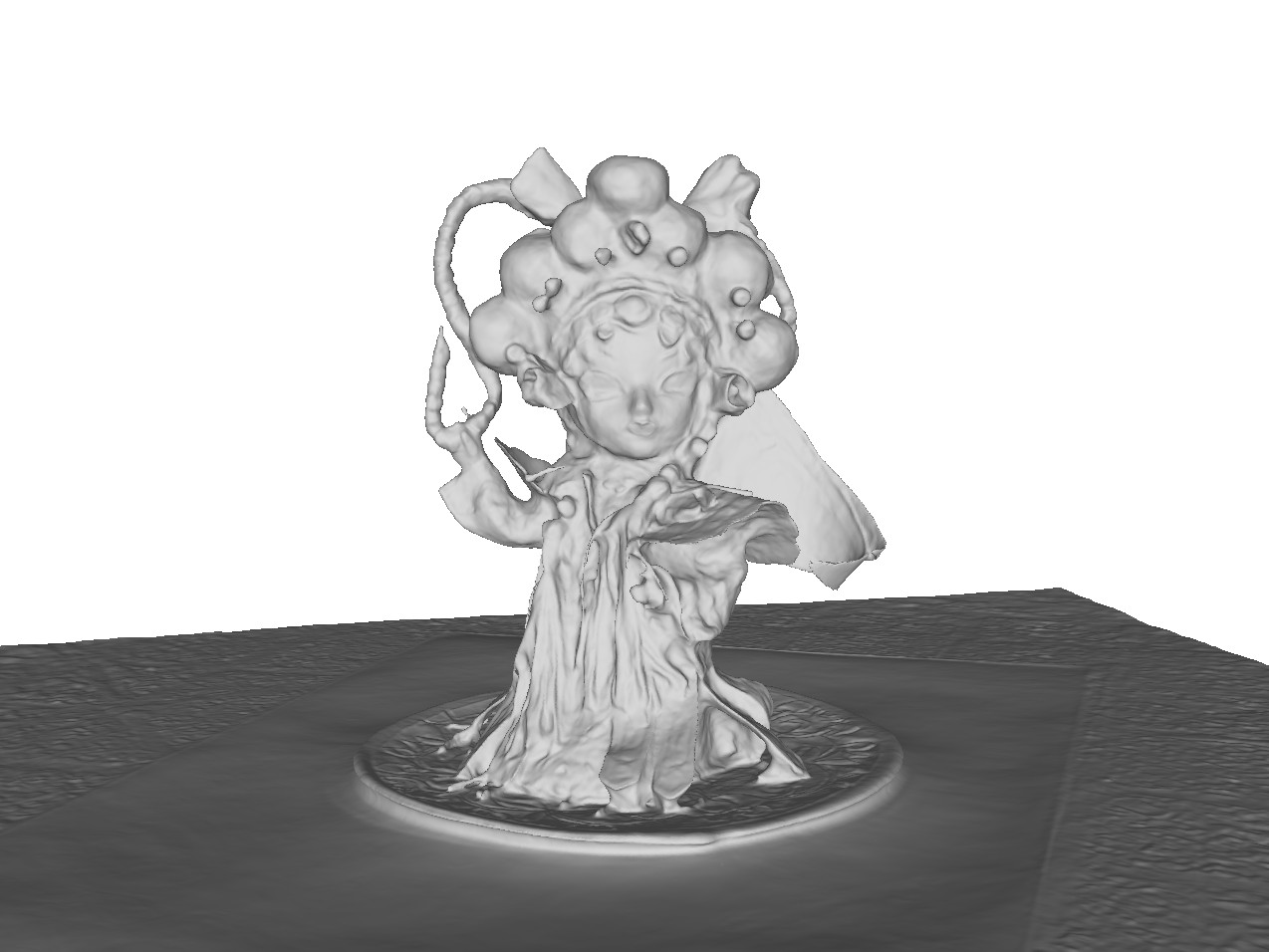}
    \end{subfigure}
    \begin{subfigure}[c]{0.18\textwidth}
        \includegraphics[width=\columnwidth]{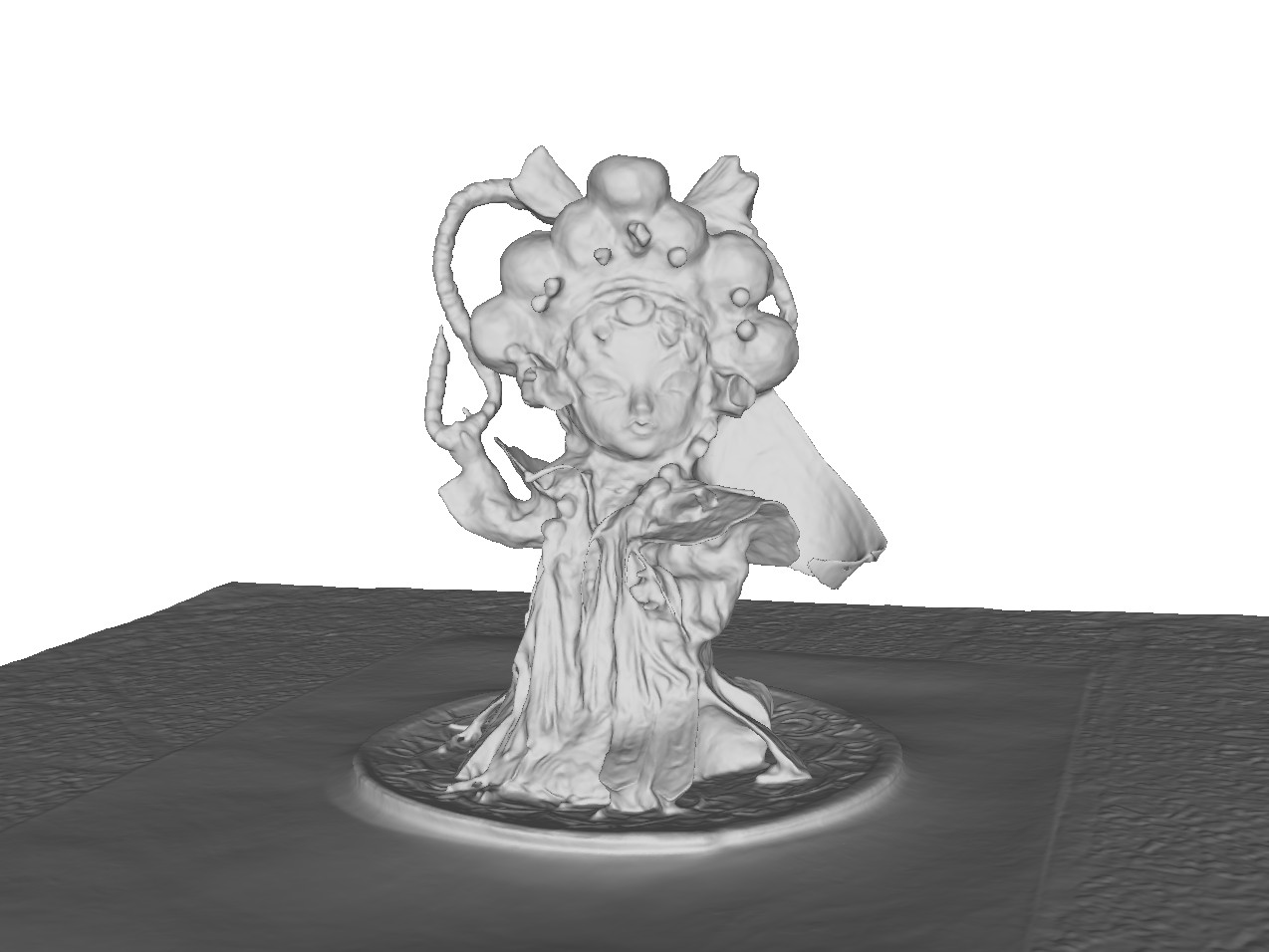}
    \end{subfigure}
    \begin{minipage}[c]{0.07\textwidth}
        \centering
        \rotatebox[origin=c]{90}{\textbf{Bull}}
    \end{minipage}
    \hfill
    \begin{subfigure}[c]{0.18\textwidth}
        \includegraphics[width=\columnwidth]{fig/bull_mesh_6freq.jpg}
    \end{subfigure}
    \begin{subfigure}[c]{0.18\textwidth}
        \includegraphics[width=\columnwidth]{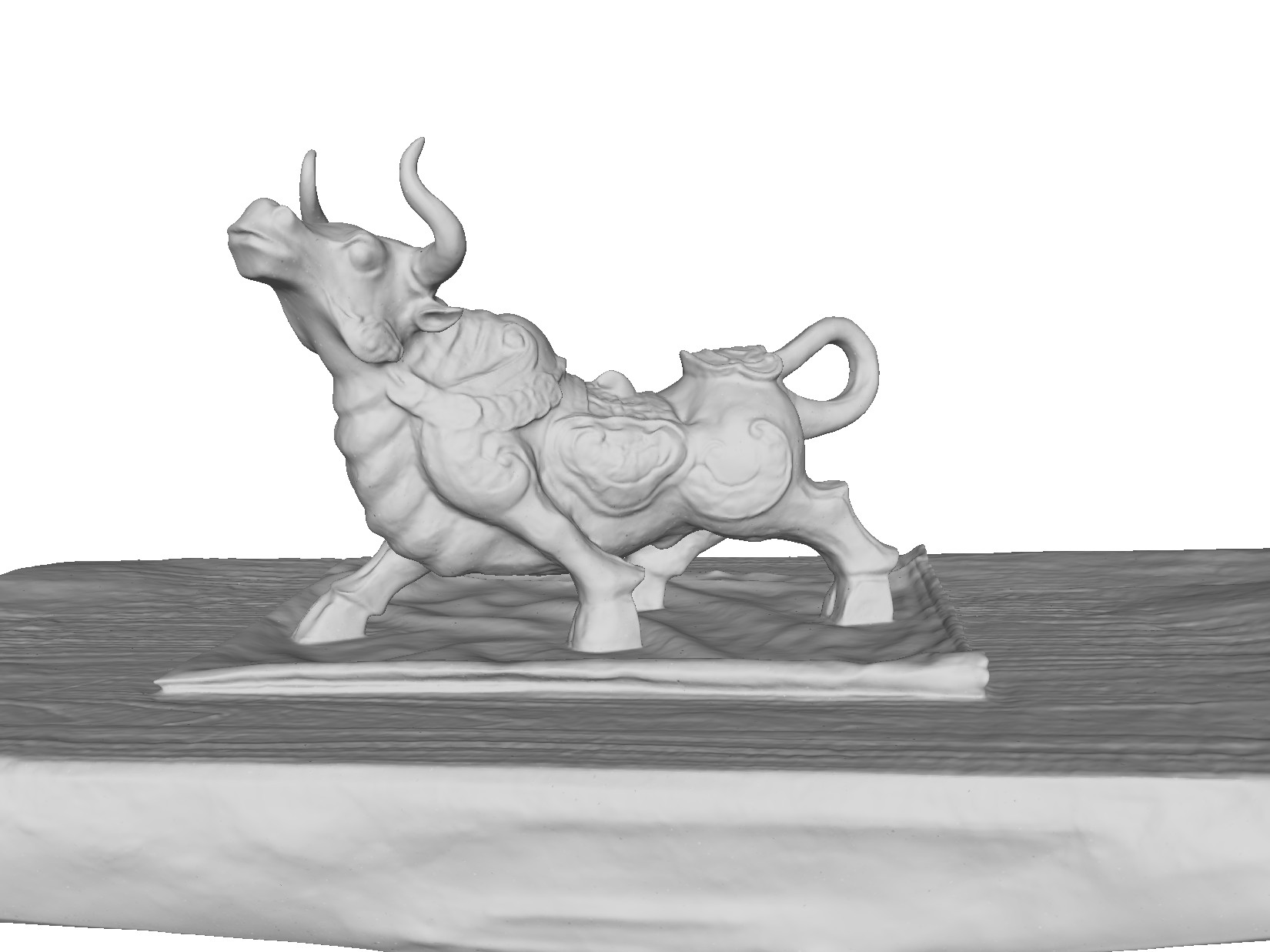}
    \end{subfigure}
    \begin{subfigure}[c]{0.18\textwidth}
        \includegraphics[width=\columnwidth]{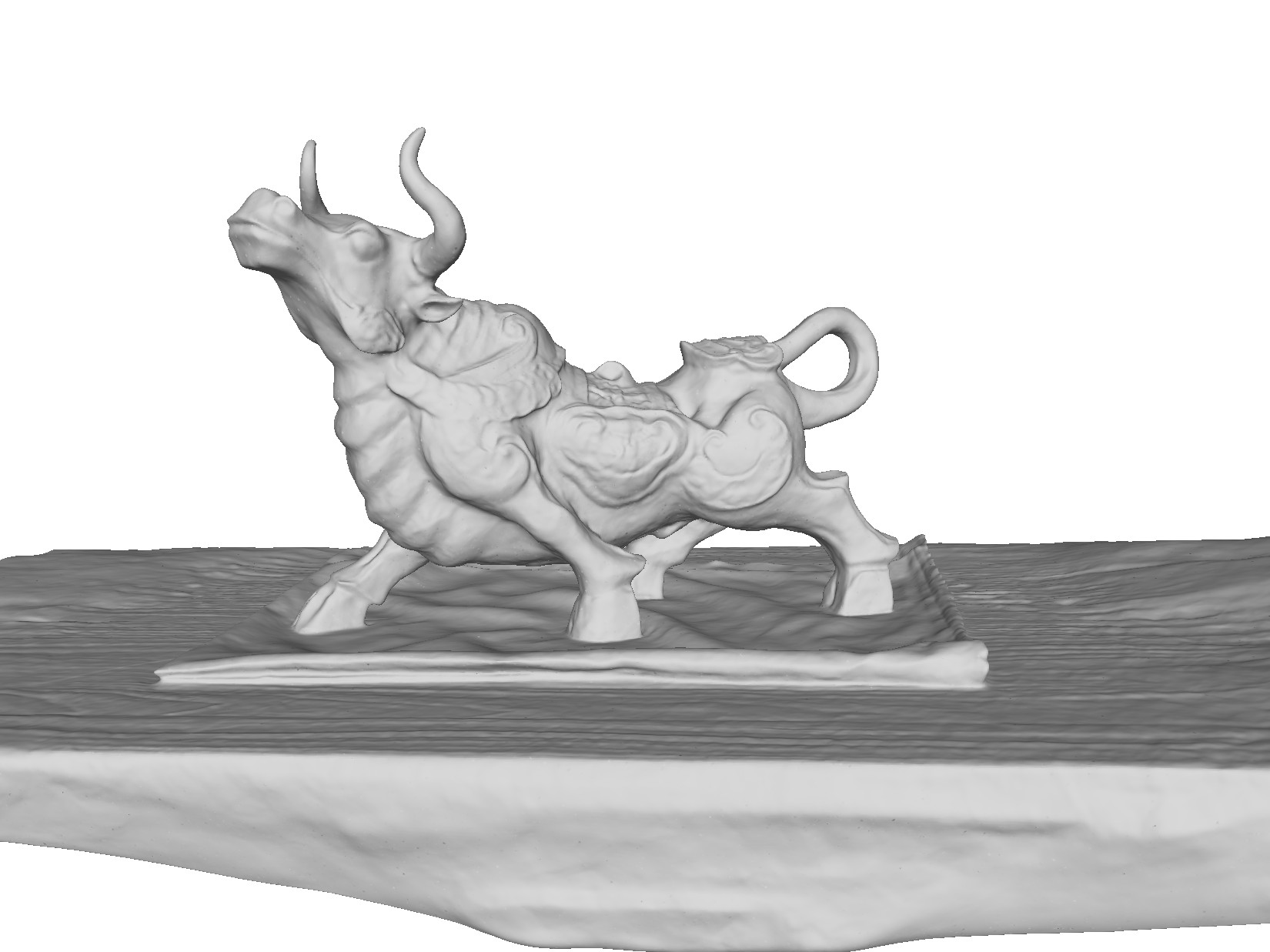}
    \end{subfigure}
    \begin{subfigure}[c]{0.18\textwidth}
        \includegraphics[width=\columnwidth]{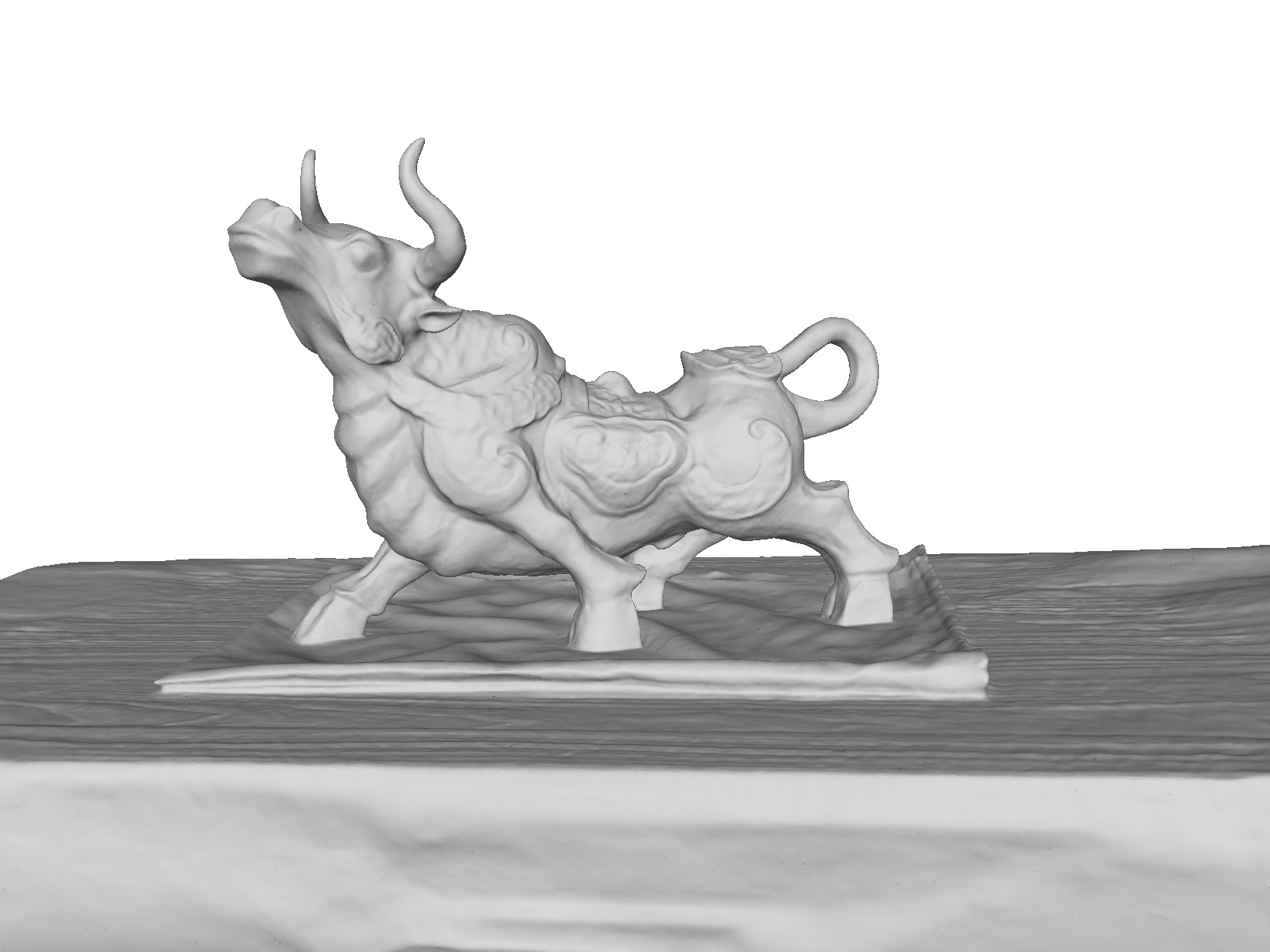}
    \end{subfigure}
    \begin{subfigure}[c]{0.18\textwidth}
        \includegraphics[width=\columnwidth]{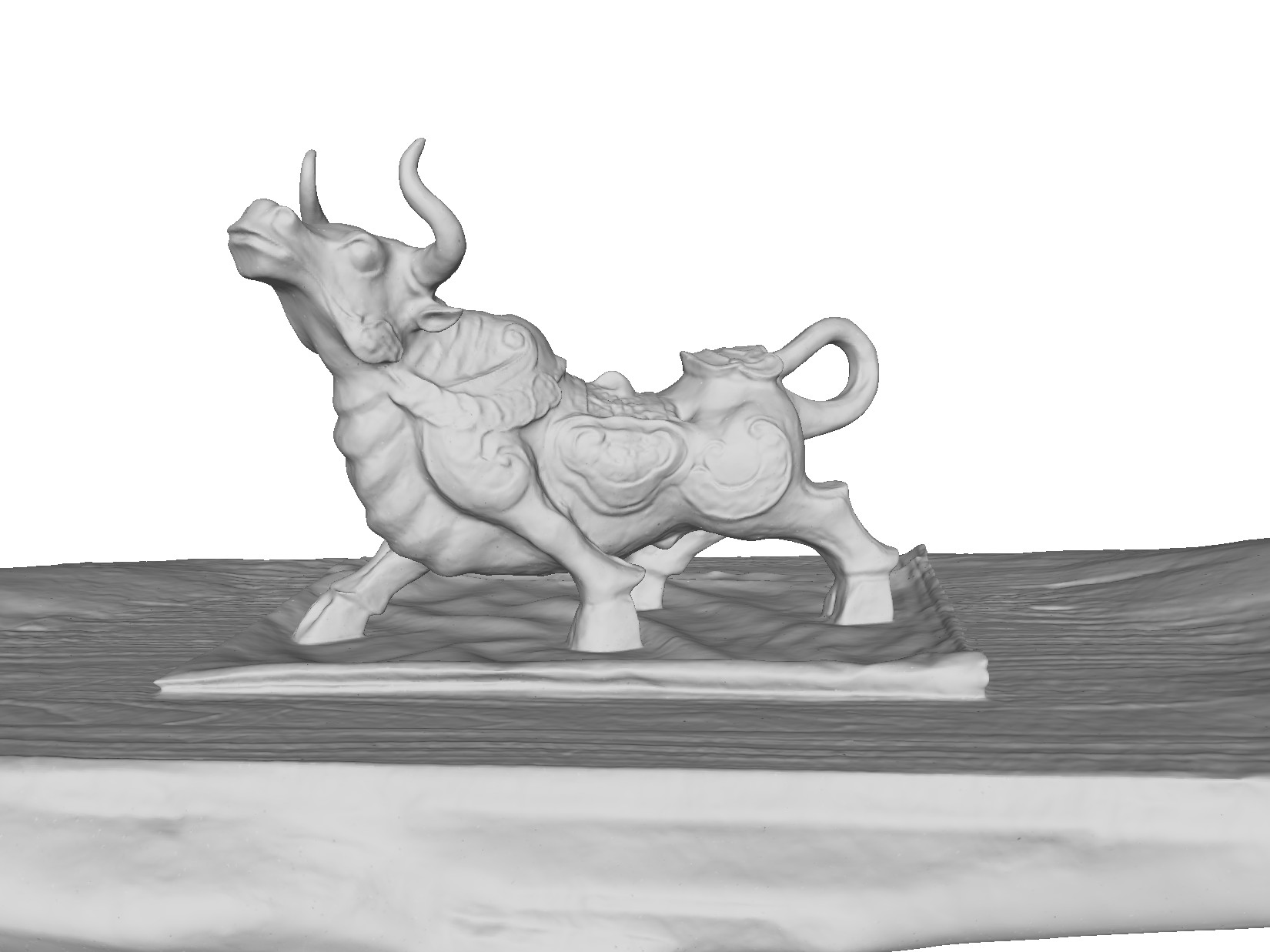}
    \end{subfigure}
    \begin{minipage}[c]{0.07\textwidth}
        \centering
        \rotatebox[origin=c]{90}{\textbf{Robot}}
    \end{minipage}
    \hfill
    \begin{subfigure}[c]{0.18\textwidth}
        \includegraphics[width=\columnwidth]{fig/robot_mesh_6freq.jpg}
        \caption{$(\rm{L,M,H})=(6,6,6)$}
        \label{fig:6layers_mesh}
    \end{subfigure}
    \begin{subfigure}[c]{0.18\textwidth}
        \includegraphics[width=\columnwidth]{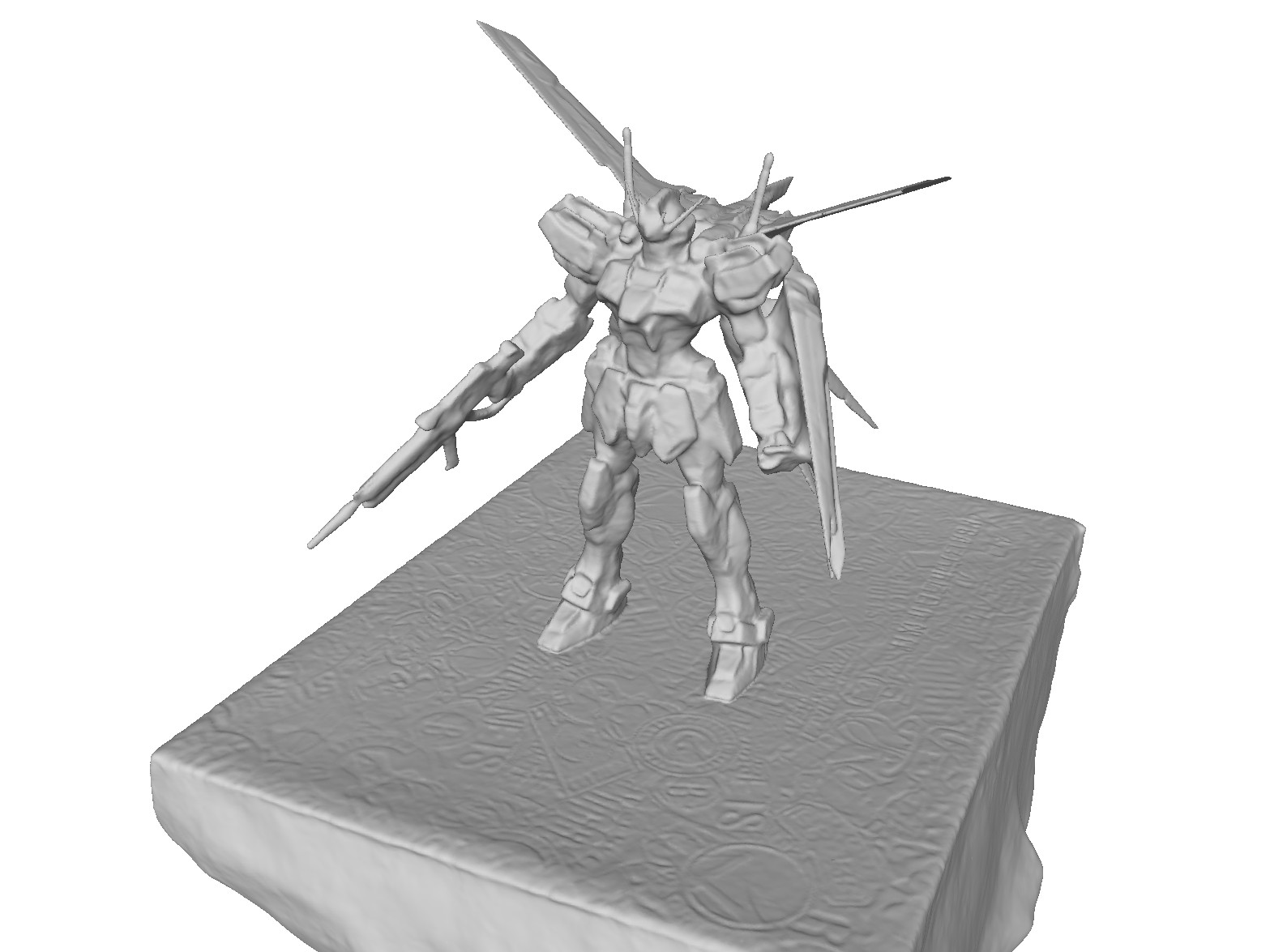}
        \caption{$(\rm{L,M,H})=(5,5,5)$}
        \label{fig:5layers_mesh}
    \end{subfigure}
    \begin{subfigure}[c]{0.18\textwidth}
        \includegraphics[width=\columnwidth]{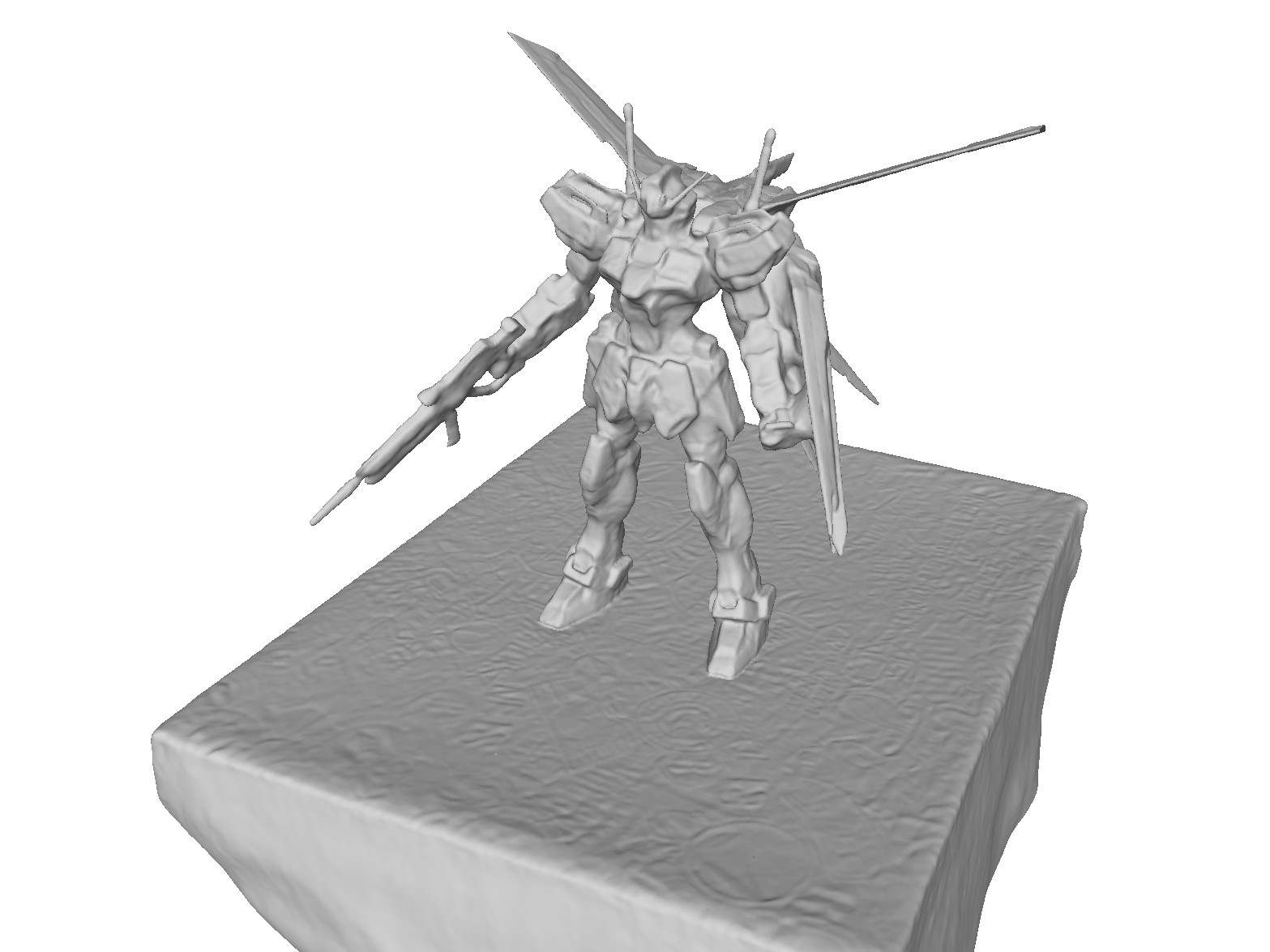}
        \caption{$(\rm{L,M,H})=(4,4,4)$}
        \label{fig:4layers_mesh}
    \end{subfigure}
    \begin{subfigure}[c]{0.18\textwidth}
        \includegraphics[width=\columnwidth]{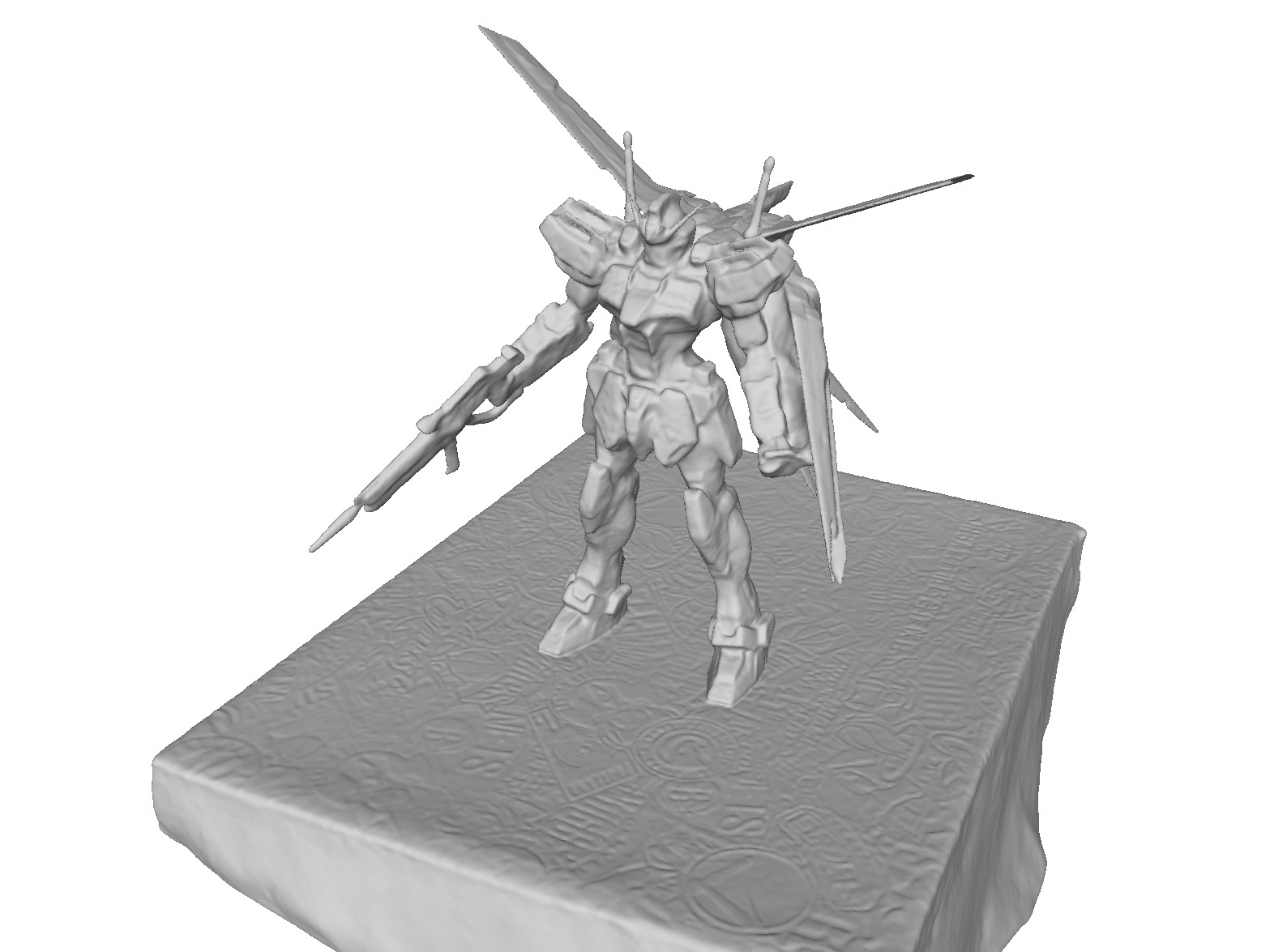}
        \caption{$(\rm{L,M,H})=(4,5,6)$}
        \label{fig:4_5_6layers_mesh}
    \end{subfigure}
    \begin{subfigure}[c]{0.18\textwidth}
        \includegraphics[width=\columnwidth]{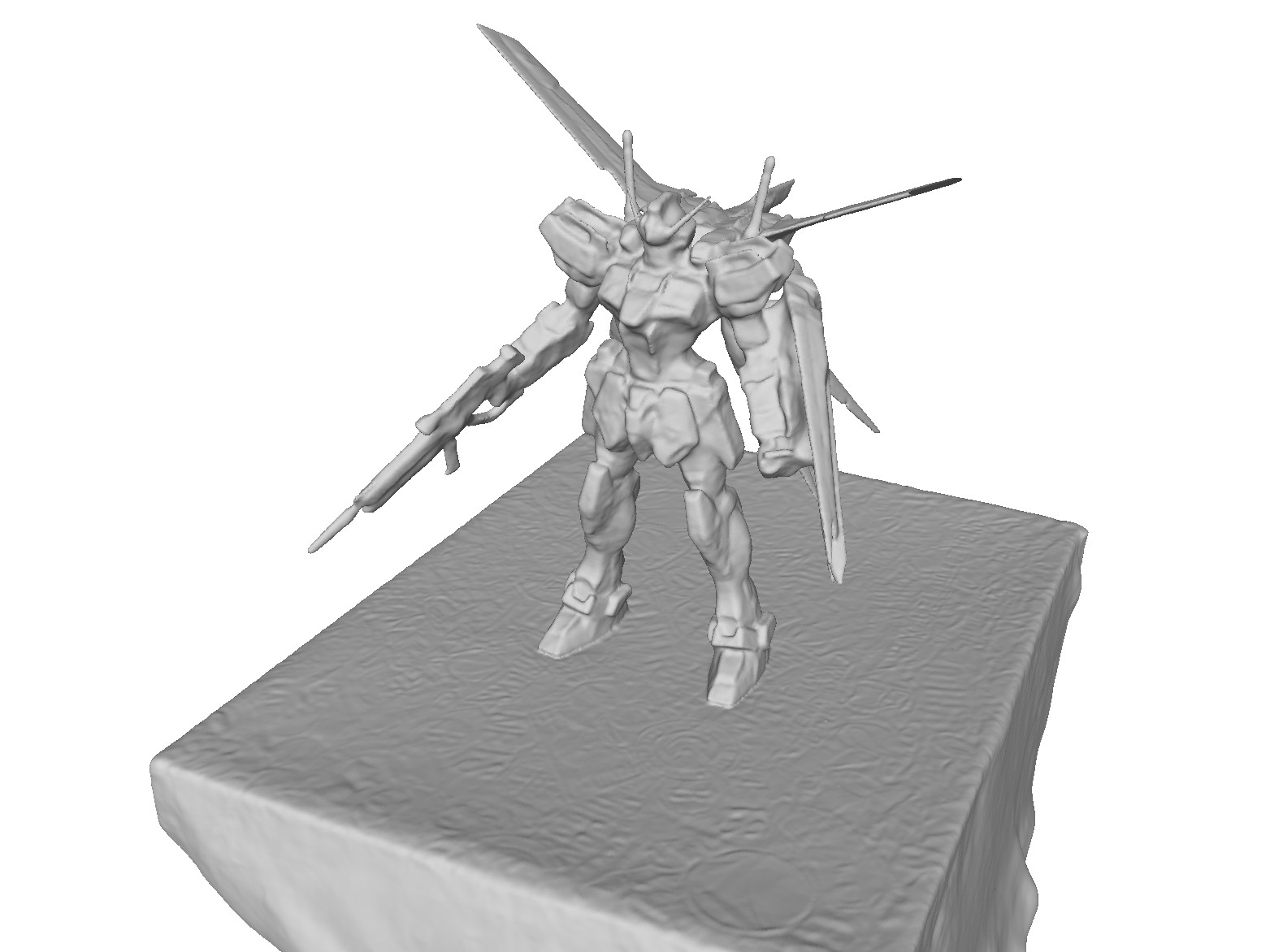}
        \caption{$(\rm{L,M,H})=(2,4,6)$}
        \label{fig:2_4_6layers_mesh}
    \end{subfigure}
    
    \caption{Qualitative comparison based on 3D surface reconstruction using \method~, obtained by varying the number of encoder layers.}
    \label{fig:layers_mesh}
\end{figure*}

\section{Ablation study of the redundancy-aware weighting module}
A key innovation of \method~ is the \emph{rendundancy-aware weighting} module which combines the complementary information from the different encoders by promoting mutual dissimilarity. Table~\ref{tab:ablaiton_redundancy-aware_supple} shows quantitative comparison results of the \method~ with and without this module. The results show that our model with this module outperforms the variant without it, where a simple averaging of the encoder features is performed, clearly bringing out its effectiveness.

\section{Comparative study of the number of frequency levels}
We conduct experiments to study the effect of the choice of frequency levels $\level~$ for both \method~ and Scaled-up VolSDF~\cite{yariv_volume_2021}. 
As shown in Table~\ref{table:ablation_frequencylevel} and Fig.~\ref{fig:comparisonfrequency},
the Scaled-up VolSDF is sensitive to the choice of frequency levels and has particular difficulty in dealing with higher frequency encodings.
In particular, the Scaled-up VolSDF with $\level~=9$ results in a reconstructed mesh with too many bumps,
while that with $\level~=12$ results in a mesh that is hard to interpret.
On the other hand, \method~ is capable of processing higher--frequency information without sacrificing information gleaned from the low--frequency bands.
Fig.~\ref{fig:freq_render} and \ref{fig:freq_mesh} show the qualitative comparisons of rendered images and reconstructed meshes with $\level~=6,9,12$ using \method~ on the \textit{Doll}, \textit{Bull}, and \textit{Robot} scenes.

\section{Comparative study of encoder architecture variants}
In order to design the encoders of \method~ optimally, we study the effect of varying the number of layers of each of the three encoders of \method~ 
and compare their performances.
As seen from the results in Table~\ref{table:ablation_layernumber} as well as   Fig.~\ref{fig:layers_render}, and Fig.~\ref{fig:layers_mesh}.
\method~ performs comparably irrespective of the choice of encoder architecture, maintaining a good performance throughout. %
Based on this analysis and in order to stay consistent with the baseline VolSDF~\cite{yariv_volume_2021} architecture, we choose the 6--layer architecture for each encoder, with each layer having 256 dimensions.

\end{document}